%% file: main.tex
\documentclass[10pt,twocolumn,letterpaper]{article}

\usepackage[pagenumbers]{cvpr} % To force page numbers, e.g. for an arXiv version

\newcommand{\myparagraph}[1]{\vspace{1pt}\noindent{\bf{#1}}~~}

\input{preamble}

\definecolor{cvprblue}{rgb}{0.21,0.49,0.74}
\usepackage[pagebackref,breaklinks,colorlinks,allcolors=cvprblue]{hyperref}

\def\confName{CVPR}
\def\confYear{2027}

\newcommand{\eaglewordmark}{{\bfseries
  \textcolor[HTML]{204E56}{E}\kern0.045em%
  \textcolor[HTML]{1D6067}{A}\kern0.045em%
  \textcolor[HTML]{1C7180}{G}\kern0.045em%
  \textcolor[HTML]{276B93}{L}\kern0.045em%
  \textcolor[HTML]{345D9F}{E}%
}}

\title{\texorpdfstring{\eaglewordmark}{EAGLE}:
\underline{E}vidence-\underline{A}ware
\underline{G}enerative \underline{L}at\underline{E}nts
for Any-Modal Earth Observation}

\author{%
Buddhi Wijenayake$^{1}$ \quad Rohita Mocharla$^{1,2}$ \quad Dhananjaya Jayasundara$^{1}$\quad Vishal M. Patel$^{1}$\\
$^1$Johns Hopkins University \quad 
$^2$Johns Hopkins Applied Physics Lab\\
\texttt{\{bwarnas1,nmochar1,vjayasu1,vpatel36\}@jhu.edu}\\
\href{https://buddhi19.github.io/EAGLE}{Buddhi19.github.io/EAGLE}
}
\begin{document}
\maketitle

\input{sec/0_abstract}

\input{sec/1_intro}

\begin{figure*}[t]
    \centering
    \includegraphics[width=1\linewidth]{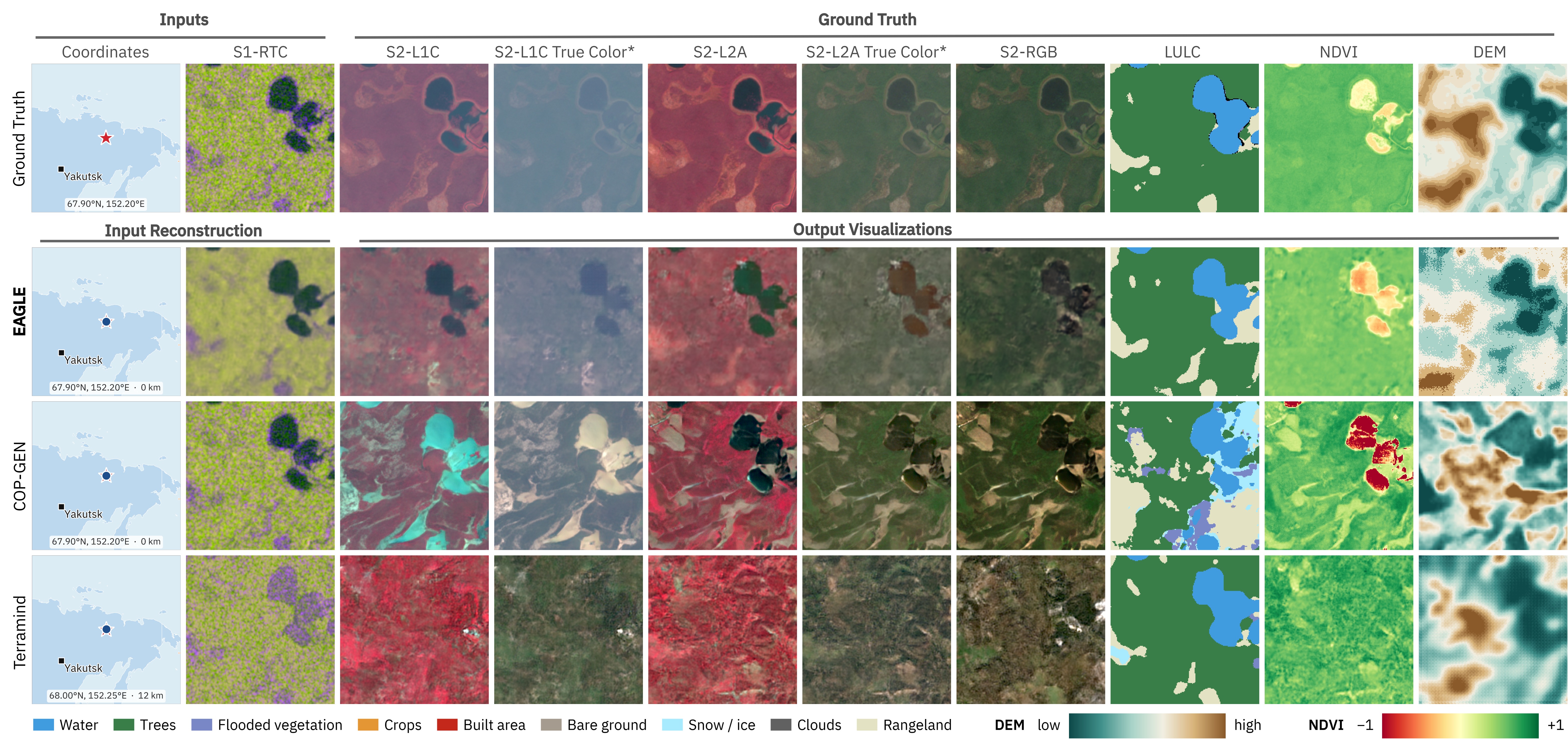}
  \vskip-8pt  \caption{\textbf{Cross-modal generation from S1 RTC and coordinates.}
The top row shows the ground truth.
$^\ast$True-colour views of S2 L1C and L2A are included
for visual comparison.}
    \label{fig:qa}
\end{figure*}

\input{sec/2_relatedwork}

\input{Tables/loo}

\input{sec/3_methodology}

\input{sec/4_experiments}

\input{sec/5_results_and_discussion}

\input{sec/6_conclusion}
% \input{sec/2_formatting}
% \input{sec/3_finalcopy}

% \section*{LLM Usage Statement}
% We used OpenAI’s GPT-6 Astra to generate the paper logo and assist with manuscript drafting and revision, coding, and the organisation, summarisation and cross-checking of a large collection of experimental results. We also used Anthropic’s Claude Code (Opus 5.5) for coding assistance. All AI-assisted outputs, including proposed text, code and result summaries, were reviewed and verified by the authors and revised where necessary. The authors developed the research ideas and made the methodological decisions and scientific judgements. The authors remain responsible for the accuracy and integrity of the final manuscript, including the reported results, their interpretation and the conclusions drawn from them.

{
    \small
    \bibliographystyle{ieeenat_fullname}
    \bibliography{main}
}

\input{sec/X_suppl}

\end{document}

%% file: preamble.tex
\usepackage{tabularx}
\usepackage{longtable}
\usepackage{booktabs,array,pifont}
\usepackage{multirow}  % Tables/best20.tex
\usepackage{xcolor}
\usepackage{colortbl}
\usepackage{adjustbox}

%% file: sec/0_abstract.tex
\begin{abstract}
% Earth observation sensors provide complementary views of a scene, but which sensors are available varies across locations and dates. We introduce Evidence-Aware Generative LatEnts (EAGLE), which combines available observations into a shared scene latent for cross-modal generation and scene understanding. EAGLE models encoded optical, SAR and elevation features as noisy linear views of a common latent, so any subset of sensors can be fused in closed form through a linear-Gaussian posterior. From this posterior, a deterministic predictor estimates the latent that the full sensor set would yield, and a conditional flow samples the remaining residual. The closed-form fusion achieves lower reconstruction error than transformer fusion in every tested input condition with about 40× fewer fusion parameters, and predicting before sampling reduces radar-to-elevation CRPS by 36\% relative to a single flow at matched backbone and training budget. Trained on 37.5\% of the TerraMesh training split, EAGLE achieves significantly lower mean absolute error than COP-GEN and TerraMind on every radar, optical and elevation target across 3,037 MajorTOM validation scenes, and adding a sensor never significantly degrades its predictions. With its core frozen, the same representation supports missing-band reconstruction, hyperspectral adaptation, cross-sensor change detection and scene question answering. Code will be released.

Earth observation sensors measure different properties of a scene, but their availability varies across locations and dates. Reconstructing missing observations requires combining the available evidence while accounting for information that remains uncertain. We introduce Evidence-Aware Generative LatEnts (EAGLE) for generation and scene understanding from incomplete observations. EAGLE models encoded optical, SAR and elevation features as noisy linear views of a common latent, so any subset of sensors can be fused in closed form through a linear-Gaussian posterior. From this posterior, a deterministic predictor estimates the latent that the full sensor set would yield, and a conditional flow samples the remaining residual. Modality decoders reconstruct missing observations, while task-specific heads support scene understanding. The fusion model achieves lower reconstruction error than the tested transformer fusion with approximately 56 times fewer fusion parameters. Across four translation tasks, our deterministic predictor followed by the flow network reduces continuous ranked probability score (CRPS) by 6--37\% compared with a flow
network without a preceding deterministic predictor. Trained on TerraMesh, EAGLE achieves lower mean absolute error (MAE) than COP-GEN and TerraMind on every optical, radar and elevation target in our main comparison across 3,037 MajorTOM scenes, with scores averaged over 20 generated samples.The representation also supports missing-band reconstruction, hyperspectral adaptation, cross-sensor change detection and scene question answering, demonstrating its broader utility across generation and scene understanding\footnote{Preprint}.

\end{abstract}

%% file: sec/1_intro.tex
\section{Introduction}
\label{sec:intro}

Remote sensing (RS) data is heterogeneous and incomplete, with availability varying by location and date. Different sensing modalities often observe the same scene on different days, providing complementary evidence about its properties~\cite{astruc2025anysat,Fallah_2026_CVPR,9691348,TANG2026115120}. Together, these observations constrain what the unobserved modalities would reveal, providing a basis for cross-modal generation~\cite{10744588,khanna2024diffusionsat,jakubik2026terramind,espinosa2026copgen}. Such generation can support continuous environmental monitoring through gaps in sensor coverage and allow models developed for one modality to be used when only another is available.~\cite{jakubik2026terramind,yu2026metaearthmmunifiedmultimodalremote}.

\begin{figure}[t]
    \centering
    \includegraphics[width=1\linewidth]{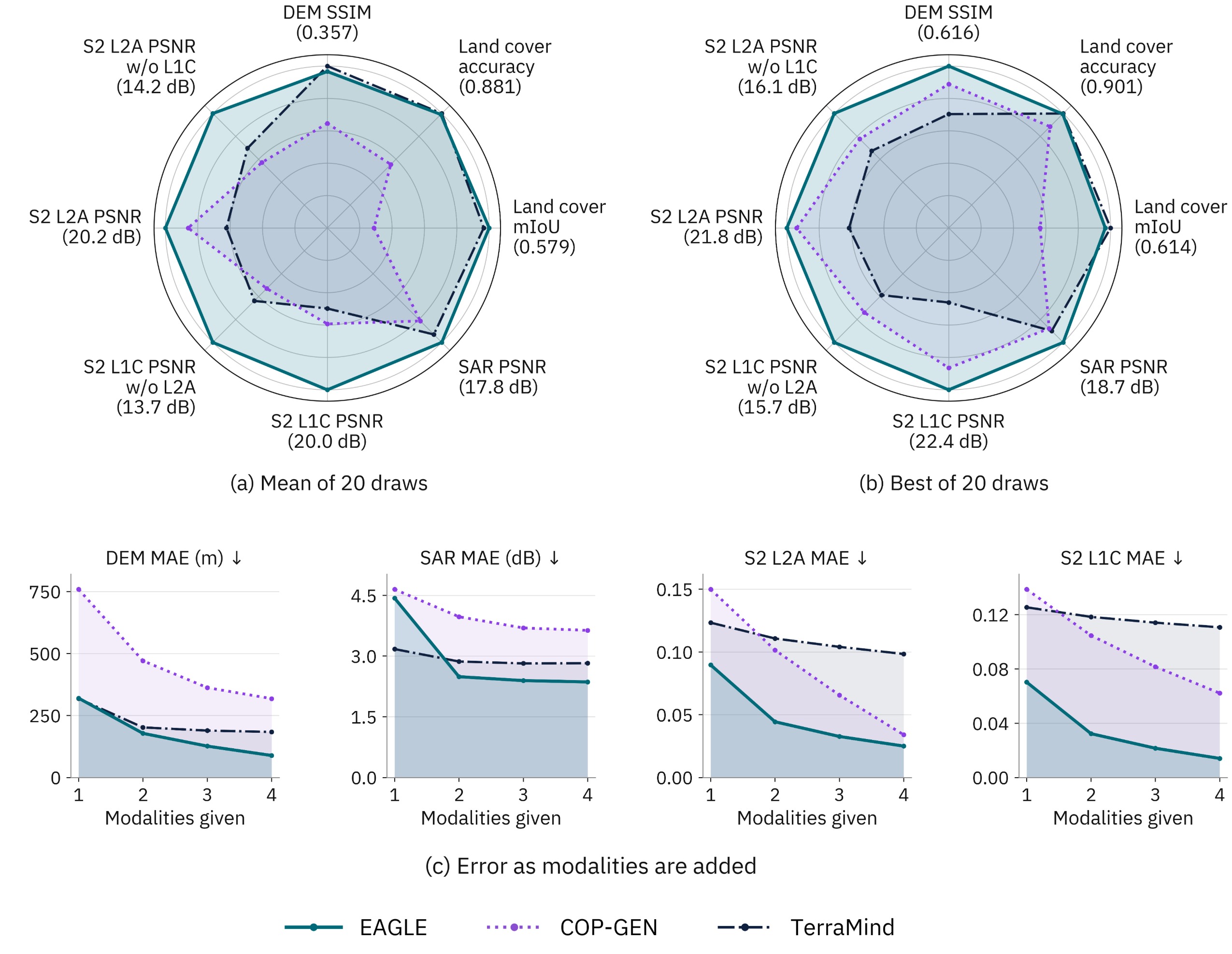}
\vskip-10pt\caption{\textbf{EAGLE leads across targets and benefits from additional inputs.}
\textbf{(a,b)}~EAGLE leads on six of eight axes for both mean
and reference-selected best scores over 20 draws per scene
on 3,037 MajorTOM scenes.
Axes are normalized to the best model; parentheses give
EAGLE's scores.
\textbf{(c)}~From one input to four, MAE falls by 47--80\%
for EAGLE and 11--42\% for TerraMind.
Points average evaluated input combinations on 500 scenes,
with eight draws per scene and combination.}
\label{fig:teaser}
\end{figure}

\begin{figure*}[t!]
    \centering
    \includegraphics[width=1\linewidth]{Figs/OA.jpg}
\vskip-8pt\caption{\textbf{Overview of EAGLE.}
Stage A learns band-aware sensor features and combines them
through closed-form linear-Gaussian fusion, appending coordinates
to form the scene latent.
With Stage A frozen, Stage B first refines the subset estimate
towards the full-evidence state through deterministic prediction,
conditioned on coordinates when available.
A conditional flow samples feature residuals and missing
coordinates on the sphere.
Modality decoders and task-specific heads use the completed
state for generation and scene understanding.}
\label{fig:overview}
    \label{fig:oa}
\end{figure*}

Early cross-modal methods learned mappings between individual sensor pairs, generally requiring separate models for different translation tasks~\cite{YANG2022108208,ijcai2022p229,8328024,Wijenayake_2026}. More recent foundation models accommodate heterogeneous bands, resolutions and sensor types within shared representations~\cite{NEURIPS2023_11822e84,pmlr-v267-tseng25a,xiong2024dofa,astruc2025anysat}. Unified models such as TerraMind and COP-GEN generate requested modalities from different subsets of the available observations~\cite{jakubik2026terramind,espinosa2026copgen}. Stochastic generation allows several plausible reconstructions from the same inputs, while shared scene representations also connect generation with downstream scene understanding~\cite{NEURIPS2023_b6446566,Jia_2025_ICCV,chen2026any2anyunifiedarbitrarymodality,yu2026metaearthmmunifiedmultimodalremote,changejepa}. Together, these capabilities place the shared scene representation
at the interface between the evidence supplied by available sensors
and the information left unresolved when sensors are missing.

% However, a shared representation does not, by itself, express how strongly each observation constrains the scene. In these models, the contribution of individual sensors remains largely implicit in learned fusion and conditioning. This limitation becomes important when the observations provide partial, unequal, or conflicting evidence. A useful scene representation should therefore make these contributions explicit while accounting for the information that the available observations leave unresolved. Beyond fusing the available evidence, cross-modal generation must also estimate the scene information that missing or partial observations would contributed.

However, a shared representation does not, by itself, provide an explicit probabilistic account of how each sensor constrains the scene. Probabilistic latent-variable models offer a basis for expressing these relationships. Recent work suggests that representations from visual encoders and generative models can be treated as noisy linear views of a common latent~\cite{bach2005probabilistic,tasker2026universal}. These findings motivate investigating whether a comparable structure can be learned across observations from different physical sensing modalities. For Earth observation, we ask the question whether this structure remains useful across optical imagery provided by Sentinel-2 (S2), synthetic aperture radar (SAR) provided by Sentinel-1 (S1) and elevation data provided by Digital Elevation Model (DEM), retaining the scene information needed for inference when the available sensors change. An explicit model of the supplied observations would provide a basis for combining their evidence, generating missing modalities also requires accounting for the scene information that this evidence leaves unresolved. To address this question, we introduce
\textbf{E}vidence-\textbf{A}ware \textbf{G}enerative
\textbf{L}at\textbf{E}nts (EAGLE), a probabilistic framework
for scene inference and completion from changing subsets of
Earth observation sensors.

Our contributions are twofold:
\textbf{(1)} We introduce EAGLE, which, to the best of our knowledge is the first
framework to learn a closed-form scene posterior for arbitrary
supported subsets of optical, SAR and elevation observations
and couple it with deterministic full-evidence latent prediction
and conditional residual generation.
\textbf{(2)} We demonstrate the practical value and generality
of this formulation. Trained on TerraMesh EAGLE achieves lower MAE, averaged over samples,
than COP-GEN and TerraMind for every SAR, optical and elevation
target in our main comparison across 3,037 scenes
(\cref{fig:teaser}). EAGLE also supports missing-band
reconstruction, hyperspectral sensor adaptation
and downstream scene understanding.

% To the best of our knowledge, EAGLE is the first framework to couple an explicit arbitrary-subset sensor posterior with deterministic full-evidence latent prediction and conditional residual generation across optical, SAR and elevation observations.

% Our contributions are:
% \textbf{(1)} We introduce EAGLE, a framework to couple closed-form inference over arbitrary optical--SAR--elevation subsets with deterministic full-evidence prediction and conditional residual generation.
% \textbf{(2)} We demonstrate the practical value and generality
% of this formulation. Trained on 37.5\% of the TerraMesh
% training split, EAGLE achieves lower MAE, averaged over
% samples, than COP-GEN and TerraMind for every SAR, optical
% and elevation target in our main comparison across 3,037
% scenes. With its pretrained core fixed, EAGLE also supports
% missing-band reconstruction, adaptation to a 368-band
% hyperspectral sensor and downstream scene understanding.

%% file: sec/2_relatedwork.tex
    \section{Related Work}

\paragraph{Multimodal Generative Representations}
Learning shared representations across modalities provides a
foundation for cross-modal inference and generation.
Multimodal VAEs~\cite{wu2018multimodal} use product-of-experts
inference to combine available modalities, learning from
randomly sampled modality subsets.
Unified Latents~\cite{heek2026unifiedlatentsultrain} learns
representations regularized by a diffusion prior, supporting
both reconstruction and generation.
MUNI~\cite{yeo2026muni} extends shared latent modeling to
any-to-any generation conditioned on arbitrary modality subsets.
Together with UNE's hypothesis of noisy linear views of a
shared latent~\cite{tasker2026universal}, these approaches
motivate our learned sensor representation: available
observations contribute explicitly to a closed-form posterior,
which provides the basis for generating missing modalities.

\myparagraph{Spectral-Aware Geospatial Models}
Earth observation data contain heterogeneous measurements
with varying bands, wavelengths and spatial resolutions.
DOFA~\cite{xiong2024dofa} uses a wavelength-aware dynamic
transformer to accommodate different remote sensing modalities.
SatMAE~\cite{cong2023satmae} uses spectral positional encodings
and band-grouped tokenisation, while SpectralGPT~\cite{Hong_2024}
learns spectral representations through generative pretraining.
SpectralEarth~\cite{Braham_2025} introduces spectral adapters
that allow vision backbones to process hyperspectral imagery.
These approaches demonstrate the value of accounting for
spectral characteristics when learning shared representations \cite{10.1007/978-3-031-73039-9_10}.
They motivate our band-aware encoding, which incorporates
band identity and resolution so that complete and partial
band sets can enter the same sensor fusion model.

\myparagraph{Geospatial Generative Models}
Recent generative models support the synthesis and
reconstruction of heterogeneous Earth observation data.
DiffusionSat~\cite{khanna2024diffusionsat} supports
metadata-conditioned generation, temporal prediction and
multispectral super-resolution.
TerraMind~\cite{jakubik2026terramind} combines pixel- and
token-level representations for generation and scene understanding,
while COP-GEN~\cite{espinosa2026copgen} models cross-modal
distributions through latent diffusion.
ProM3E~\cite{sastry2025prom3e} reconstructs missing modality
representations probabilistically in embedding space.
CorrDiff~\cite{mardani2024residualcorrectivediffusionmodeling}
combines mean prediction with generative residual correction
for atmospheric downscaling.
Any2Any~\cite{chen2026any2anyunifiedarbitrarymodality} uses
modality-specific VAEs and a shared diffusion transformer for
target-latent regression, followed by deterministic target-specific
adapters that correct latent mismatches.
MetaEarth-MM~\cite{yu2026metaearthmmunifiedmultimodalremote}
infers scene embeddings through joint attention over noisy
modality latents, with scene-consistency regularisation and
modality-routed generation.
EAGLE instead fits a linear-Gaussian observation model whose
product-of-experts (PoE) posterior combines any available subset of
supported sensors in closed form, weighting evidence by its
fitted precision. The posterior mean and covariance condition
a deterministic estimate of the full-evidence scene state,
a conditional flow then samples its remaining feature error.

%% file: Tables/loo.tex
% Review draft: combines reported main results with the supplied leave-one-out table.
% Needs booktabs, multirow, graphicx, colortbl, xcolor[table].
\definecolor{A2Eink}{HTML}{204E56}
\definecolor{A2Eshade}{HTML}{EDF4F4}
\providecommand{\aBest}[1]{\textcolor{A2Eink}{\textbf{#1}}}
% Main values and 95% intervals transcribed from the existing main-results table.
% Leave-one-out sample count confirmed by the author: 3,037 scenes.
\providecommand{\aScoreCI}[2]{#1\,{\fontsize{6}{7}\selectfont[#2]}}
\begin{table}[t]
\centering
\scriptsize
\setlength{\tabcolsep}{2pt}
\renewcommand{\arraystretch}{1.05}
\caption{\textbf{Main and leave-one-out generation.}
\textbf{(a)} 3,037 MajorTOM scenes;
\textbf{(b)} 500-scene subset.
Inputs: S2-L1C, S2-L2A, S1-RTC, DEM and coordinates,
excluding the target and, in (b), the listed input.
S2-RGB is excluded. Because S2 L1C and L2A derive from the same observations, withholding one retains overlapping optical information. We therefore also evaluate generation with both products withheld to assess cross-sensor inference, with results in the supplementary material.
Baselines also receive LULC unless removed
($^{\ddagger}$); COP-GEN receives cloud masks and dates.
Scores use 20 draws; best-of-20 uses ground truth.
EAGLE uses one location estimate in both columns.
Brackets: 95\% geographic block-bootstrap intervals.
MAE units: DEM, m; S1, dB; S2, reflectance.
Location: great-circle distance, km.
TerraM.: TerraMind-B.
\textbf{Bold}: best; \underline{underline}: second-best.}
\begin{tabular*}{\columnwidth}{@{\extracolsep{\fill}}l
    >{\columncolor{A2Eshade}}rrr@{}}
\toprule
\multicolumn{4}{@{}l}{\textbf{(a) Main comparison: mean error $\downarrow$}} \\
\textbf{Target} & \textbf{EAGLE} & \textbf{COP-GEN} & \textbf{TerraMind-B} \\
\midrule
DEM & \aScoreCI{\aBest{88.2}}{84.9, 91.7}
    & \aScoreCI{318.2}{307.4, 328.8}
    & \aScoreCI{\underline{198.3}}{174.5, 223.3} \\
S1 RTC & \aScoreCI{\aBest{2.36}}{2.33, 2.38}
    & \aScoreCI{3.65}{3.61, 3.70}
    & \aScoreCI{\underline{2.84}}{2.79, 2.89} \\
S2 L1C & \aScoreCI{\aBest{0.0140}}{0.0136, 0.0144}
    & \aScoreCI{\underline{0.0627}}{0.0619, 0.0635}
    & \aScoreCI{0.1083}{0.1042, 0.1126} \\
S2 L2A & \aScoreCI{\aBest{0.0243}}{0.0233, 0.0254}
    & \aScoreCI{\underline{0.0376}}{0.0366, 0.0385}
    & \aScoreCI{0.0957}{0.0906, 0.1011} \\
Location & \aScoreCI{\underline{866}}{773, 957}
    & \aScoreCI{1188}{1116, 1265}
    & \aScoreCI{\textbf{623}}{566, 684} \\
\bottomrule
\end{tabular*}\par

\makebox[\columnwidth][l]{\textbf{(b) Leave-one-out generation}}\par
\resizebox{\columnwidth}{!}{%
\begin{tabular}{@{}ll >{\columncolor{A2Eshade}}rrr >{\columncolor{A2Eshade}}rrr@{}}
% \toprule
& & \multicolumn{3}{c}{\textbf{Mean of 20}} & \multicolumn{3}{c}{\textbf{Best of 20}} \\
\cmidrule(lr){3-5}\cmidrule(l){6-8}
\textbf{Target} & \textbf{Removed} & \textbf{EAGLE} & \textbf{COP-GEN} & \textbf{TerraM.} & \textbf{EAGLE} & \textbf{COP-GEN} & \textbf{TerraM.} \\
\midrule
\multirow{5}{*}{\shortstack[l]{DEM\\{\scriptsize MAE (m) $\downarrow$}}} & w/o LatLon & \underline{223.4} & 330.1 & \textbf{185.0} & \aBest{37.1} & \underline{41.0} & 171.3 \\
 & w/o LULC$^{\ddagger}$ & \aBest{90.5} & 320.1 & \underline{183.9} & \aBest{17.8} & \underline{39.7} & 171.0 \\
 & w/o S1 & \aBest{98.1} & 341.7 & \underline{184.1} & \aBest{20.7} & \underline{55.4} & 171.1 \\
 & w/o L1C & \aBest{92.5} & 450.6 & \underline{192.5} & \aBest{18.6} & \underline{53.9} & 174.9 \\
 & w/o L2A & \aBest{98.5} & 331.8 & \underline{193.8} & \aBest{21.4} & \underline{42.1} & 177.6 \\
\midrule
\multirow{5}{*}{\shortstack[l]{LULC\\{\scriptsize top-1 $\uparrow$}}} & w/o LatLon & \underline{0.862} & 0.493 & \textbf{0.891} & \aBest{0.911} & 0.803 & \underline{0.907} \\
 & w/o DEM & \underline{0.878} & 0.494 & \textbf{0.890} & \underline{0.905} & 0.805 & \textbf{0.907} \\
 & w/o S1 & \underline{0.876} & 0.486 & \textbf{0.888} & \aBest{0.907} & 0.799 & \underline{0.904} \\
 & w/o L1C & \underline{0.879} & 0.515 & \textbf{0.892} & \underline{0.906} & 0.800 & \textbf{0.908} \\
 & w/o L2A & \underline{0.871} & 0.510 & \textbf{0.890} & \underline{0.894} & 0.802 & \textbf{0.906} \\
\midrule
\multirow{5}{*}{\shortstack[l]{S1\\{\scriptsize MAE (dB) $\downarrow$}}} & w/o LatLon & \aBest{2.43} & 3.60 & \underline{2.83} & \aBest{2.10} & \underline{2.71} & 2.72 \\
 & w/o LULC$^{\ddagger}$ & \aBest{2.36} & 3.67 & \underline{2.82} & \aBest{2.09} & 2.75 & \underline{2.72} \\
 & w/o DEM & \aBest{2.41} & 3.82 & \underline{2.82} & \aBest{2.10} & 2.81 & \underline{2.71} \\
 & w/o L1C & \aBest{2.36} & 3.59 & \underline{2.82} & \aBest{2.08} & \underline{2.71} & \underline{2.71} \\
 & w/o L2A & \aBest{2.38} & 3.58 & \underline{2.81} & \aBest{2.10} & \underline{2.68} & 2.71 \\
\midrule
\multirow{5}{*}{\shortstack[l]{S2 L1C\\{\scriptsize MAE $\downarrow$}}} & w/o LatLon & \aBest{0.0151} & \underline{0.0636} & 0.1111 & \aBest{0.0120} & \underline{0.0179} & 0.1068 \\
 & w/o LULC$^{\ddagger}$ & \aBest{0.0141} & \underline{0.0620} & 0.1106 & \aBest{0.0116} & \underline{0.0178} & 0.1064 \\
 & w/o DEM & \aBest{0.0148} & \underline{0.0647} & 0.1107 & \aBest{0.0120} & \underline{0.0179} & 0.1064 \\
 & w/o S1 & \aBest{0.0145} & \underline{0.0625} & 0.1108 & \aBest{0.0117} & \underline{0.0173} & 0.1065 \\
 & w/o L2A & \aBest{0.0421} & 0.1314 & \underline{0.1238} & \aBest{0.0329} & \underline{0.0563} & 0.1150 \\
\midrule
\multirow{5}{*}{\shortstack[l]{S2 L2A\\{\scriptsize MAE $\downarrow$}}} & w/o LatLon & \aBest{0.0257} & \underline{0.0361} & 0.0987 & \aBest{0.0210} & \underline{0.0220} & 0.0938 \\
 & w/o LULC$^{\ddagger}$ & \aBest{0.0251} & \underline{0.0345} & 0.0984 & \aBest{0.0210} & \underline{0.0220} & 0.0935 \\
 & w/o DEM & \aBest{0.0257} & \underline{0.0363} & 0.0987 & \aBest{0.0217} & \underline{0.0221} & 0.0939 \\
 & w/o S1 & \aBest{0.0247} & \underline{0.0350} & 0.0988 & \aBest{0.0203} & \underline{0.0220} & 0.0939 \\
 & w/o L1C & \aBest{0.0547} & 0.1496 & \underline{0.1192} & \aBest{0.0436} & \underline{0.0668} & 0.1098 \\
\midrule
\multirow{5}{*}{\shortstack[l]{LatLon\\{\scriptsize km $\downarrow$}}} & w/o LULC$^{\ddagger}$ & \underline{703} & 1317 & \textbf{644} & 703 & \underline{236} & \textbf{188} \\
 & w/o DEM & \underline{970} & 1967 & \textbf{683} & 970 & \underline{343} & \textbf{189} \\
 & w/o S1 & \underline{1131} & 1545 & \textbf{643} & 1131 & \underline{323} & \textbf{199} \\
 & w/o L1C & \underline{821} & 1395 & \textbf{673} & 821 & \underline{250} & \textbf{218} \\
 & w/o L2A & 2103 & \underline{1462} & \textbf{655} & 2103 & \underline{337} & \textbf{216} \\
\bottomrule
\end{tabular}}
\label{tab:loo}
\label{tab:generation_combined}
\end{table}

%% file: sec/3_methodology.tex
\section{Methodology}
\label{sec:method}
We train EAGLE in two stages as shown in \cref{fig:overview}. Stage A fuses the encoded sensor features using a linear-Gaussian model and appends the RS scene coordinates to form the scene latent (\cref{sec:stage_a}). Freezing Stage A, Stage B uses this fusion model to combine
the available sensor features into a subset posterior.
A deterministic predictor then uses this posterior to estimate
the latent that the full sensor set would yield.
A flow model samples a residual correction to this estimate
(\cref{sec:stage_b}). Modality-aware decoders reconstruct the requested outputs from the resulting latent, and task-specific heads
use it for scene understanding (\cref{sec:downstream}).

\subsection{Stage A: learning a shared scene latent}
\label{sec:stage_a}

\textbf{Band Aware Encoder} To learn spatial features jointly from S2-L2A, S2-RGB, S2-L1C, S1-RTC, S1-GRD and DEM data, we use a shared encoder $E_{\mathrm{img}}$. Since the number and meaning of the bands vary across these modalities, we embed each band separately, following a band-wise approach similar to~\cite{xiong2024dofa}. Each embedding includes the band's identity and native resolution. We combine the available band embeddings at each spatial location into a single token and process the resulting tokens with a pretrained DINOv2 ViT-B~\cite{dinov2}. For modality $m_i$, the encoder output is a feature map $Y_i \in \mathbb{R}^{K \times 32 \times 32}$. Its feature vector $y_{i,j} \in \mathbb{R}^K$ serves as the observation from modality $i$ at a spatial location $j$, which we define as a spatial cell, in the fusion model.

\begin{figure*}[t]
    \centering
    \includegraphics[width=1\linewidth]{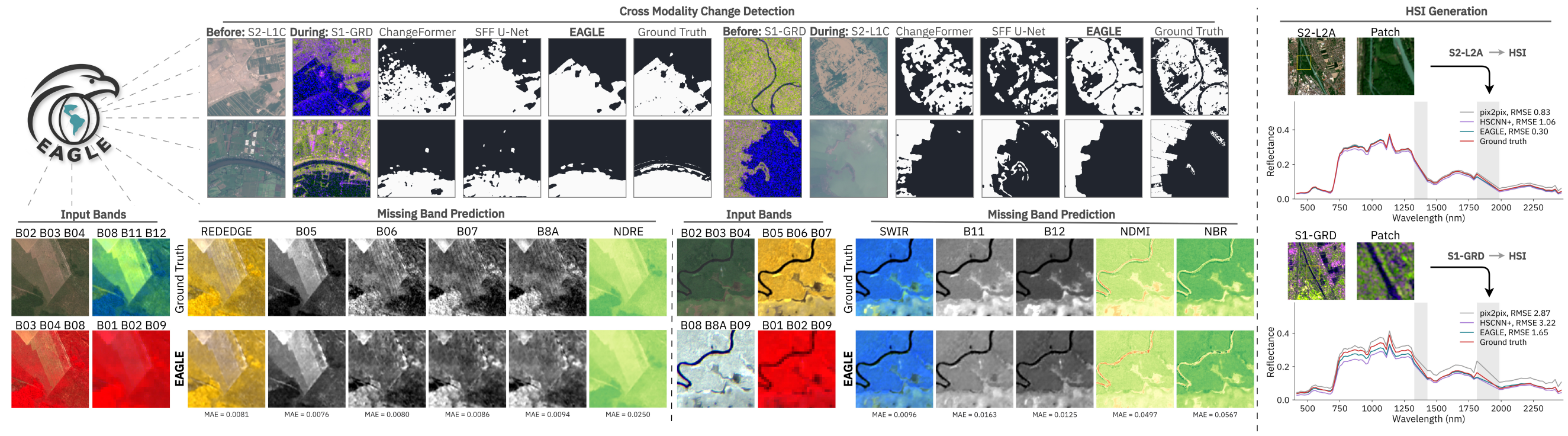}
\vskip-10pt\caption{\textbf{Generation and scene understanding.}
\textbf{Upper left:} cross-sensor flood change detection;
white marks newly flooded pixels.
\textbf{Lower left:} withheld S2 band reconstruction
and derived spectral indices.
\textbf{Right:} HSI spectra predicted from S2 and S1,
compared with ground truth and baselines.}
    \label{fig:fullresult}
\end{figure*}

\myparagraph{Linear Gaussian Sensor Fusion}
We hypothesize that the encoded features of different sensors
are noisy linear views of a shared sensor latent
$h \in \mathbb{R}^{K \times 32 \times 32}$, following the latent
variable models of~\cite{bach2005probabilistic,tasker2026universal}.
Let $h_j \in \mathbb{R}^K$ denote its value at spatial cell $j$. 
For modality $i$, we write $y_{i,j} = A_i h_j + b_i + \varepsilon_{i,j},$
where $A_i$ and $b_i$ map $h_j$ to the feature space of modality $i$. The noise $\varepsilon_{i,j} \sim \mathcal{N}(0,R_i)$ has diagonal covariance $R_i$, fitted per sensor variant and shared across cells. 

For an input subset $S \subseteq F$, let $S_j$ index valid
observations at cell $j$, with at most one variant per
optical, SAR or elevation group, when both S2-L1C and
S2-L2A are available, they are encoded together as a
single optical variant. Assuming conditional independence across sensor groups given $h_j$, the posterior is proportional to the product of the prior $\mathcal{N}(0,I)$ and the Gaussian likelihoods for $S_j$~\cite{hinton2002training,wu2018multimodal}. Its mean $\mu_{S,j}$ gives the most probable value of $h_j$,
while its covariance $\Sigma_{S,j}$ describes uncertainty
under this observation model. Both are obtained in closed form,
\setlength{\belowdisplayskip}{0pt} \setlength{\belowdisplayshortskip}{0pt}
\setlength{\abovedisplayskip}{0pt} \setlength{\abovedisplayshortskip}{0pt}
\begin{equation}
\begin{aligned}
  \Sigma_{S,j}^{-1}
  &= I + \sum_{i \in S_j} A_i^{\top} R_i^{-1} A_i, \\
  \mu_{S,j}
  &= \Sigma_{S,j} \sum_{i \in S_j}
     A_i^{\top} R_i^{-1}(y_{i,j}-b_i),
\end{aligned}
\label{eq:posterior}
\end{equation}
where $\Sigma_{S,j}^{-1}$ is the precision. With no valid observations, the posterior equals the prior. For fixed observation variants, adding a sensor group
adds positive semidefinite precision and cannot increase
posterior variance.

Let $x_F$ denotes the observations from the full sensor set $F$.
Applying Eq.~\eqref{eq:posterior} at each cell gives the full-evidence
posterior $q(h \mid x_F) = \mathcal{N}(\mu_F,\Sigma_F)$ as shown in
Fig.~\ref{fig:overview}, where the posterior means $\mu_{F,j}$ across cells form the fused
sensor feature map $\mu_F \in \mathbb{R}^{K \times 32 \times 32}$,
and the cell covariances $\Sigma_{F,j}$ form the blocks of the
block-diagonal covariance $\Sigma_F$.
Location is a property of the scene, so we introduce it after sensor
fusion. We represent the scene center at latitude $\phi$ and longitude
$\lambda$ by
$\ell=\sqrt{3}\,(\cos\phi\cos\lambda,\,
\cos\phi\sin\lambda,\,\sin\phi)$.
Repeating $\ell$ across the spatial grid and appending it to
$\mu_F$ gives the full-evidence scene latent
$H=[\mu_F;\ell]\in\mathbb{R}^{(K+3)\times32\times32}$.

\textbf{Decoders }Optical reflectance, radar backscatter and elevation are distinct
physical quantities, so we use separate optical, SAR and DEM decoders\cite{10.1007/978-3-031-73390-1_24,Qu_Yang_Dong_Yang_2024}.
Each decoder uses Swin-style transformer blocks~\cite{Liu_2021_ICCV}
followed by pixel-shuffle upsampling.
For a requested modality $i$, the corresponding decoder maps the
shared scene latent $H$ to the reconstruction $\hat{m}_i = D_i(H)$.
The optical decoder shares a backbone across S2 L1C, L2A and RGB,
with a separate output head for each.
We compute NDVI from the red and near-infrared bands of the
reconstructed L2A product.
The SAR decoder shares a backbone across S1 GRD and RTC, with the
requested product conditioning the backbone and selecting the
output head.
A separate geolocation head estimates the scene center from the
scene features using geocell classification and retrieval\cite{Haas_2024_CVPR,Berton_2024_CVPR}.

We train Stage A on the full sensor set $F$ and a sampled sensor
subset from the same scene. For either input set $S$,
the variational loss is,
\begin{equation}
  \mathcal{L}_{\mathrm{var}}(S)
  = \mathcal{L}_{dec}
  + \beta\,D_{\mathrm{KL}}\!\left(
    q(h\mid x_S)\,\|\,\mathcal{N}(0,I)\right).
  \label{eq:stage_a}
\end{equation}
The decoder loss $\mathcal{L}_{\mathrm{dec}}$ combines
mean absolute reconstruction error with modality-specific
spectral and spatial penalties, and cross-entropy for
land-cover prediction. We evaluate it using the posterior
mean and one posterior sample, then average the two losses.
The KL term, $D_{\mathrm{KL}}$  regularizes the posterior, $q(h\mid x_S)$, towards the standard Gaussian prior, $\mathcal{N}(0,I)$. Full loss definitions and weights are given in \cref{sup:decoder_loss}.

We first learn the representation with $A_i=I$ and
encoder-predicted precisions per channel and cell.
We then freeze the encoder and fit $A_i$, $b_i$ and
cell-shared diagonal $R_i$ per variant by maximising
the marginal likelihood of jointly observed features,
integrating out $h\sim\mathcal{N}(0,I)$.
The six variants contain $6(K^2+2K)=5{,}040$ fitted
parameters at $K=28$. These define subsequent fusion,
including Stage B and inference. Finally, we adapt the
decoders with the encoder and observation parameters fixed.

\myparagraph{Land-cover segmentation}
During Stage A, land-cover labels supervise a decoder that predicts a class at each pixel. At inference, we apply this frozen decoder to the scene state inferred from the available sensors.

\subsection{Stage B: learning the correction to the full-evidence state}
\label{sec:stage_b}

With Stage A frozen, we form the full-evidence target
$H=[\mu_F;\ell]$ for each training scene and obtain the subset
posterior from a sampled sensor set $S$.
Our key idea is to estimate the scene from the available
evidence before sampling its remaining feature error.
The deterministic predictor learns a correction to the subset
posterior mean, and a conditional flow samples a further
correction to its feature estimate.
The Stage A decoders map the completed state $\hat{H}$
to the requested outputs.

\myparagraph{B1: Deterministic prediction}
A SiT-L/2 transformer backbone~\cite{ma2024sit} refines the subset
posterior mean $\mu_S$ using evidence from surrounding cells.
It takes $\mu_S$ and the channel-wise log variances from
$\Sigma_S$ as inputs, with scene coordinates as additional
conditioning when available. The transformer predicts a
feature correction $\Delta h_\theta$ and, when coordinates
are withheld, estimates the scene coordinates jointly with
this correction. The resulting deterministic state is
$\hat{H}_{\mathrm{det}} = [\hat{\mu}_{F_{det}};\,\hat{\ell}_{\mathrm{det}}]
=[\mu_S+\Delta h_\theta;\,\hat{\ell}_{\mathrm{det}}]$,
where $\hat{\ell}_{\mathrm{det}}$ contains the supplied
coordinates when available and the predicted coordinates
otherwise.

We train B1 by regression~\cite{mardani2024residualcorrectivediffusionmodeling} against
the full-evidence state $H=[\mu_F;\,\ell]$ with the loss,
\begin{equation}
  \mathcal{L}_{B1}
  = \lVert \hat{H}_{\mathrm{det}}-H \rVert^2
  + 0.1\,\mathcal{L}_{\mathrm{dec}}.
  \label{eq:b1}
\end{equation}
The decoder loss $\mathcal{L}_{\mathrm{dec}}$ compares
$D_i(\hat{H}_{\mathrm{det}})$ with ground truth for all modalities.

\myparagraph{B2: Stochastic prediction}
We model uncertainty after deterministic prediction by jointly
sampling feature residuals and withheld scene coordinates.
During training, the feature residual is
$r_\mu=\mu_F-\hat{\mu}_{F_{\mathrm{det}}}$.
We train a single SiT-B/2 transformer by flow
matching~\cite{liu2022flow,lipman2023flow}.
For sensor features, the path
$z_t^\mu=(1-t)z_0^\mu+t r_\mu$
interpolates linearly from Gaussian noise
$z_0^\mu\sim\mathcal{N}(0,I)$ to the feature residual.
For withheld coordinates, $z_t^\ell$ follows a great circle
from $z_0^\ell$, sampled uniformly on
$\sqrt{3}\mathbb{S}^2$, to the true scene coordinates
$\ell$~\cite{chen2024flow,Dufour_2025_CVPR}.
The deterministic coordinate estimate conditions this flow
through $\hat H_{\mathrm{det}}$.
When coordinates are supplied, the coordinate state remains
fixed at $\ell$.

At time $t$, the transformer uses the joint state
$[z_t^\mu;\,z_t^\ell]$, the available sensor evidence and
$\hat H_{\mathrm{det}}$ to predict both velocities.
The training loss is
\begin{equation}
  \mathcal{L}_{B2}
  = \mathbb{E}_{t,z_0^\mu,z_0^\ell}
    \big\|\hat v_t^\mu-(r_\mu-z_0^\mu)\big\|^2
  + \lambda_{\mathrm{loc}}\,\mathcal{L}_{\mathrm{sph}}.
  \label{eq:flow}
\end{equation}
The first term matches the feature velocity along the linear
path. The spherical loss $\mathcal{L}_{\mathrm{sph}}$
penalises the squared difference between the predicted tangent
velocity and the target great-circle velocity, averaged over
scenes with withheld coordinates.

\input{Tables/ablation2}

\subsection{Inference}
\label{sec:inference}

Given observations $x_S$, we fuse the available sensor features
and predict $\hat H_{\mathrm{det}}$ from the subset posterior
(Fig.~\ref{fig:overview}).
We integrate the conditional flow with 64 Heun
steps~\cite{karras2022elucidatingdesignspacediffusionbased}
to obtain the feature residual $\hat r_\mu$ and, when
coordinates are withheld, the spherical endpoint $z_1^\ell$.
The completed state is
$\hat H=[\hat{\mu}_{F_{\mathrm{det}}}+\hat r_\mu;\,\hat\ell]$,
where $\hat\ell=\ell$ when coordinates are supplied and
$\hat\ell=z_1^\ell$ otherwise.
The frozen modality decoder reconstructs
$\hat m_i=D_i(\hat H)$ using the coordinate component
$\hat\ell$ of this state.
Downstream heads also use $\hat H$, including the geocell
classification and retrieval head that provides the reported
geolocation estimate.
Different noise draws yield alternative scene completions
and corresponding modality reconstructions.

%% file: Tables/ablation2.tex
% Requires booktabs, pifont, and xcolor[table] (or xcolor + colortbl).
% Replace both original table inputs with this file.
% The original labels are retained as aliases for the combined table.
\definecolor{A2Eshade}{HTML}{EDF4F4}
\definecolor{A2Eink}{HTML}{204E56}
\definecolor{A2Emuted}{HTML}{B0B0B0}
\providecommand{\aYes}{\ding{51}}
\providecommand{\aNo}{\textcolor{A2Emuted}{\ding{55}}}
\providecommand{\aBest}[1]{\textcolor{A2Eink}{\textbf{#1}}}
\providecommand{\aSecond}[1]{\underline{#1}}
\providecommand{\aOurs}{\textcolor{A2Eink}{\textbf{EAGLE}}}
\providecommand{\aPanel}[1]{\textcolor{A2Eink}{\textbf{#1}}}
\providecommand{\aHead}[1]{{\scriptsize\bfseries #1}}

\begin{table}[t]
\centering
\scriptsize
\setlength{\tabcolsep}{1pt}
\renewcommand{\arraystretch}{1.05}
\renewcommand{\aHead}[1]{{\scriptsize\bfseries #1}}
\caption{\textbf{Ablation study.}
Units: DEM, m; S1, dB; S2-L2A, reflectance.
\textbf{(a)} Params: fusion parameters;
Prec.: learned encoder precision; Subset: subset training.
S2$\to$S2 also averages S2+S1 inputs;
S1/DEM: either alone; S1+DEM is held out in training.
\textbf{(b,c)} Coordinates supplied.
DEM: 4,423 scenes, 32 draws;
S1 and S2: 600 scenes, 16 draws.
Stage A samples its posterior; B1-only CRPS equals MAE.
$^{\S}$Full-latent flow target.
\textbf{(d)} Stage A with known coordinates, 4,423 scenes.
\textbf{(e)} 3,037 scenes; flow uses 20 draws,
geocell reports geometric-median estimate.
\textbf{Bold}: best; \underline{underline}: second-best
within each panel.}
% Stage A: fusion
\begin{tabular*}{\columnwidth}{@{\extracolsep{\fill}}lcccrrr@{}}
\toprule
\multicolumn{7}{@{}l}{\aPanel{(a) Fusion: S2-L2A MAE $\downarrow$}} \\
\aHead{Method} & \aHead{Params} & \aHead{Prec.} & \aHead{Subset}
& \aHead{S2$\to$S2} & \aHead{S1/DEM$\to$S2} & \aHead{S1+DEM$\to$S2} \\
\midrule
Transformer & 282K & \textemdash & \aYes
& 0.0200 & 0.0807 & 0.0862 \\
PoE, uniform & 5.04K & \aNo & \aYes
& 0.0176 & \aSecond{0.0747} & \textbf{0.0751} \\
PoE, full only & 5.04K & \aYes & \aNo
& \aSecond{0.0158} & 0.0926 & 0.0864 \\
\rowcolor{A2Eshade}
\aOurs{} (PoE) & 5.04K & \aYes & \aYes
& \aBest{0.0151} & \aBest{0.0733} & \aSecond{0.0756}\\
\end{tabular*}\par

% Shared target columns for completion and location input
\begin{tabular*}{\columnwidth}{@{\extracolsep{\fill}}lccrrrr@{}}
\midrule
\multicolumn{7}{@{}l}{\aPanel{(b) Completion and latent width: CRPS $\downarrow$}} \\
\aHead{Method} & \aHead{B1} & \aHead{B2}
& \aHead{S1$\to$DEM} & \aHead{S2+S1$\to$DEM}
& \aHead{S1$\to$S2} & \aHead{S2$\to$S1} \\
\midrule
Stage A posterior & \aNo & \aNo
& 245.3 & 226.0 & 0.0599 & 2.17 \\
B1 only & \aYes & \aNo
& 72.3 & 64.4 & 0.0438 & 2.40 \\
Full-latent flow$^{\S}$ & \aYes & \aYes
& 57.2 & 51.5 & \textbf{0.0341} & \aSecond{1.80} \\
EAGLE, $K=36$ & \aYes & \aYes
& \aSecond{56.6} & \textbf{50.4} & \textbf{0.0341} & \textbf{1.79} \\
\rowcolor{A2Eshade}
\aOurs{}, $K=28$ & \aYes & \aYes
& \aBest{56.4} & \aSecond{50.8} & \aSecond{0.0342} & \aBest{1.79} \\
\midrule
\multicolumn{7}{@{}l}{\aPanel{(c) SiT-B/2, CRPS $\downarrow$}} \\
Single flow & \aNo & \aYes
& \aSecond{114.9} & \aSecond{107.7} & \aSecond{0.0467} & \aSecond{1.96} \\
\rowcolor{A2Eshade}
B1 + residual flow & \aYes & \aYes
& \aBest{73.7} & \aBest{67.8} & \aBest{0.0364} & \aBest{1.84} \\
\midrule
\multicolumn{7}{@{}l}{\aPanel{(d) Location input: MAE $\downarrow$}} \\
\multicolumn{3}{@{}l}{Fused as an expert}
& \aSecond{185.3} & \aSecond{165.1} & \aSecond{0.0699} & \aSecond{2.57} \\
\rowcolor{A2Eshade}
\multicolumn{3}{@{}l}{Appended (\aOurs{})}
& \aBest{173.7} & \aBest{160.5} & \aBest{0.0677} & \aBest{2.52} \\
\end{tabular*}\par

\begin{tabular*}{\columnwidth}{@{\extracolsep{\fill}}lcrr@{}}
\midrule
\multicolumn{4}{@{}l}{\aPanel{(e) Geolocation error (km) $\downarrow$}} \\
\aHead{Method} & \aHead{Geocell training scenes}
& \aHead{Mean} & \aHead{Median} \\
\midrule
Flow + decoder & \textemdash & \aSecond{1483} & \aSecond{796} \\
\rowcolor{A2Eshade}
\aOurs{} (geocell) & 1.39M & \aBest{866} & \aBest{141} \\
\bottomrule
\end{tabular*}

\label{tab:ablations_combined}
\label{tab:completion_ablation}
\label{tab:fusion_location_ablation}
\end{table}

%% file: sec/4_experiments.tex
\section{Experiments}
\label{sec:experiments}
\input{Tables/changedetection}

\subsection{Data}
% \textbf{Data.}
We train EAGLE on 3.35 million TerraMesh scenes, drawing from both MajorTOM and SSL4EO-S12~\cite{terramesh}. The data provide co-registered optical imagery, SAR, elevation, land cover and scene coordinates. Our main comparison (\cref{tab:loo}) uses one scene from each of the 3,037 eligible MajorTOM locations in the TerraMesh validation split, covering over six times as many locations as COP-GEN's released 495-scene test set~\cite{espinosa2026copgen}. We further evaluate 211 input configurations on a fixed subset of 500 scenes to assess performance as the available inputs change. Eligibility criteria, geographic coverage and evaluation on additional locations are detailed in \cref{sup:eval_sets,sup:additional_locations}.

% \textbf{Generation and evaluation.}
\subsection{Evaluation and Generation}
We compare EAGLE with the released checkpoints of COP-GEN~\cite{espinosa2026copgen} and TerraMind~\cite{jakubik2026terramind}, which support generation across optical, SAR and elevation modalities from partial observations. We follow COP-GEN's evaluation protocol, scoring image outputs on the central $192\times192$ pixels of each scene. The experiments assess generation across input subsets and robustness to predictive uncertainty. Detailed evaluation settings, metrics and statistical procedures are provided in \cref{sup:eval}.

\myparagraph{Hyperspectral reconstruction}
Similar to the other sensors, we add a band identity embedding to each hyperspectral channel and pass the result through the band-aware encoder $E_{\mathrm{img}}$. We fuse the encoded features using~\cref{eq:posterior}, and adapt the Stage B predictor and flow using LoRA~\cite{hu2021lora} and train a new decoder to reconstruct the hyperspectral bands from the scene latent.

% \textbf{Downstream tasks.}
\subsection{Downstream tasks}
\label{sec:downstream}

\myparagraph{Missing-band prediction}
For each evaluated S2 or HySpex band subset, we fit
subset-specific observation parameters $(A_i,b_i,R_i)$
on development data, keeping neural weights fixed.
The encoder processes the available bands, and the
completion model reconstructs the withheld bands.

\myparagraph{Change detection}
The sensors available before and after an event may differ. We estimate a scene state for each date so that one change decoder can compare S1, S2 or joint S1 and S2 observations in any pairing. We use the same noise draws at both dates and average the sampled states at each date to reduce variation due to sampling. A transformer compares the two sets of spatial tokens. An upsampling head combines its output with features from the frozen optical and SAR decoders to map changed areas.

\myparagraph{Scene question answering}
A transformer encodes the question and attends to spatial features from the completed scene latent $\hat{H}$ after stochastic refinement. A scoring head ranks the candidate answers. With EAGLE's pretrained weights frozen, we train one QA head across all seven non-empty subsets of S2 L2A, S1 RTC and DEM, using one randomly sampled scene state per training example. At inference, we average answer probabilities over sampled states.

\myparagraph{Evaluating Downstream tasks}
We evaluate missing-band reconstruction on S2 against COP-GEN, SatMAE and SpectralGPT+, and on the 368-band HySpex imagery in MDAS~\cite{mdas} against DHP\cite{Sidorov_2019_ICCV} (\cref{tab:missing_bands_published}). MDAS also supports hyperspectral transfer comparisons with pix2pix\cite{isola2017pix2pix} and HSCNN+\cite{Shi_2018_CVPR_Workshops} (\cref{tab:hsi}). For scene understanding, we evaluate flood change detection on SenForFlood~\cite{senforflood} against SFF U-Net and ChangeFormer~\cite{bandara2022changeformer} (\cref{tab:flood_cd}), and scene question answering on BigEarthNet.txt~\cite{herzog2026bigearthnettxt} against GeoChat\cite{Kuckreja_2024_CVPR}, EarthMind\cite{shu2025earthmindleveragingcrosssensordata}, GPT-5.2, Qwen3-VL\cite{bai2025qwen3vltechnicalreport} and fine-tuned RS-InternVL\cite{herzog2026bigearthnettxt}, together with a question-only control (\cref{tab:qa}).

% \textbf{Land-cover segmentation.}
% During Stage A, land-cover labels supervise a decoder that predicts a class at each pixel. At inference, we apply this frozen decoder to the scene state inferred from the available sensors.

%% file: Tables/changedetection.tex
\definecolor{A2Eshade}{HTML}{EDF4F4}
\definecolor{A2Eink}{HTML}{204E56}
\providecommand{\aBest}[1]{\textcolor{A2Eink}{\textbf{#1}}}
\providecommand{\aSecond}[1]{\underline{#1}}
\providecommand{\aOurs}{\textcolor{A2Eink}{\textbf{EAGLE}}}
\providecommand{\aHead}[1]{{\scriptsize\bfseries #1}}

\begin{table}[t]
\centering
\scriptsize
\setlength{\tabcolsep}{2pt}
\renewcommand{\arraystretch}{1.0}
\caption{\textbf{Cross-sensor flood change detection}
on SenForFlood~\cite{senforflood}.
Pixel-pooled F1 and IoU (\%) for newly flooded pixels.
SFF U-Net and ChangeFormer~\cite{bandara2022changeformer}
are trained for each setting; EAGLE uses one head on frozen features.
\textbf{Bold}: best; \underline{underline}: second-best.}
\begin{tabular*}{\columnwidth}{
    @{\extracolsep{\fill}}ll rr rr
    >{\columncolor{A2Eshade}}r
    >{\columncolor{A2Eshade}}r@{}
}
\toprule
& & \multicolumn{2}{c}{\aHead{SFF U-Net}}
& \multicolumn{2}{c}{\aHead{ChangeFormer}}
& \multicolumn{2}{>{\columncolor{A2Eshade}}c@{}}{
    \aHead{\aOurs{}}
} \\
\cmidrule(lr){3-4}
\cmidrule(lr){5-6}
\cmidrule(l){7-8}

\aHead{Before} & \aHead{During}
& \aHead{F1$\uparrow$} & \aHead{IoU$\uparrow$}
& \aHead{F1$\uparrow$} & \aHead{IoU$\uparrow$}
& \aHead{F1$\uparrow$} & \aHead{IoU$\uparrow$} \\
\midrule

S1 & S2
& \textbf{74.4}& \textbf{59.3}
& 71.8 & 56.0
& \aSecond{73.3} & \aSecond{57.9} \\

S2 & S1
& \aSecond{71.4} & \aSecond{55.5}
& 69.3 & 53.0
& \aBest{73.6} & \aBest{58.2} \\

S1\,+\,S2 & S2
& \textbf{75.6} & \textbf{60.7}
& 64.4 & 47.5
& \aSecond{71.1} & \aSecond{55.1} \\

S2 & S1\,+\,S2
& 72.2 & 56.5
& \aSecond{72.4} & \aSecond{56.7}
& \aBest{75.4} & \aBest{60.5} \\

S1 & S1\,+\,S2
& \aSecond{72.2} & \aSecond{56.4}
& 70.4 & 54.4
& \aBest{77.7} & \aBest{63.6} \\

S1\,+\,S2 & S1
& 71.4 & 55.5
& \aSecond{71.7} & \aSecond{55.9}
& \aBest{76.3} & \aBest{61.7} \\

\bottomrule
\end{tabular*}

\label{tab:flood_cd}
\end{table}

%% file: sec/5_results_and_discussion.tex
\section{Results and Discussions}
\input{Tables/scene_QA}

\subsection{Evaluating the Design of EAGLE}
\label{sec:results_ablations}

\input{Tables/hsi}

Stage A supports our hypothesis that learned sensor features
permit effective linear-Gaussian fusion
(\cref{tab:fusion_location_ablation}(a)).
EAGLE achieves lower reconstruction error than transformer fusion
in all three tested input conditions, using 5.04K rather than
282K fusion parameters, approximately 56 times fewer.
Although transformers can represent both linear and nonlinear
relations, their additional flexibility offers no improvement
in these comparisons.
Subset training lowers error in all three input conditions,
and learned precisions lower it in both conditions seen during training.
% Fitting full sensor mappings reduces held-out negative
% log-likelihood from 1.08 with identity mappings to 0.28;
% spatially varying noise gives no further improvement.
Appending coordinates to the latent lowers MAE on all four
tasks compared with fusing them as an expert, and the geocell
head reduces median geolocation error from 796\,km for
flow-sampled coordinates to 141\,km
(\cref{tab:fusion_location_ablation}(d,e)).

Deterministic completion improves CRPS on three of four
tasks (\cref{tab:completion_ablation}(b)).
For S2-to-S1, CRPS rises from 2.17 to 2.40 dB before
residual sampling reduces it to 1.79 dB.
With SiT-B/2 used for each network, the
deterministic predictor-flow system reduces CRPS by 6--37\% across
four tasks relative to a single flow (Tab.~\ref{tab:completion_ablation}(c)).
With the predictor fixed, residual and full-latent
targets differ in CRPS by at most 1.5\% (Tab.~\ref{tab:completion_ablation}(b)).

\subsection{Any-modal generation}
\label{sec:results_generation}

Trained on 37.5\% of TerraMesh, EAGLE achieves lower
mean-of-20 MAE than COP-GEN and TerraMind-B across the
SAR, optical and elevation targets on 3,037 scenes
(\cref{tab:loo}(a)).
When one further input is withheld, EAGLE keeps the lowest
mean-of-20 S1 and S2 MAE in every case and the lowest DEM MAE
in four of five (\cref{tab:loo}(b)).
The comparison with TerraMind-L and oracle best-of-20
results is reported in \cref{tab:peak_capability}; for example, DEM MAE
is 88.2\,m for EAGLE and 194.1\,m for TerraMind-L,
falling to 18.5\,m and 182.9\,m under best-of-20 selection.
These gains required only 941\,M parameters, 16\% fewer than
COP-GEN (1,124\,M) and 10\% fewer than TerraMind-L
(1,042\,M), with the lowest peak GPU memory in our cost
benchmark (3.9\,GB, \cref{tab:cost}).
\Cref{fig:teaser}(a,b) summarizes the gains over COP-GEN
and TerraMind-B. \Cref{fig:qa} shows generations from S1
and coordinates, with further examples in
\cref{fig:supp_qual1,fig:supp_qual2,fig:supp_qual3}.

Across the 500-scene sweep with eight draws, EAGLE ranks first in 148 of 181 configurations with non-coordinate targets (\cref{tab:context_sweep}), and its MAE falls by 47--80\% from one input to four (\cref{fig:teaser}(c)). Across 243 input additions, the largest observed degradation is 0.09\,dB in S2 L2A PSNR (adding S1 to DEM, L1C and coordinates; 95\% CI 0.03--0.15\,dB); none is significant after Holm correction (\cref{tab:supp_additions}). In the uncertainty evaluation, EAGLE achieves the lowest CRPS in nine of ten settings against COP-GEN and TerraMind-B, using eight draws on 500 scenes
(\cref{tab:uncertainty}). The exception is L2A with all other inputs, where
COP-GEN scores 0.0181 against 0.0200. For S1-to-DEM with coordinates withheld, CRPS is 197.1\,m, compared with 376.4\,m for COP-GEN and 273.3\,m for TerraMind-B;
nominal 90\% coverage is 93\%, 95\% and 63\%, respectively. Across settings, EAGLE's spread--error correlation is 0.52--0.73, while coverage is 90--93\% for DEM and 73--88\% for S1 and S2, indicating undercoverage
for the latter.

EAGLE extends to 368-band HySpex imagery. On MDAS, the mean of eight reconstructions achieves lower RMSE and higher PSNR and SSIM than pix2pix and HSCNN+ for HSI generation from S1, S2 and joint inputs
(\cref{tab:hsi}). The adapted model also reconstructs S2 from HSI, although
HSCNN+ performs better in this direction.

\input{Tables/hsi_translation}

\subsection{Downstream tasks}
\label{sec:results_downstream}
\myparagraph{Missing-band reconstruction.}
Using the mean of 20 samples, EAGLE outperforms SatMAE and
SpectralGPT+ in the red-edge/NIR and SWIR settings
(\cref{tab:missing_bands_published}).
COP-GEN has lower MAE for the 10\,m-to-20\,m setting,
while EAGLE has higher SSIM.
The reconstructed bands retain the main field and river
boundaries (\cref{fig:fullresult}).
After hyperspectral adaptation, EAGLE also reconstructs both
contiguous spectral gaps and regularly spaced missing bands.
Adding S2 improves all four hyperspectral settings, reducing
RMSE for the withheld VNIR range by 48\%.

\myparagraph{Change detection.}
% EAGLE performs change detection through the shared scene latent,
% allowing different sensor combinations at either date.
% One change head on the frozen representation handles all six
% cross-sensor pairings on SenForFlood without retraining for each
% pairing (\cref{tab:flood_cd}).
% The comparison includes SFF U-Net adapted to the flood-date
% modality and the separately trained
% ChangeFormer model for each pairing.
EAGLE achieves the highest F1 and IoU when S1 is
available during the flood, improving F1 by
2.2--5.5 percentage points over the strongest baseline,
but trails SFF U-Net when only S2 is available
(\cref{tab:flood_cd}). S1 observations provide evidence of flooding through changes in radar backscatter, including reduced returns from newly flooded open surfaces. %\TODO{What made S2 perform well}

\myparagraph{Scene question answering.}
EAGLE reaches 61.3\% multiple-choice accuracy with S1 alone
(\cref{tab:qa}).
Adding S2 raises accuracy to 68.7\%, which is 17.2 percentage points above fine-tuned RS-InternVL.
Adding DEM improves it further to 69.4\%.
The question-only control reaches 42.7\%, confirming
the benefit of scene information.

%% file: Tables/scene_QA.tex
\begin{table}[t]
\centering
\scriptsize
\setlength{\tabcolsep}{2pt}
\renewcommand{\arraystretch}{1.0}
\caption{\textbf{Scene question answering}
on BigEarthNet.txt (bench split; accuracy, \%).
No location or date inputs.
$^{\ast}$Fine-tuned.
$^{\dagger}$Question-only control with zero image latent.
Published baselines: Herzog et al.~\cite{herzog2026bigearthnettxt};
Qwen3-VL: our evaluation.
\textbf{Bold}: best; \underline{underline}: second-best.}
\begin{tabular*}{\columnwidth}{
    @{\extracolsep{\fill}}lcccccc@{}
}
\toprule
\textbf{Method} & RGB & S2 & S1 & DEM
& \textbf{Binary}$\uparrow$ & \textbf{MCQ}$\uparrow$ \\
\midrule

GeoChat
& \aYes & \aNo & \aNo & \aNo & 50.8 & 28.4 \\
EarthMind
& \aNo & \aYes & \aYes & \aNo & 57.8 & 35.3 \\
GPT-5.2
& \aYes & \aNo & \aNo & \aNo & 60.4 & 34.9 \\
Qwen3-VL
& \aYes & \aNo & \aNo & \aNo & 62.0 & 37.5 \\
RS-InternVL$^\ast$
& \aYes & \aYes & \aYes & \aNo & 73.3 & 51.5 \\
Question only$^\dagger$
& \aNo & \aNo & \aNo & \aNo & 63.8 & 42.7 \\

\rowcolor{A2Eshade}
\aOurs-QA
& \aNo & \aYes & \aNo & \aNo & 73.2 & 68.2 \\
\rowcolor{A2Eshade}
\aOurs-QA
& \aNo & \aNo & \aYes & \aNo & 70.2 & 61.3 \\
\rowcolor{A2Eshade}
\aOurs-QA
& \aNo & \aNo & \aNo & \aYes & 71.7 & 58.2 \\
\rowcolor{A2Eshade}
\aOurs-QA
& \aNo & \aYes & \aYes & \aNo
& \underline{73.5} & \underline{68.7} \\
\rowcolor{A2Eshade}
\aOurs-QA
& \aNo & \aYes & \aYes & \aYes
& \aBest{75.0} & \aBest{69.4} \\
\bottomrule
\end{tabular*}

\label{tab:qa}
\end{table}

%% file: Tables/hsi.tex
\definecolor{A2Eshade}{HTML}{EDF4F4}
\definecolor{A2Eink}{HTML}{204E56}
\providecommand{\aBest}[1]{\textcolor{A2Eink}{\textbf{#1}}}
\providecommand{\aSecond}[1]{\underline{#1}}
\providecommand{\aOurs}{\textcolor{A2Eink}{\textbf{EAGLE}}}
\providecommand{\aPanel}[1]{\textcolor{A2Eink}{\textbf{#1}}}
\providecommand{\aHead}[1]{{\scriptsize\bfseries #1}}

\begin{table*}[t!]
\centering
\scriptsize
\setlength{\tabcolsep}{3pt}
\renewcommand{\arraystretch}{1.05}
\caption{\textbf{Missing-band reconstruction.}
\textbf{(a)} S2-L2A: 500 scenes;
EAGLE and COP-GEN score the mean of 20 draws.
\textbf{(b)} HySpex: 368 bands, MDAS test area,
EAGLE fits observation parameters per band subset
on development data with neural weights fixed.
MAE and RMSE: reflectance $\times100$;
PSNR: dB; SAM: degrees.
\textbf{Bold}: best; \underline{underline}: second-best.}
\begin{tabular*}{\textwidth}{
    @{\extracolsep{\fill}}l*{12}{r}@{}
}
\toprule

\aPanel{(a) S2 L2A}
& \multicolumn{4}{c}{\aHead{10\,m$\to$20\,m bands}}
& \multicolumn{4}{c}{\aHead{Red-edge + NIR$_n$ withheld}}
& \multicolumn{4}{c}{\aHead{SWIR withheld}} \\
\cmidrule(lr){2-5}
\cmidrule(lr){6-9}
\cmidrule(l){10-13}

\aHead{Method}
& \aHead{MAE$\downarrow$}
& \aHead{PSNR$\uparrow$}
& \aHead{SSIM$\uparrow$}
& \aHead{SAM$\downarrow$}
& \aHead{MAE$\downarrow$}
& \aHead{PSNR$\uparrow$}
& \aHead{SSIM$\uparrow$}
& \aHead{SAM$\downarrow$}
& \aHead{MAE$\downarrow$}
& \aHead{PSNR$\uparrow$}
& \aHead{SSIM$\uparrow$}
& \aHead{SAM$\downarrow$} \\
\midrule

COP-GEN
& \textbf{2.33} & \textbf{20.34} & 0.546 & \textbf{5.53}
& \textemdash & \textemdash & \textemdash & \textemdash
& \textemdash & \textemdash & \textemdash & \textemdash \\

SatMAE
& 6.66 & 15.02 & \aSecond{0.556} & 12.19
& \aSecond{5.24} & \aSecond{19.68}
& \aSecond{0.682} & \aSecond{5.14}
& 7.61 & \aSecond{14.49}
& \aSecond{0.546} & \aSecond{6.55} \\

SpectralGPT+
& \textemdash & \textemdash & \textemdash & \textemdash
& 5.40 & 16.09 & 0.475 & 9.51
& \aSecond{5.85} & 14.19 & 0.531 & 7.41 \\

\rowcolor{A2Eshade}
\aOurs{}
& \aSecond{2.77} & \aSecond{19.16}
& \aBest{0.639} & \aSecond{6.30}
& \aBest{2.13} & \aBest{23.03}
& \aBest{0.699} & \aBest{3.16}
& \aBest{3.36} & \aBest{17.48}
& \aBest{0.613} & \aBest{3.84} \\

\midrule
\aPanel{(b) HySpex}
& \multicolumn{3}{c}{\aHead{Visible: 417--678\,nm}}
& \multicolumn{3}{c}{\aHead{VNIR: 690--1000\,nm}}
& \multicolumn{3}{c}{\aHead{SWIR: $\geq$1400\,nm}}
& \multicolumn{3}{c}{\aHead{Every fourth band}} \\
\cmidrule(lr){2-4}
\cmidrule(lr){5-7}
\cmidrule(lr){8-10}
\cmidrule(l){11-13}

\aHead{Method}
& \aHead{RMSE$\downarrow$}
& \aHead{PSNR$\uparrow$}
& \aHead{SAM$\downarrow$}
& \aHead{RMSE$\downarrow$}
& \aHead{PSNR$\uparrow$}
& \aHead{SAM$\downarrow$}
& \aHead{RMSE$\downarrow$}
& \aHead{PSNR$\uparrow$}
& \aHead{SAM$\downarrow$}
& \aHead{RMSE$\downarrow$}
& \aHead{PSNR$\uparrow$}
& \aHead{SAM$\downarrow$} \\
\midrule

DHP
& 42.31 & 7.48 & 38.08
& 29.55 & 10.76 & 25.47
& 35.19 & 9.15 & 32.21
& 34.25 & 9.59 & 31.49 \\

\rowcolor{A2Eshade}
\aOurs{} (HSI only)
& \aSecond{3.64} & \aSecond{28.83} & \aSecond{4.13}
& \aSecond{9.31} & \aSecond{21.13} & \aSecond{10.06}
& \aSecond{6.45} & \aSecond{23.86} & \aSecond{8.07}
& \aSecond{5.15} & \aSecond{26.36} & \aSecond{6.73} \\

\rowcolor{A2Eshade}
\aOurs{} (HSI + S2)
& \aBest{3.02} & \aBest{30.44} & \aBest{3.33}
& \aBest{4.88} & \aBest{26.55} & \aBest{5.16}
& \aBest{5.79} & \aBest{24.88} & \aBest{7.15}
& \aBest{4.04} & \aBest{28.32} & \aBest{5.18} \\

\bottomrule
\end{tabular*}
\label{tab:missing_bands_published}
\end{table*}

%% file: Tables/hsi_translation.tex
\begin{table}[t]
\centering
\scriptsize
\setlength{\tabcolsep}{2pt}
\renewcommand{\arraystretch}{1.08}
\caption{\textbf{Hyperspectral transfer on MDAS}
(368 bands, 10\,m grid; S2-L2A).
All methods use the same training area.
EAGLE averages eight reconstructions.
RMSE: reflectance $\times100$;
SAM: degrees; PSNR: dB.
\textbf{Bold}: best; \underline{underline}: second-best
per transfer.}
\begin{tabular*}{\columnwidth}{
    @{\extracolsep{\fill}}llrrrr@{}
}
\toprule
\aHead{Transfer} & \aHead{Method}
& \aHead{RMSE$\downarrow$}
& \aHead{SAM$\downarrow$}
& \aHead{PSNR$\uparrow$}
& \aHead{SSIM$\uparrow$} \\
\midrule

\aPanel{S1$\to$HSI}
& pix2pix
& 10.74 & 16.36 & 19.99 & 0.216 \\
& HSCNN+
& \aSecond{9.44} & \aSecond{14.65}
& \aSecond{21.09} & \aSecond{0.290} \\
\rowcolor{A2Eshade}
\cellcolor{white} & \aOurs{}
& \aBest{8.28} & \aBest{12.97}
& \aBest{22.17} & \aBest{0.315} \\
\midrule

\aPanel{S2$\to$HSI}
& pix2pix
& 4.30 & 7.75 & 27.90 & 0.811 \\
& HSCNN+
& \aSecond{3.60} & \textbf{5.25}
& \aSecond{29.73} & \aSecond{0.873} \\
\rowcolor{A2Eshade}
\cellcolor{white} & \aOurs{}
& \aBest{3.48} & \aSecond{5.40}
& \aBest{30.00} & \aBest{0.878} \\
\midrule

\aPanel{S1+S2$\to$HSI}
& pix2pix
& 4.02 & 6.91 & 28.50 & 0.835 \\
& HSCNN+
& \aSecond{3.63} & \textbf{5.26}
& \aSecond{29.64} & \aSecond{0.868} \\
\rowcolor{A2Eshade}
\cellcolor{white} & \aOurs{}
& \aBest{3.49} & \aSecond{5.46}
& \aBest{29.98} & \aBest{0.877} \\
\midrule

\aPanel{HSI$\to$S2}
& pix2pix
& 2.94 & 5.00 & 30.95 & \aSecond{0.887} \\
& HSCNN+
& \textbf{2.57} & \textbf{4.02}
& \textbf{32.15} & \textbf{0.920} \\
\rowcolor{A2Eshade}
\cellcolor{white} & \aOurs{}
& \aSecond{2.78} & \aSecond{4.62}
& \aSecond{31.61} & 0.883 \\
\bottomrule
\end{tabular*}

\label{tab:hsi}
\end{table}

%% file: sec/6_conclusion.tex
\section{Conclusion and Limitations}

We introduced EAGLE, a probabilistic framework for scene inference and generation from arbitrary subsets of Earth observation sensors. Our contribution lies in integrating explicit linear-Gaussian sensor fusion with deterministic prediction and stochastic residual completion. Our results support the hypothesis that heterogeneous sensor representations can be modeled as noisy linear views of a shared latent, achieving lower reconstruction error than transformer fusion with 56 times fewer fusion parameters.

Across 3,037 scenes, EAGLE achieves lower mean reconstruction error than COP-GEN and TerraMind across optical, SAR, and elevation targets. The shared representation also supports missing-band reconstruction, hyperspectral adaptation, cross-sensor change detection, and scene question answering, demonstrating its utility for both generation and downstream scene understanding.
Our approach assumes conditional independence and fixed sensor noise estimates, while generated samples underestimate uncertainty for optical and SAR observations. Fine spatial details and elevation prediction without coordinates also remain challenging. Future work will explore more flexible observation models and improved uncertainty calibration to support reliable inference under heterogeneous sensing conditions.

%% file: sec/X_suppl.tex
\clearpage
\setcounter{page}{1}
\maketitlesupplementary

% Supplement numbering: S1, S2, ... for sections, tables, figures and equations (main-paper references read
% "Sec. S2", "Tab. S3"); \theH* keeps hyperref anchors distinct from the main paper's.
\setcounter{section}{0}\setcounter{table}{0}\setcounter{figure}{0}\setcounter{equation}{0}
\renewcommand{\thesection}{S\arabic{section}}
\renewcommand{\thetable}{S\arabic{table}}
\renewcommand{\thefigure}{S\arabic{figure}}
\renewcommand{\theequation}{S\arabic{equation}}
\providecommand{\theHsection}{}\renewcommand{\theHsection}{S\arabic{section}}
\providecommand{\theHtable}{}\renewcommand{\theHtable}{S\arabic{table}}
\providecommand{\theHfigure}{}\renewcommand{\theHfigure}{S\arabic{figure}}
\providecommand{\theHequation}{}\renewcommand{\theHequation}{S\arabic{equation}}

% =====================================================================================================================
% Numbering fixed by the submitted main paper (do not reorder):
%   Sec. S2.4 = sup:decoder_loss; Sec. S3 = sup:eval; Sec. S3.1 = sup:eval_sets; Sec. S4.5 = sup:additional_locations
%   Tab. S1 = tab:cost; Tab. S4 = tab:peak_capability; Tab. S5 = tab:context_sweep; Tab. S13 = tab:supp_additions;
%   Tab. S14 = tab:uncertainty; Figs. S2-S4 = fig:supp_qual1-3.
% Table numbers follow the source order of \caption: exactly two tables (S2, S3) between Sup/cost and Tables/best20,
% the seven tables of Sup/fullcontext (S6-S12) between context_sweep and supp_additions; only Fig. S1 before the
% qualitative figures. New tables go after Tab. S14.
% Previous versions: scratchpad X_suppl.before_rewrite_19nov.tex, X_suppl.before_trim.tex, X_suppl.before_constraint.tex.
% RULE (author, 19 Nov; CVPR): no results from an improved method (extra tuning or training) and no corrected main-paper
% content. Author's decision (19 Nov): the perceptual decoder stays, presented only as a fidelity-for-realism trade-off.
% =====================================================================================================================

% Table style of the main paper (Tables/*.tex define these first; repeated here so the supplement also compiles alone).
\definecolor{A2Eshade}{HTML}{EDF4F4}\definecolor{A2Eink}{HTML}{204E56}
\providecommand{\aBest}[1]{\textcolor{A2Eink}{\textbf{#1}}}
\providecommand{\aOurs}{\textcolor{A2Eink}{\textbf{EAGLE}}}
\providecommand{\aScoreCI}[2]{#1\,{\fontsize{6}{7}\selectfont[#2]}}

% Dense float packing in the supplement.
\renewcommand{\dbltopfraction}{0.99}\setcounter{dbltopnumber}{4}\renewcommand{\dblfloatpagefraction}{0.5}
\renewcommand{\topfraction}{0.95}\setcounter{topnumber}{4}\renewcommand{\bottomfraction}{0.9}\setcounter{bottomnumber}{2}
\setcounter{totalnumber}{8}\renewcommand{\textfraction}{0.05}\renewcommand{\floatpagefraction}{0.5}

\section{Overview}
\label{sup:overview}
This supplement details the implementation (\cref{sup:impl}) and evaluation (\cref{sup:eval}) of \textbf{EAGLE}, and
adds S2 generation with both S2 products withheld (\cref{tab:supp_paired}), fusion of a combination held out in
training (\cref{tab:supp_fusion}), completion against a matched-size flow (\cref{tab:supp_completion}), a held-out
test of the fitted fusion model (\cref{tab:supp_calibration}) and downstream tasks (\cref{sup:downstream}).

\input{Sup/implementation}

\input{Sup/eval}
\input{Tables/best20}

\section{Additional generation results}
\label{sup:gen}
\begin{figure*}
    \centering
    \includegraphics[width=1\linewidth]{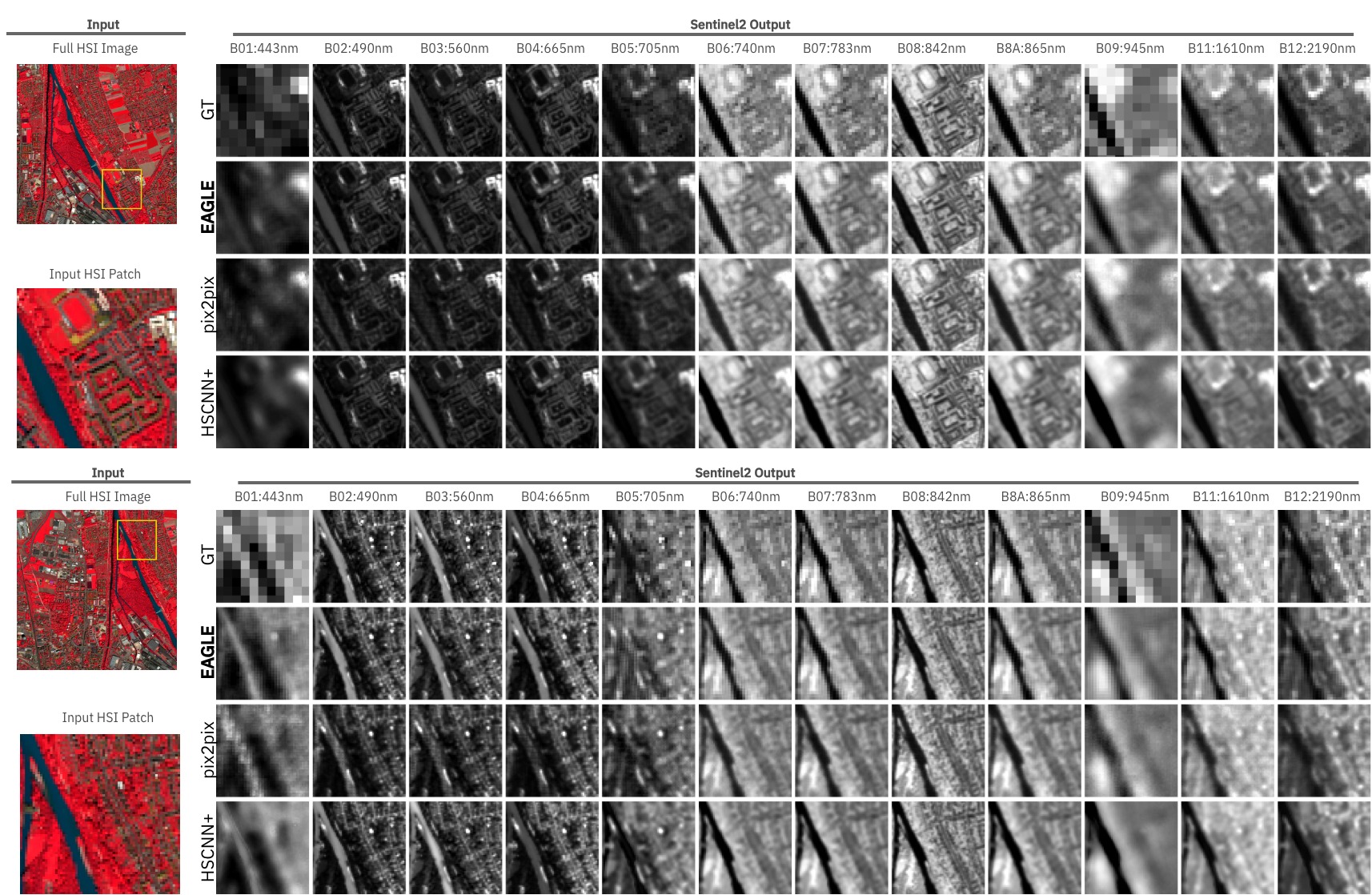}
    \caption{\textbf{S2 generated from HySpex} on two MDAS test patches (\cref{tab:hsi}). Left: the HySpex test window in
    false colour with the patch (yellow box), and the patch enlarged. Right: the 12 S2 L2A bands of the patch: ground
    truth (GT), EAGLE (mean of eight draws), pix2pix and HSCNN+; one stretch per band, set by the ground truth. Typical
    patches: among patches with at least median detail, EAGLE's RMSE minus the better baseline's is closest to its
    median. Over all test patches, HSCNN+ has the lowest error in this direction (\cref{tab:hsi}).}
    \label{fig:supp_hsi}
\end{figure*}

\Cref{fig:supp_qual1,fig:supp_qual2,fig:supp_qual3} show generation from one, two and three inputs,
\cref{fig:supp_hsi} S2 generated from HySpex, and \cref{fig:supp_dist_all} per-band S2 L2A pixel distributions over 500
scenes as inputs are added, with \cref{fig:supp_dist1} showing one of them.

\providecommand{\aQualKey}{Per scene, top: inputs and references; below: each model's reconstructed inputs and
generated outputs. S2 L1C and L2A in false colour, $^{*}$true colour. Map: star, true location; dot, prediction.}

\subsection{Main comparison}
\label{sup:best20}
\label{sup:paired}
\Cref{tab:peak_capability} adds TerraMind-L and best-of-20 scores to \cref{tab:loo}(a); EAGLE has the lowest MAE on every
sensor target under both. Paired tests confirm all MAE gains in \cref{tab:loo}(a), including S2 generated with both S2
products withheld (\cref{tab:supp_paired}): from S1 RTC, DEM and coordinates, EAGLE's MAE is 0.0405 for L1C and 0.0527
for L2A, less than half that of every baseline. With each draw's MAE averaged over all scenes, the worst of EAGLE's 20 draws
beats the best of either baseline's 20 draws on each sensor target (DEM: 90.1\,m versus 197.3\,m). Error does not increase with distance from the training
scenes, and EAGLE has the lowest MAE in every distance bin (\cref{tab:train_distance}).

\subsection{Input configurations}
\label{sup:sweep}
\label{sup:additions}
\Cref{tab:context_sweep} summarises the 211 input configurations, with full results in
\crefrange{tab:full-dem}{tab:full-geolocation}; rows of \cref{tab:full-s2l1c,tab:full-s2l2a} with neither S2 product as
input give generation with both withheld. Across 243 input additions, none makes EAGLE significantly worse after Holm
correction, against 3 for COP-GEN and 2 for TerraMind-B on the same additions, and 174 make EAGLE significantly better,
against 127 and 91 (\cref{tab:supp_additions}).
Over the 500 scenes of \cref{tab:uncertainty}, EAGLE's pooled S2 L2A pixel distribution is the closest to the reference
in every band at each step from DEM to all four inputs (Wasserstein-1 distance; \cref{fig:supp_dist_all}). From S1 RTC
alone, the per-band distribution of EAGLE's eight draws is closer than COP-GEN's in 86--93\% of scenes and closer than
TerraMind-B's in 65--88\%. \Cref{fig:supp_dist1} shows a typical scene.

\subsection{Uncertainty}
\label{sup:uncertainty}
EAGLE has the lowest CRPS in nine of ten settings (\cref{tab:uncertainty}).

\subsection{Image realism and fidelity}
\label{sup:realism}
EAGLE prioritises reconstruction of the reference scene from the available observations. In \cref{tab:supp_realism}, the standard EAGLE decoder gives the lowest MAE and highest SSIM, while COP-GEN achieves the lowest FID and KID. These measures assess different properties: MAE and SSIM compare each reconstruction with its scene reference, whereas FID and KID compare image-feature distributions without assessing correspondence to individual scenes.

We examine this trade-off through a diagnostic decoder ablation that combines latent smoothing with perceptual and adversarial training. FID falls by 38--41, while MAE rises by 1.6--4.0\% and SSIM falls. This ablation is used only in \cref{tab:supp_realism}; all other results use the standard EAGLE decoder.

For the ablation, we train a copy of the optical decoder while keeping the original model fixed. Latents are Gaussian-smoothed with $\sigma=2$ cells. We train on full-evidence states for 4,000 updates using reconstruction, LPIPS and PatchGAN losses (batch 8, learning rate $10^{-4}$). A further 1,000 updates add an adversarial loss on sampled states and consistency with the original decoder at 80\,m resolution (learning rate $5\times10^{-5}$, EMA 0.999).

\subsection{Additional locations}
\label{sup:additional_locations}
We evaluate EAGLE and TerraMind-B on 1,386 SSL4EO-S12 scenes, using S1 GRD in place of S1 RTC. COP-GEN is excluded because it supports only S1 RTC. EAGLE has lower MAE in all twelve settings, for DEM generation from S1 and coordinates, MAE is 91.9\,m compared with 132.6\,m for TerraMind-B (\cref{tab:supp_additional_locations}).

\section{Ablations}
\label{sup:fusion}
Every gap between EAGLE and transformer fusion in \cref{tab:fusion_location_ablation}(a) is significant, including
S1+DEM, a combination held out in training ($-$12.3\%; \cref{tab:supp_fusion}). The arms differ only in the fusion module; the transformer receives the encoder's
means and precisions (the Prec.\ column of \cref{tab:fusion_location_ablation}(a) applies to the PoE arms). Whenever two or more sensors are fused (S2 and S1, S1 and DEM, or all), EAGLE's error on S2, S1 and
DEM is also significantly lower in 8 of these 9 input--target pairs ($-$12\% to $-$48\%) and equal in the ninth (DEM
from S2 and S1, $-$0.1\% [$-$2.5, $+$2.5]; \cref{tab:supp_fusion}, bottom).
\Cref{tab:supp_calibration} tests the fitted fusion model on held-out scenes. For every withheld sensor, the
posterior-predictive error of its features matches the realised error, averaged over scenes, to within 12\% (ratio
0.98--1.12).

Predicting before sampling lowers CRPS significantly on all four tasks of \cref{tab:completion_ablation}(c), at the same
backbone and number of updates (\cref{tab:supp_completion}, top). The gain does not come from the second network's
parameters: against a single flow of matched size (24 blocks, 308.1\,M parameters against 311.1\,M for B1 + residual flow),
trained for 16k or 32k updates, B1 + residual flow still lowers CRPS significantly on all four tasks, by 5--42\% and 3--37\%
(\cref{tab:supp_completion}, middle and bottom).

% ---------------------------------------------------------------------------------------------------------------------
\section{Downstream tasks}
\label{sup:downstream}
\begin{figure*}
    \centering
    \includegraphics[width=0.72\linewidth]{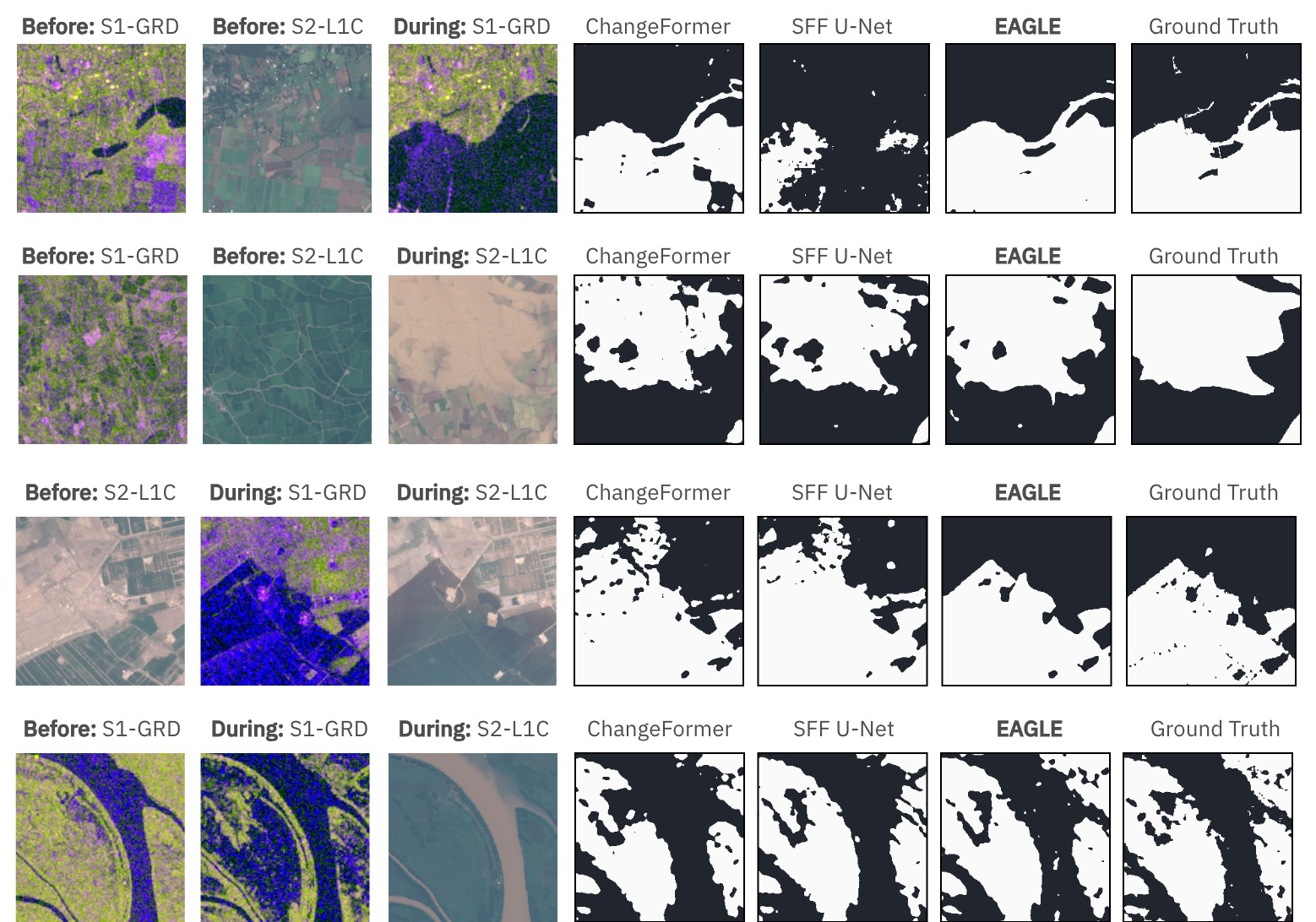}
    \caption{\textbf{Cross-sensor flood change detection} on SenForFlood test tiles (\cref{tab:flood_cd}). Rows, before
    $\to$ during: S1+S2$\to$S1, S1+S2$\to$S2, S2$\to$S1+S2 and S1$\to$S1+S2. S1 GRD in false colour (VV, VH and
    VV$-$VH as red, green and blue); S2 L1C in true colour; white: newly flooded. }
    \label{fig:supp_cd}
\end{figure*}

\begin{figure*}
    \centering
    \includegraphics[width=1\linewidth]{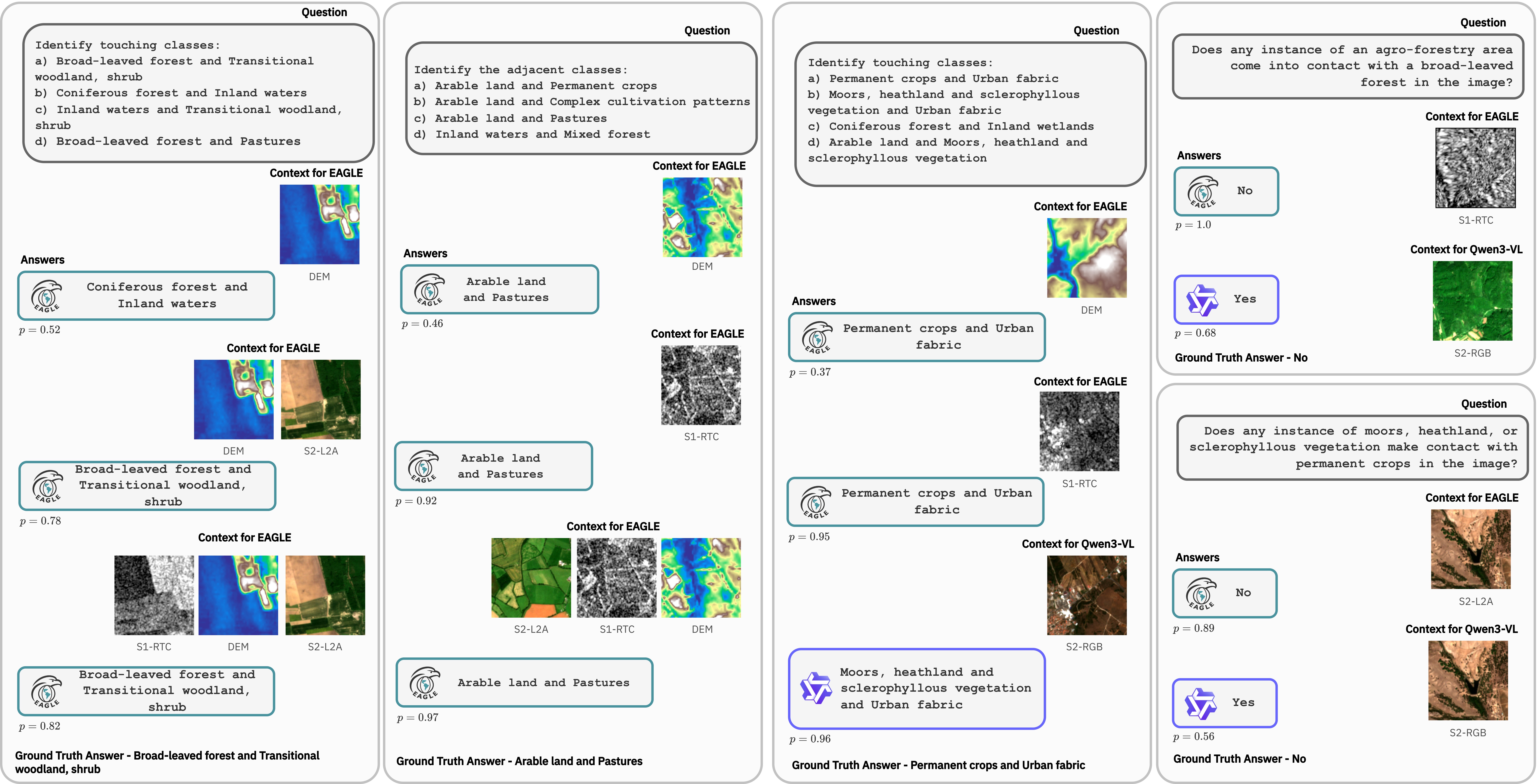}
\caption{\textbf{Scene question answering: EAGLE's confidence increases with context.} BigEarthNet.txt bench
    questions (\cref{tab:qa}). Each box gives a model's answer from the context beside it and its confidence $p$.
    Qwen3-VL receives BigEarthNet's RGB patch. Left: a question answered wrongly from DEM and correctly once S2 and then
    S1 are added. Columns 2--4: questions EAGLE-QA answers correctly from every sensor set, while Qwen3-VL (columns 3
    and 4) is wrong; in columns 2 and 3, EAGLE-QA's confidence is lowest from DEM alone (0.46 and 0.37) and reaches
    0.92--0.97 with S1 or all three sensors. Over the bench, adding S2 and S1 to DEM corrects 1,551 answers, a net gain
    of 840, and raises the mean confidence in the correct answer from 0.62 to 0.69.}
    \label{fig:supp_qa}
\end{figure*}

\Cref{fig:supp_cd} shows flood change detection (\cref{tab:flood_cd}) and \cref{fig:supp_qa} scene question answering
(\cref{tab:qa}). With EAGLE's pretrained weights frozen, one QA head is trained across all seven non-empty subsets of
S2 L2A, S1 RTC and DEM, with no location or date inputs; the question-only control uses a zero image latent.

% ---------------------------------------------------------------------------------------------------------------------
% Floats after the text (Tab. S4 sits on the last text page): Figs. S2--S8 (S5 with Tab. S5; S8 with Tab. S6), Tabs.
% S6--S14, Tabs. S15--S19 in one float (one page), then Fig. S9 with Tab. S20 on the last page.
\clearpage

\begin{figure*}
    \centering
    \includegraphics[width=0.87\linewidth]{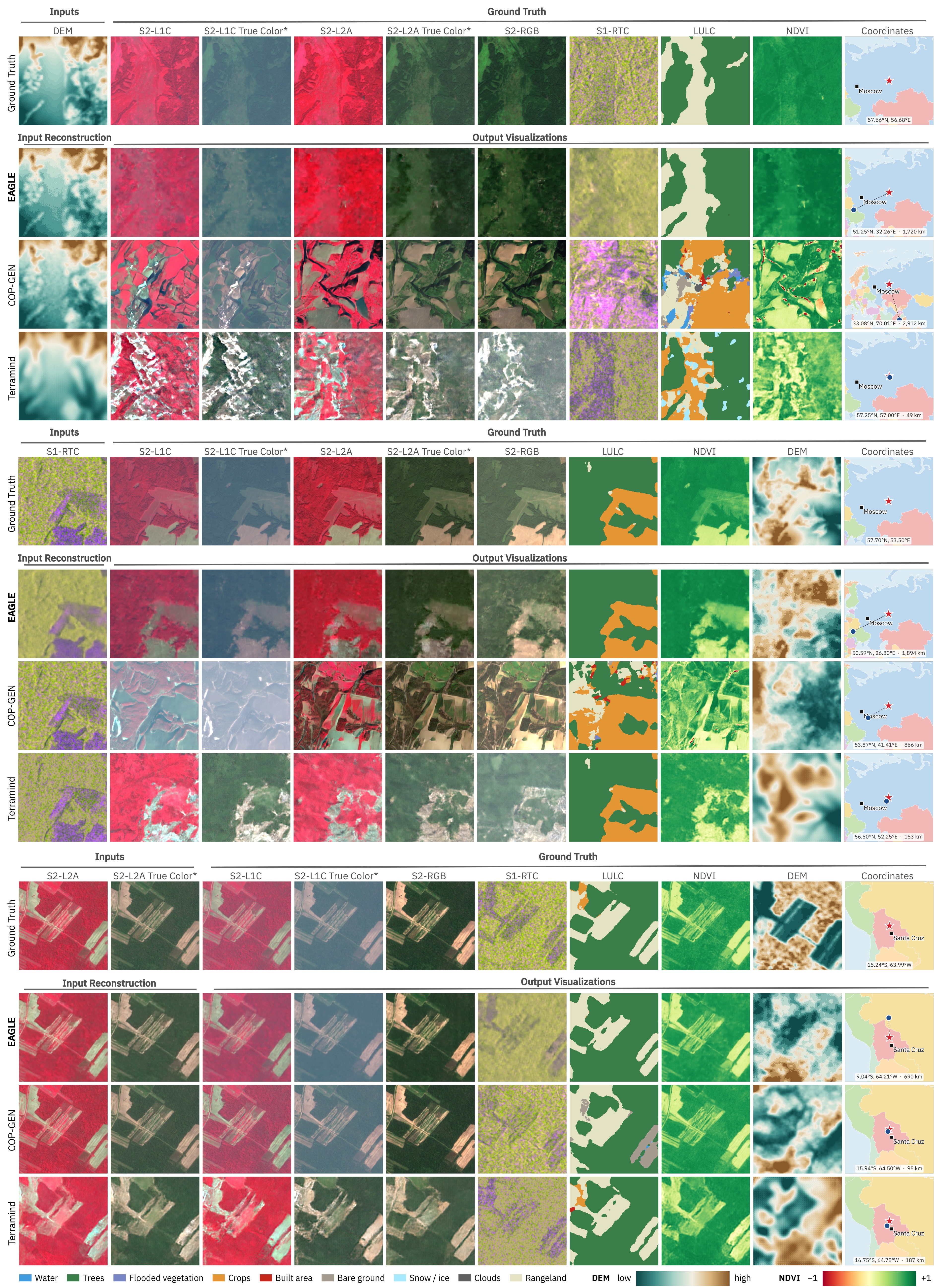}
    \caption{\textbf{Generation from one input:} DEM (top), S1 RTC (middle), S2 L2A (bottom). \aQualKey}
    \label{fig:supp_qual1}
\end{figure*}

\begin{figure*}
    \centering
    \includegraphics[width=0.87\linewidth]{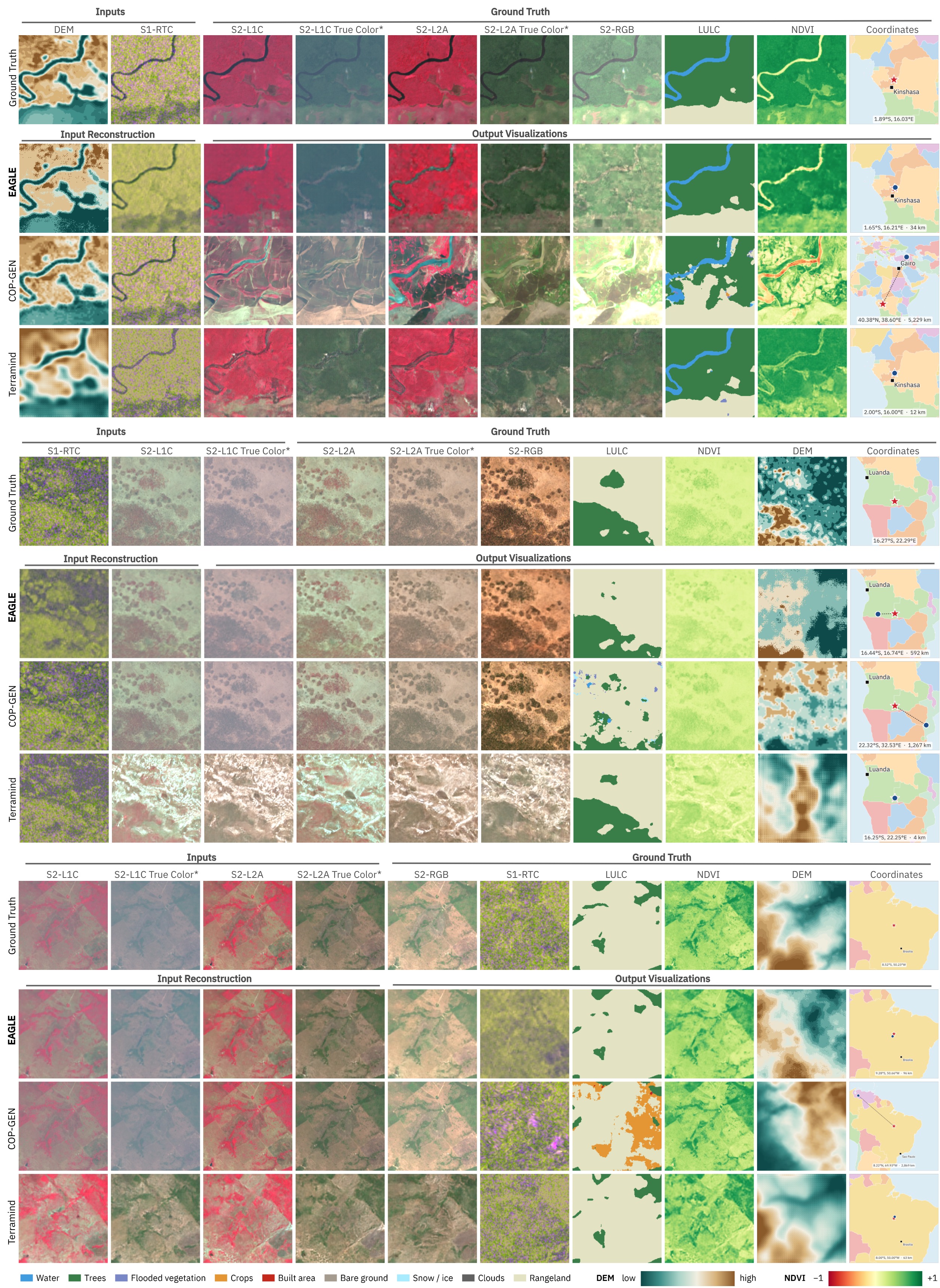}
    \caption{\textbf{Generation from two inputs:} DEM and S1 RTC (top), S1 RTC and S2 L1C (middle), S2 L1C and L2A
    (bottom). \aQualKey}
    \label{fig:supp_qual2}
\end{figure*}

\begin{figure*}
    \centering
    \includegraphics[width=0.87\linewidth]{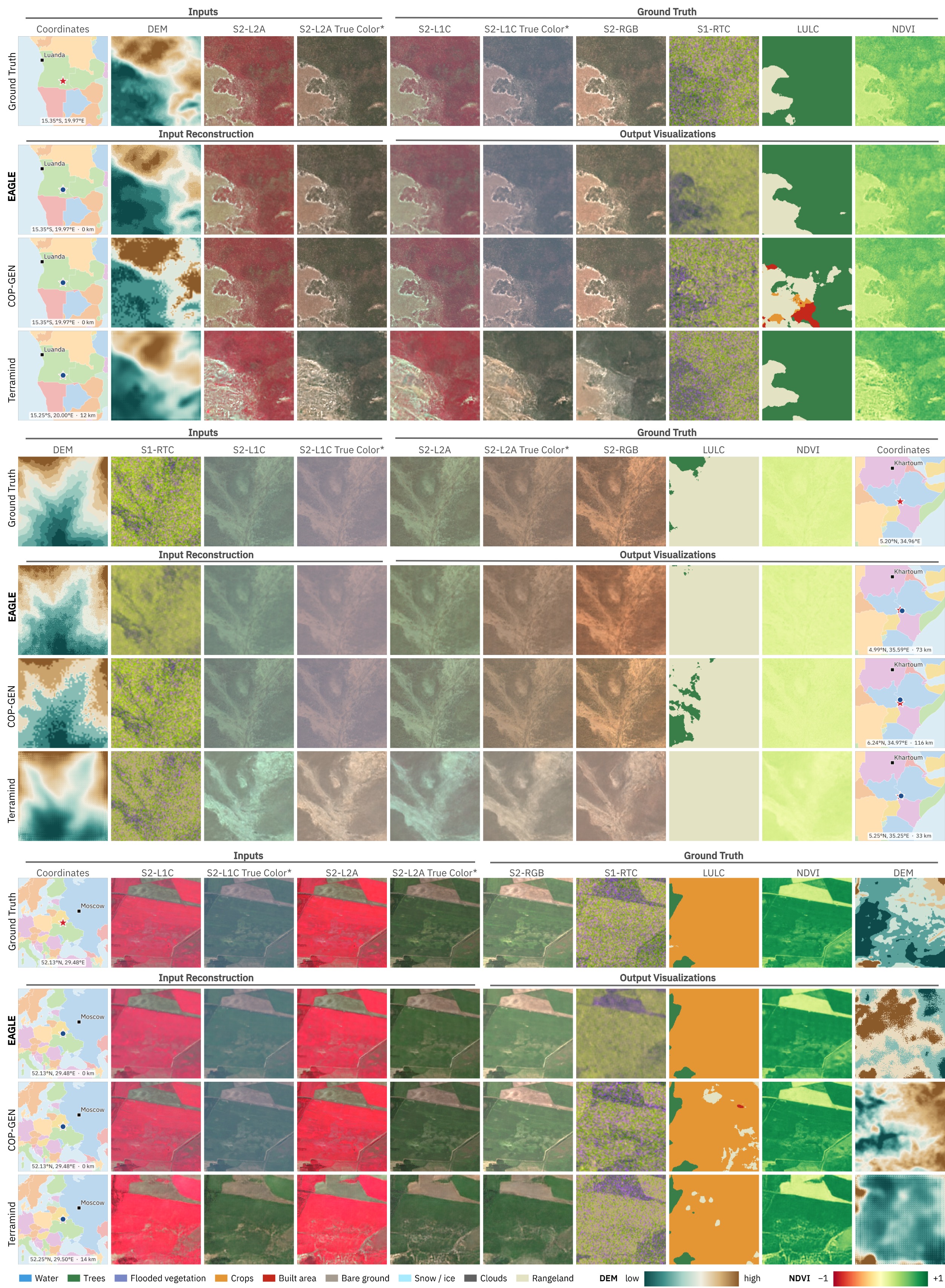}
    \caption{\textbf{Generation from three inputs:} coordinates, DEM and S2 L2A (top); DEM, S1 RTC and S2 L1C (middle);
    coordinates, S2 L1C and L2A (bottom). \aQualKey}
    \label{fig:supp_qual3}
\end{figure*}

\input{Tables/context_sweep}

\begin{figure*}
    \centering
    \includegraphics[width=0.95\linewidth]{Figs/fulltest.jpg}
    \caption{\textbf{Over 500 scenes, EAGLE's per-band S2 L2A distribution is closest to the reference in every band and at
    every input step.} For each band, a density histogram with a kernel density estimate of the pixel values over the
    central $192\times192$ pixels, pooled over the 500 scenes of \cref{tab:uncertainty}: the references (GT) and eight
    draws per model and scene. Each row adds an input: DEM, then S1 RTC, coordinates and S2 L1C. A model's distribution is
    closest when its Wasserstein-1 distance to the reference is smallest. Pooled, EAGLE's distance is 20--88\% lower than
    that of the closer baseline. Per scene and band, EAGLE's distribution is closer than COP-GEN's in 82--96\% of scenes
    without S2 input and in 53--71\% given S2 L1C, and closer than TerraMind-B's in 57--91\% and 67--99\%.}
    \label{fig:supp_dist_all}
\end{figure*}

\begin{figure*}
    \centering
    \includegraphics[width=0.97\linewidth]{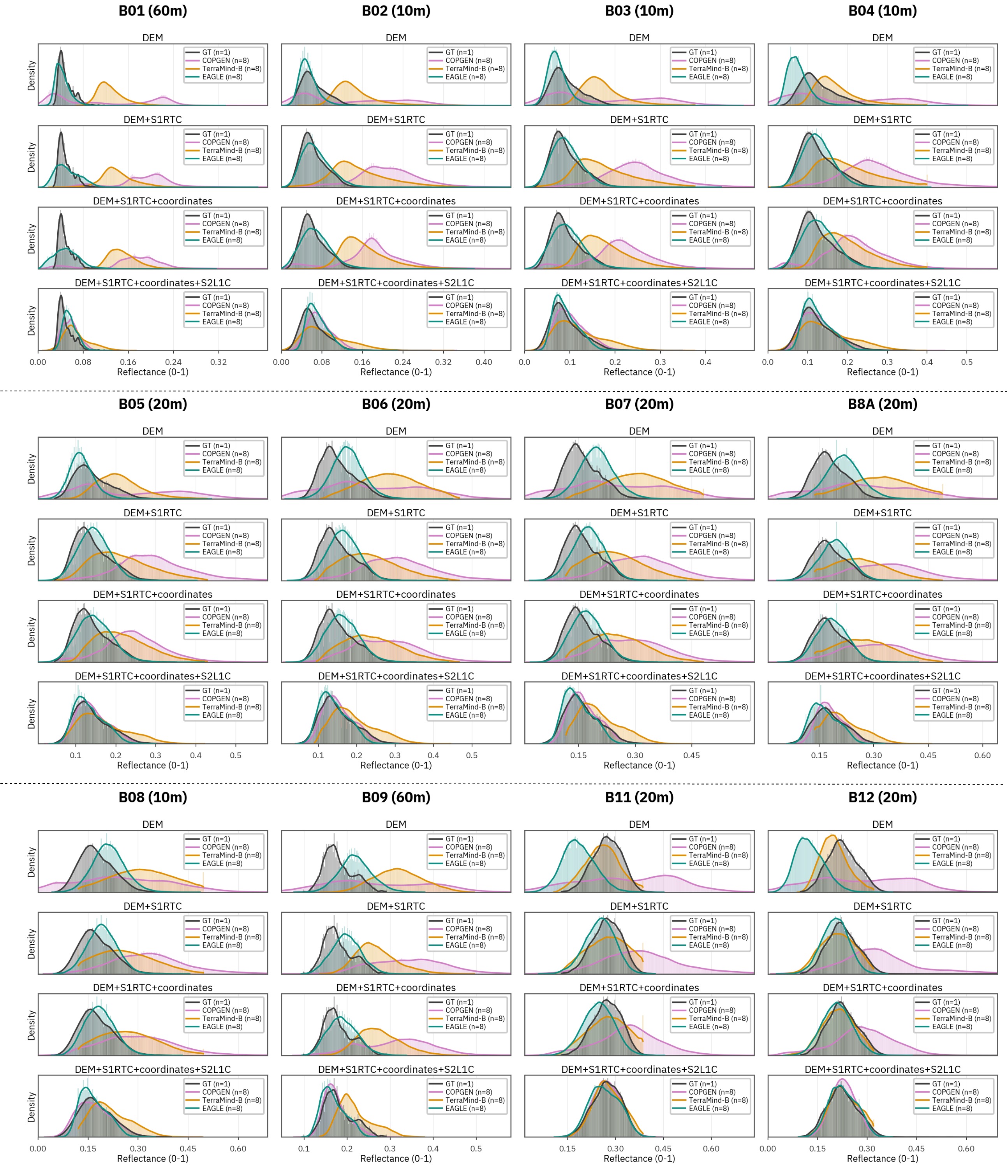}
    \caption{\textbf{Without S2 input, EAGLE's per-band S2 L2A distribution is closest to the reference in most bands.}
    As \cref{fig:supp_dist_all}, for MajorTOM validation scene 0042409 (one reference image). EAGLE's distribution is
    closest in 9--10 of the 12 bands without S2 input, COP-GEN's in 8 given S2 L1C. The scene is typical: EAGLE's
    Wasserstein-1 distance to the reference (B04, all four inputs) is closest to its median over the 500 scenes.}
    \label{fig:supp_dist1}
\end{figure*}

\input{Sup/fullcontext}

\begin{table*}[t]
\centering
\begin{minipage}[t]{0.485\textwidth}
\centering
\caption{\textbf{None of 243 input additions makes EAGLE significantly worse after Holm correction.} 500 scenes, mean of 8
draws; each input set against the same set plus one input. Top: EAGLE; gain: median improvement. Bottom: the same 243
additions for each model; Sig.: Holm-corrected $p<0.05$. \textbf{Bold}: best.}
\label{tab:supp_additions}
\scriptsize
\setlength{\tabcolsep}{2pt}
\renewcommand{\arraystretch}{1.05}
\begin{tabular*}{\linewidth}{@{\extracolsep{\fill}}llrrr@{}}
\toprule
\textbf{Target} & \textbf{Metric} & \textbf{Additions} & \textbf{Worse} & \textbf{Median gain} \\
\midrule
DEM & MAE (m) & 28 & 0 & 71.9 \\
S1 RTC & MAE (dB) & 28 & 0 & 0.08 \\
S2 L1C & PSNR (dB) & 28 & 0 & 0.99 \\
S2 L2A & PSNR (dB) & 28 & 2 & 0.79 \\
NDVI & MAE & 28 & 2 & 0.012 \\
LULC & mIoU & 75 & 6 & 0.017 \\
Location & error (km) & 28 & 0 & 712 \\
\midrule
\textbf{Model} & & \textbf{Sig.\ better}$\uparrow$ & \textbf{Worse}$\downarrow$ & \textbf{Sig.\ worse}$\downarrow$ \\
\midrule
COP-GEN & & 127 & 49 & 3 \\
TerraMind-B & & 91 & 54 & 2 \\
\rowcolor{A2Eshade}
\aOurs & & \aBest{174} & \aBest{10} & \aBest{0} \\
\bottomrule
\end{tabular*}
\end{minipage}
\end{table*}

\input{Sup/tab_uncertainty}

% Tabs. S15-S19 on one page: left paired, realism, additional locations; right fusion, completion.
\begin{table*}[t]
\begin{minipage}[t]{0.485\textwidth}
\centering
\caption{\textbf{Every MAE lead of EAGLE is significant.} EAGLE's mean-of-20 MAE and its difference to each baseline on
the 3,037 scenes of \cref{tab:loo}(a) [95\% paired block-bootstrap interval]. Both S2 withheld: inputs S1 RTC, DEM and
coordinates. S2 in reflectance. \textbf{Bold}: best.}
\label{tab:supp_paired}
\scriptsize
\setlength{\tabcolsep}{2pt}
\renewcommand{\arraystretch}{1.05}
\begin{tabular*}{\linewidth}{@{\extracolsep{\fill}}l>{\columncolor{A2Eshade}}ccc@{}}
\toprule
\textbf{Target} & \aOurs & \textbf{$-$ COP-GEN} & \textbf{$-$ TerraMind-B} \\
\midrule
DEM (m) & \aBest{88.2} & \aScoreCI{$-$230.0}{$-$239.0, $-$220.8} & \aScoreCI{$-$110.1}{$-$133.8, $-$87.6} \\
S1 RTC (dB) & \aBest{2.36} & \aScoreCI{$-$1.30}{$-$1.33, $-$1.26} & \aScoreCI{$-$0.48}{$-$0.53, $-$0.44} \\
S2 L1C & \aBest{0.0140} & \aScoreCI{$-$0.0487}{$-$0.0495, $-$0.0480} & \aScoreCI{$-$0.0944}{$-$0.0984, $-$0.0905} \\
S2 L2A & \aBest{0.0243} & \aScoreCI{$-$0.0132}{$-$0.0140, $-$0.0126} & \aScoreCI{$-$0.0713}{$-$0.0761, $-$0.0668} \\
\midrule
\multicolumn{4}{@{}l}{\textit{Both S2 products withheld}} \\
S2 L1C & \aBest{0.0405} & \aScoreCI{$-$0.0908}{$-$0.0930, $-$0.0885} & \aScoreCI{$-$0.0817}{$-$0.0852, $-$0.0783} \\
S2 L2A & \aBest{0.0527} & \aScoreCI{$-$0.0974}{$-$0.1003, $-$0.0946} & \aScoreCI{$-$0.0646}{$-$0.0689, $-$0.0601} \\
\bottomrule
\end{tabular*}

\vspace{65pt}
\caption{\textbf{EAGLE has lower reconstruction error, while COP-GEN has lower FID and KID.} S2 generated from
coordinates, DEM and S1 RTC (3,037 scenes, one draw, true colour). FID and KID ($\times1000$) against the references;
MAE (reflectance) and SSIM on the true-colour bands against each scene's own reference. Other real scene: real images of different scenes in
place of generations. $D^{p}$: decoder trained with adversarial and perceptual losses (\cref{sup:realism}).
\textbf{Bold}: best; \underline{underline}: second-best.}
\label{tab:supp_realism}
\scriptsize
\setlength{\tabcolsep}{2.5pt}
\resizebox{\linewidth}{!}{%
\begin{tabular}{@{}l cccc cccc@{}}
\toprule
& \multicolumn{4}{c}{\textbf{S2 L2A}} & \multicolumn{4}{c}{\textbf{S2 L1C}} \\
\cmidrule(lr){2-5}\cmidrule(l){6-9}
\textbf{Model} & FID$\downarrow$ & KID$\downarrow$ & MAE$\downarrow$ & SSIM$\uparrow$
               & FID$\downarrow$ & KID$\downarrow$ & MAE$\downarrow$ & SSIM$\uparrow$ \\
\midrule
\textit{Other real scene} & \textit{30.7} & \textit{0.1} & \textit{0.1748} & \textit{0.155}
                          & \textit{26.8} & \textit{0.1} & \textit{0.1386} & \textit{0.198} \\
\midrule
COP-GEN & \textbf{73.5} & \textbf{41.5} & 0.1620 & 0.137 & \textbf{66.1} & \textbf{38.5} & 0.1386 & 0.155 \\
TerraMind-B & 125.7 & 81.4 & 0.1224 & 0.147 & 148.7 & 105.9 & 0.1050 & 0.118 \\
\rowcolor{A2Eshade}
\aOurs & 139.3 & 100.3 & \aBest{0.0486} & \aBest{0.276} & 134.0 & 100.4 & \aBest{0.0350} & \aBest{0.316} \\
\rowcolor{A2Eshade}
\aOurs, $D^{p}$ & \underline{100.9} & \underline{60.7} & \underline{0.0494} & \underline{0.258}
               & \underline{92.8} & \underline{51.9} & \underline{0.0364} & \underline{0.267} \\
\bottomrule
\end{tabular}}

\vspace{62pt}
\caption{\textbf{EAGLE has lower MAE than TerraMind-B in every setting at 1,386 further locations.} SSL4EO-S12 scenes,
S1 GRD in place of RTC, coordinates given; MAE $\downarrow$, mean of 20 draws; S2 in reflectance. COP-GEN takes S1 RTC
only. \textbf{Bold}: best.}
\label{tab:supp_additional_locations}
\scriptsize
\setlength{\tabcolsep}{2pt}
\renewcommand{\arraystretch}{1.05}
\begin{tabular*}{\linewidth}{@{\extracolsep{\fill}}ll>{\columncolor{A2Eshade}}rr@{}}
\toprule
\textbf{Target} & \textbf{Inputs} & \aOurs & \textbf{TerraMind-B} \\
\midrule
\multirow{3}{*}{DEM (m)} & S1 & \aBest{91.9} & 132.6 \\
 & S1, S2 L2A & \aBest{78.0} & 105.9 \\
 & S1, S2 RGB & \aBest{84.0} & 113.1 \\
\midrule
\multirow{2}{*}{S1 GRD (dB)} & S2 L2A & \aBest{2.93} & 3.22 \\
 & S2 L2A, DEM & \aBest{2.85} & 3.22 \\
\midrule
\multirow{4}{*}{S2 L1C} & S1 & \aBest{0.0369} & 0.1015 \\
 & S1, DEM & \aBest{0.0361} & 0.1015 \\
 & S1, S2 L2A & \aBest{0.0112} & 0.0922 \\
 & S1, S2 RGB & \aBest{0.0188} & 0.0926 \\
\midrule
\multirow{3}{*}{S2 L2A} & S1 & \aBest{0.0527} & 0.0730 \\
 & S1, DEM & \aBest{0.0518} & 0.0733 \\
 & S1, S2 RGB & \aBest{0.0251} & 0.0573 \\
\bottomrule
\end{tabular*}
\end{minipage}\hfill
\begin{minipage}[t]{0.485\textwidth}
\centering
\caption{\textbf{EAGLE's fusion beats transformer fusion in all three conditions, including one held out in training,
with approximately 56 times fewer fusion parameters.} Paired over the 4,423 scenes of \cref{tab:fusion_location_ablation}(a); S2 L2A MAE in reflectance.
$\Delta$: relative difference [95\% interval], all $p<0.001$ except $^{\dagger}$ ($p>0.9$); Better: share of scenes
with lower EAGLE error. Bottom: $\Delta$ per target, two or more sensors fused. \textbf{Bold}: best.}
\label{tab:supp_fusion}
\scriptsize
\setlength{\tabcolsep}{2pt}
\renewcommand{\arraystretch}{1.05}
\begin{tabular*}{\linewidth}{@{\extracolsep{\fill}}lcccc@{}}
\toprule
\textbf{Condition} & \cellcolor{A2Eshade}\aOurs & \textbf{Transformer} & \textbf{$\Delta$ [95\% CI]} & \textbf{Better} \\
\midrule
S2$\to$S2 & \cellcolor{A2Eshade}\aBest{0.0151} & 0.0200 & \aScoreCI{$-$24.5\%}{$-$24.8, $-$24.2} & 99.4\% \\
S1/DEM$\to$S2 & \cellcolor{A2Eshade}\aBest{0.0733} & 0.0807 & \aScoreCI{$-$9.1\%}{$-$10.0, $-$8.2} & 69.0\% \\
S1+DEM$\to$S2 & \cellcolor{A2Eshade}\aBest{0.0756} & 0.0862 & \aScoreCI{$-$12.3\%}{$-$13.4, $-$11.2} & 69.5\% \\
\midrule
Fusion parameters & \cellcolor{A2Eshade}5.04\,K & 282\,K & & \\
\midrule
\textbf{Inputs} & \textbf{S2} & \textbf{S1} & \textbf{DEM} & \\
\midrule
S2+S1 & $-$24.5\% & $-$19.0\% & $-$0.1\%$^{\dagger}$ & \\
S1+DEM & $-$12.3\% & $-$19.3\% & $-$47.8\% & \\
All sensors & $-$22.1\% & $-$19.0\% & $-$34.6\% & \\
\bottomrule
\end{tabular*}

\vspace{10pt}
\caption{\textbf{Predicting before sampling beats a single flow, even one of matched size.} CRPS paired over the scenes
of \cref{tab:completion_ablation}(c). B1 + residual flow: two SiT-B/2 networks (311.1\,M), 16k updates each. Single flow:
SiT-B/2 (155.7\,M), 32k updates (top); 24 blocks (308.1\,M), 16k (middle) or 32k updates (bottom). $\Delta$ [95\%
interval], from unrounded values, all $p<0.001$. \textbf{Bold}: best.}
\label{tab:supp_completion}
\scriptsize
\setlength{\tabcolsep}{2pt}
\renewcommand{\arraystretch}{1.05}
\begin{tabular*}{\linewidth}{@{\extracolsep{\fill}}lc>{\columncolor{A2Eshade}}cc@{}}
\toprule
\textbf{Task} & \textbf{Single flow} & \textbf{B1 + residual flow} & \textbf{$\Delta$ [95\% CI]} \\
\midrule
\multicolumn{4}{@{}l}{\textit{Single flow: SiT-B/2, 32k updates}} \\
S1$\to$DEM (m) & 114.9 & \aBest{73.7} & \aScoreCI{$-$35.8\%}{$-$39.0, $-$32.8} \\
S2+S1$\to$DEM (m) & 107.7 & \aBest{67.8} & \aScoreCI{$-$37.0\%}{$-$39.9, $-$34.1} \\
S1$\to$S2 & 0.0467 & \aBest{0.0364} & \aScoreCI{$-$22.1\%}{$-$27.4, $-$16.8} \\
S2$\to$S1 (dB) & 1.96 & \aBest{1.84} & \aScoreCI{$-$6.4\%}{$-$7.9, $-$4.8} \\
\midrule
\multicolumn{4}{@{}l}{\textit{Single flow: 24 blocks (matched parameters), 16k updates}} \\
S1$\to$DEM (m) & 126.2 & \aBest{73.7} & \aScoreCI{$-$41.6\%}{$-$44.5, $-$38.5} \\
S2+S1$\to$DEM (m) & 115.0 & \aBest{67.8} & \aScoreCI{$-$41.1\%}{$-$44.0, $-$38.0} \\
S1$\to$S2 & 0.0474 & \aBest{0.0364} & \aScoreCI{$-$23.3\%}{$-$28.8, $-$17.6} \\
S2$\to$S1 (dB) & 1.94 & \aBest{1.84} & \aScoreCI{$-$5.3\%}{$-$7.1, $-$3.7} \\
\midrule
\multicolumn{4}{@{}l}{\textit{Single flow: 24 blocks (matched parameters), 32k updates}} \\
S1$\to$DEM (m) & 116.5 & \aBest{73.7} & \aScoreCI{$-$36.7\%}{$-$39.6, $-$33.7} \\
S2+S1$\to$DEM (m) & 108.1 & \aBest{67.8} & \aScoreCI{$-$37.3\%}{$-$40.2, $-$34.2} \\
S1$\to$S2 & 0.0442 & \aBest{0.0364} & \aScoreCI{$-$17.7\%}{$-$23.2, $-$12.3} \\
S2$\to$S1 (dB) & 1.89 & \aBest{1.84} & \aScoreCI{$-$3.0\%}{$-$4.5, $-$1.5} \\
\bottomrule
\end{tabular*}

\vspace{12pt}
\caption{\textbf{On held-out scenes, the fitted posterior predicts the average error of every withheld sensor to
within 12\%.} 4,423 evaluation scenes; $A_i,b_i,R_i$ fitted on training scenes; S2: L2A; S1: RTC or GRD. Each
sensor's encoder features are predicted from the others as $A_t\mu_{S,j}+b_t$, with covariance
$V=A_t\Sigma_{S,j}A_t^\top+R_t$; $e$: prediction error; $K=28$. Ratio: realised over predicted mean squared error
(valid cells, scenes weighted equally) [95\% block-bootstrap interval]; it and $e^\top V^{-1}e/K$ are 1 for a
calibrated model.}
\label{tab:supp_calibration}
\scriptsize
\setlength{\tabcolsep}{2pt}
\renewcommand{\arraystretch}{1.05}
\begin{tabular*}{\linewidth}{@{\extracolsep{\fill}}ll>{\columncolor{A2Eshade}}c>{\columncolor{A2Eshade}}c@{}}
\toprule
\textbf{Target} & \textbf{Inputs} & \textbf{Ratio $\to$ 1} & \textbf{$e^\top V^{-1}e/K$ $\to$ 1} \\
\midrule
\multirow{3}{*}{S2 L2A} & S1      & \aScoreCI{0.98}{0.97, 1.00} & 0.99 \\
                        & DEM     & \aScoreCI{0.98}{0.96, 0.99} & 0.99 \\
                        & S1, DEM & \aScoreCI{1.01}{1.00, 1.03} & 1.02 \\
\midrule
\multirow{3}{*}{S1}     & S2      & \aScoreCI{1.00}{0.98, 1.01} & 0.98 \\
                        & DEM     & \aScoreCI{1.02}{1.00, 1.03} & 0.99 \\
                        & S2, DEM & \aScoreCI{1.04}{1.02, 1.06} & 1.02 \\
\midrule
\multirow{3}{*}{DEM}    & S2      & \aScoreCI{1.05}{1.00, 1.09} & 1.03 \\
                        & S1      & \aScoreCI{1.10}{1.04, 1.15} & 1.04 \\
                        & S2, S1  & \aScoreCI{1.12}{1.07, 1.18} & 1.07 \\
\bottomrule
\end{tabular*}
\end{minipage}
\end{table*}

\input{Sup/train_distance}

%% file: Sup/implementation.tex
% ---------------------------------------------------------------------------------------------------------------------
\section{Implementation details}
\label{sup:impl}

\subsection{Data}
\label{sup:data}
We train on 3.35\,M TerraMesh~\cite{terramesh} scenes from MajorTOM and SSL4EO-S12, centre-cropped to $256\times256$ pixels on a common 10\,m grid, to which TerraMesh resamples the 20\,m and 60\,m S2 bands. We correct the inconsistent $+1{,}000$ offset in TerraMesh S2 data per scene, using the same correction for all models' inputs and references. 

\subsection{Architecture and training}
\label{sup:arch}
\label{sup:training}
\Cref{tab:supp_impl} lists the encoder, fusion model, decoders and Stage B networks, and the training settings of both
stages. Stage A first learns the representation with $A_i=I$ and encoder-predicted precisions, then fits
$A_i,b_i,R_i$ with the encoder frozen, and finally adapts the decoders to the fitted fusion. With Stage A frozen,
Stage B trains the predictor B1 and then the flow B2.

\subsection{Computational cost}
\label{sup:cost}
Under matched settings (\cref{tab:cost}), EAGLE needs the least peak GPU memory (3.9\,GB). With the 64 Heun steps
used for all reported results, its time per tile is comparable to COP-GEN's; with 16 steps it is the fastest model,
and its MAE changes by at most 3.1\% (mean of 8 draws, 100 scenes).

\input{Sup/loss}

\input{Sup/cost}

\begin{table*}[t]
\centering
\caption{\textbf{Architecture and training settings.} Batch sizes are global.}
\label{tab:supp_impl}
\scriptsize
\begin{tabular}{@{}p{0.15\textwidth} p{0.81\textwidth}@{}}
\toprule
\multicolumn{2}{@{}l}{\textbf{Architecture}} \\
\midrule
Input &
$256\times256$ pixels on a 10\,m grid \\

Band tokens &
Shared $16\times16$ patch projection ($1\to768$) applied separately to each band, with a band-identity embedding and projected descriptor: sensor group, variable type, product level, polarisation, wavelength, bandwidth and native resolution \\

Band merging &
One learned query per patch attends to the available band tokens through 12 attention heads; absent bands are masked \\

Encoder &
DINOv2 ViT-B/14 blocks on a $16\times16$ patch grid; pretrained $37\times37$ positional embeddings resized by bicubic interpolation; sensor supplied through zero-initialised adaptive LayerNorm \\

Cell features &
$2\times$ pixel shuffle gives $32\times32$ cells, each covering $8\times8$ pixels. A parallel branch embeds bands at each cell and combines them with four learned queries. Output heads predict $K=28$ means and variances per cell \\

Fusion &
\Cref{eq:posterior}, with $A_i,b_i,R_i$ fitted for six sensor variants; 5,040 parameters \\

Scene latent &
$H=[\mu_F;\ell]\in\mathbb{R}^{(K+3)\times32\times32}$, comprising fused features and coordinates \\

Decoders &
Swin-style blocks with pixel-shuffle upsampling to $256\times256$; a separate geocell classification and retrieval head estimates location \\

Stage B &
SiT-L/2 deterministic predictor (B1) and SiT-B/2 residual flow (B2) on the $32\times32$ grid; 64 Heun steps at inference \\

Total &
941\,M parameters \\
\midrule
\multicolumn{2}{@{}l}{\textbf{Stage A}: AdamW, weight decay 0.01, gradient clipping at 1} \\
\midrule
Representation &
144,720 updates; batch 140; $A_i=I$ with encoder-predicted precisions; bf16 and two-step gradient accumulation; learning rate $10^{-4}$ ($10^{-5}$ for DINOv2), with 5\% warmup \\

Sensor fitting &
Encoder frozen; $A_i,b_i,R_i$ fitted by maximising marginal likelihood (\cref{sec:stage_a}) \\

Decoder adaptation &
8,000 updates; batch 100; encoder and fitted $A_i,b_i,R_i$ fixed \\
\midrule
\multicolumn{2}{@{}l}{\textbf{Stage B}: Stage A frozen; AdamW, weight decay 0.01} \\
\midrule
B1 &
\Cref{eq:b1}; 67,549 updates at batch 320; learning rate $10^{-4}$ with 5\% warmup, followed by 4,000 updates at $2\times10^{-5}$; coordinates supplied for 60\% of scenes \\

B2 &
\Cref{eq:flow}; 16,000 updates at batch 260; learning rate $10^{-4}$ with 500 warmup updates; $\lambda_{\mathrm{loc}}=1$; coordinates supplied for 30\% of scenes \\
\bottomrule
\end{tabular}
\end{table*}
% ---------------------------------------------------------------------------------------------------------------------

%% file: Sup/loss.tex
\subsection{Stage A objective}
\label{sup:decoder_loss}
Each step uses the full sensor set $F$ and a subset $S$ sampled for each scene. The objective is
\begin{equation}
\mathcal{L}_{A}
= \mathcal{L}_{\mathrm{var}}(F)
+ 0.5\,\mathcal{L}_{\mathrm{var}}(S)
+ \mathcal{L}_{\mathrm{aux}},
\end{equation}
where $\mathcal{L}_{\mathrm{var}}$ is defined in \cref{eq:stage_a}. We increase $\beta$ linearly to $10^{-4}$ over the first 10\% of updates and average $D_{\mathrm{KL}}$ over cells and channels.
The decoder loss averages the losses from the posterior mean and one sample $z_S\sim q(h\mid x_S)$:
\begin{equation}
\begin{aligned}
\mathcal{L}_{\mathrm{dec}}(S)
&= \tfrac12\,\mathcal{R}_S([\mu_S;\ell])
 + \tfrac12\,\mathcal{R}_S([z_S;\ell]),\\
\mathcal{R}_S(H)
&= \frac{1}{|\mathcal{P}_S|}
   \sum_{i\in\mathcal{P}_S}\rho_i(D_i(H),m_i).
\end{aligned}
\end{equation}
Here, $\mathcal{P}_F$ contains all products in the scene, and $\mathcal{P}_S$ contains those supplied by $S$. Losses use valid pixels only; S2 products with an unresolved offset are excluded. Reconstruction errors use per-band standardised units unless stated otherwise.

\myparagraph{Product losses.}
For S2 L1C and L2A, $\rho_i$ combines per-band L1 with $0.05$ times the spectral angle in reflectance; 20\,m bands are pooled over $2\times2$ pixels before computing L1. S2 RGB and S1 use L1. NDVI uses L1 on the index computed from decoded L2A. DEM uses L1 plus $0.1$ times the L1 slope error, and land cover uses cross-entropy over labelled pixels.

S2 L2A, S2 RGB, NDVI and DEM also use $0.5\,\mathcal{L}_{\nabla}$, computed per band and scene:
\begin{equation}
\begin{aligned}
\mathcal{L}_{\nabla}(\hat m,m)
= 1-\frac{\langle\nabla\hat m,\nabla m\rangle}{\tilde g\,g}
 + a\left|\log\frac{\tilde g}{g}\right|,\\
g = \lVert\nabla m\rVert,\qquad
\tilde g = \sqrt{\lVert\nabla\hat m\rVert^2+(0.1g)^2}.
\end{aligned}
\end{equation}
Here, $\nabla$ denotes horizontal and vertical differences. For DEM, we apply $4\times4$ pooling and set $a=0.5$; otherwise, $a=0$. We omit this term when the reference gradient RMS is below $10^{-3}$ ($10^{-4}$ for DEM) in standardised units.

\myparagraph{Auxiliary losses.}
\begin{equation}
\begin{aligned}
\mathcal{L}_{\mathrm{aux}}
={}& 0.1\,\mathcal{L}_{\times}
+ 0.1\big(\mathcal{L}_{\mathrm{loc}}(F)
+ 0.5\,\mathcal{L}_{\mathrm{loc}}(S)\big)\\
&+ w_{\mathrm{terr}}\,\mathcal{L}_{\mathrm{terr}}.
\end{aligned}
\end{equation}
$\mathcal{L}_{\times}$ is the L1 error at 40\,m for products decoded from $[\mu_S;\ell]$ but absent from $S$: S2 L2A to DEM and S1, and S1 to DEM and S2 L2A. S1--S2 pairs must be at most seven days apart.

$\mathcal{L}_{\mathrm{loc}}(S)$ is the mean great-circle angle between the predicted and true scene centres. $\mathcal{L}_{\mathrm{terr}}$, weighted by $w_{\mathrm{terr}}$, trains an auxiliary head to predict relief from S1 features alone, using L1 plus $0.5\,\mathcal{L}_{\nabla}$. Relief is elevation minus its scene mean, expressed in 100\,m units.

Representation learning uses $A_i=I$ and $w_{\mathrm{terr}}=1$. Decoder adaptation fixes the encoder and fitted $A_i,b_i,R_i$, and sets $w_{\mathrm{terr}}=0$. The relief head is used only during training.

%% file: Sup/cost.tex
% Generated by scripts/paper/build_cost_table.py (results/efficiency). Needs booktabs, colortbl, xcolor[table].
\definecolor{A2Eshade}{HTML}{EDF4F4}
\definecolor{A2Eink}{HTML}{204E56}
\providecommand{\aBest}[1]{\textcolor{A2Eink}{\textbf{#1}}}
\providecommand{\aOurs}{\textcolor{A2Eink}{\textbf{EAGLE}}}
\begin{table}[t]
\centering
\caption{\textbf{EAGLE uses the least GPU memory and is fastest at 16 steps.} One RTX A6000, 8 draws per tile at 192\,px,
median over 10 tiles. Inputs: coordinates, DEM, S1. One output: S2 L2A; three: S2 L1C, L2A and land cover. GB: peak
memory. \textbf{Bold}: best; \underline{underline}: second-best.}
\label{tab:cost}
\scriptsize
\setlength{\tabcolsep}{2pt}
\renewcommand{\arraystretch}{1.05}
\begin{tabular*}{\columnwidth}{@{\extracolsep{\fill}}lccccc@{}}
\toprule
 & & & \multicolumn{2}{c}{\textbf{s / tile}$\downarrow$} & \\
\cmidrule(lr){4-5}
\textbf{Model} & \textbf{Params} & \textbf{Steps} & one output & three outputs & \textbf{GB}$\downarrow$ \\
\midrule
COP-GEN~\cite{espinosa2026copgen} & 1,124\,M & 50 & 4.22 & \underline{4.29} & \underline{5.2} \\
TerraMind-B~\cite{jakubik2026terramind} & 524\,M & 50 & \underline{2.41} & 5.01 & 6.5 \\
TerraMind-L~\cite{jakubik2026terramind} & 1,042\,M & 50 & 2.50 & 5.19 & 12.3 \\
\rowcolor{A2Eshade} \aOurs & 941\,M & 64 & 4.39 & 4.42 & \aBest{3.9} \\
\rowcolor{A2Eshade} \aOurs & 941\,M & 16 & \aBest{1.20} & \aBest{1.22} & \aBest{3.9} \\
\bottomrule
\end{tabular*}
\end{table}

%% file: Sup/eval.tex
\section{Evaluation protocol}
\label{sup:eval}

\subsection{Evaluation sets}
\label{sup:eval_sets}
\begin{figure}[t]
    \centering
    \includegraphics[width=\linewidth]{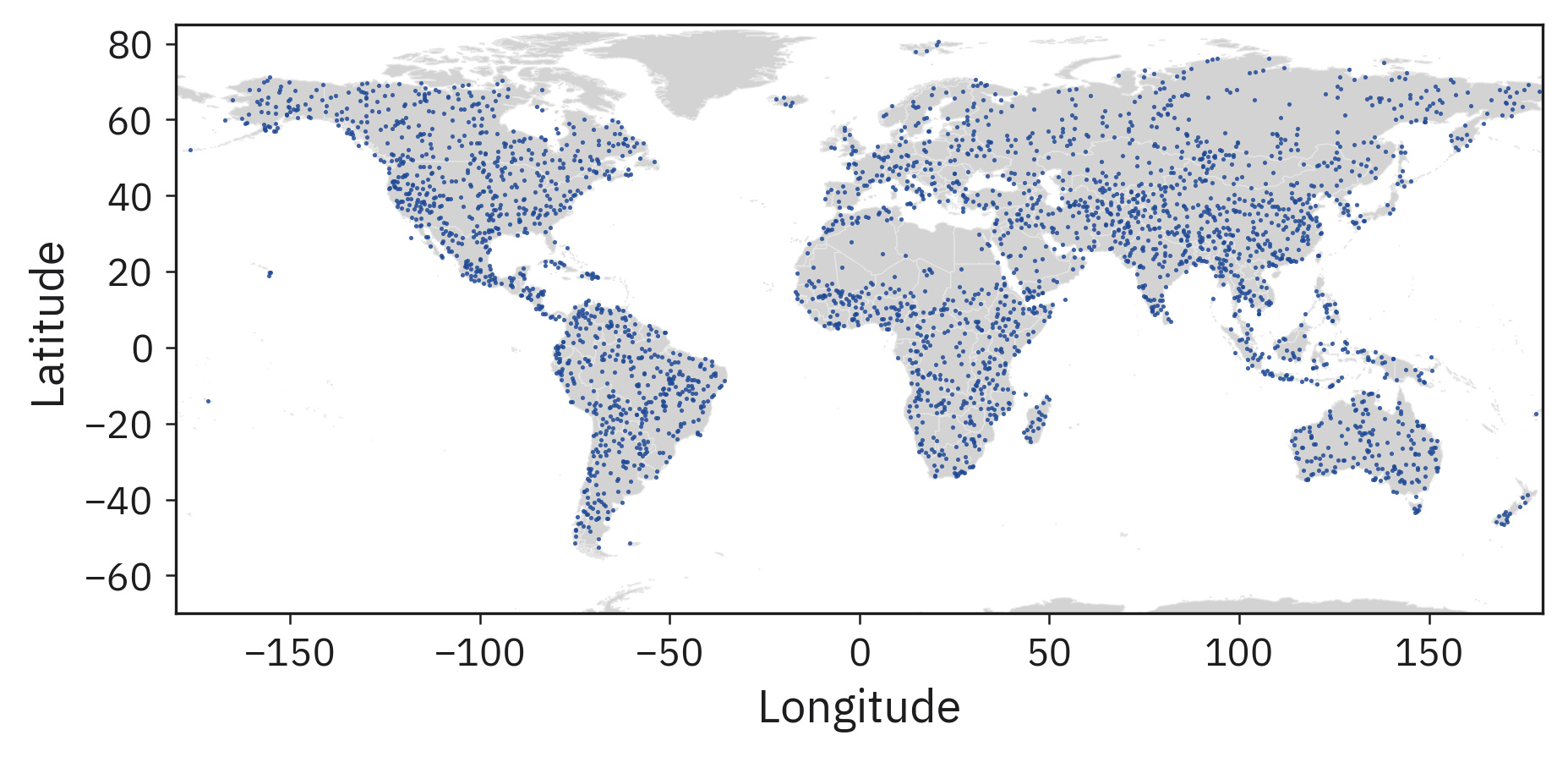}
    \caption{\textbf{The main comparison covers 3,037 MajorTOM locations.}}
    \label{fig:supp_eval_split}
\end{figure}

We use the TerraMesh validation split, whose locations are absent from training. We exclude locations used for development and scenes with unresolved S2 offsets, then retain one scene per location: 3,037 MajorTOM and 1,386 SSL4EO-S12 scenes. These sets were fixed before scoring any model. \Cref{fig:supp_eval_split} shows the MajorTOM locations; \cref{tab:supp_eval_sets} summarises the evaluation settings.

% Keep this table before Tables/best20.

\subsection{Baselines}
\label{sup:baselines}
We use the released COP-GEN and TerraMind checkpoints and inference code listed in \cref{tab:supp_baselines}. In the main and leave-one-out comparisons, both baselines receive land cover unless it is withheld or targeted; COP-GEN also receives cloud masks and dates. EAGLE does not use land-cover inputs. S2 RGB is excluded from these comparisons.
Our baseline scores are close to those reported by COP-GEN~\cite{espinosa2026copgen}, despite the different evaluation scenes.

COP-GEN uses its own MajorTOM split. Its released split lists place 2,309 of our 3,037 evaluation locations in its training set; none of our evaluation locations were used to train EAGLE. EAGLE has lower mean-of-20 MAE on every sensor target in both this group and the remaining 728 locations, with DEM MAE lower by 238\,m and 205\,m, respectively. All paired 95\% confidence intervals for these MAE differences exclude zero.

\input{Sup/implementation_table}

\subsection{Metrics and statistics}
\label{sup:metrics}
\label{sup:stats}
For image generation, we use COP-GEN's metric code on the central $192\times192$ pixels. MAE is measured in reflectance for S2, dB for S1 (clipped to $[-50,10]$) and metres for DEM. PSNR and SSIM use each tile's reference range; constant references are excluded from these two metrics for every model. Land cover uses top-1 accuracy and mIoU, and geolocation uses great-circle distance in kilometres.

Mean over draws scores each draw before averaging. Best of 20 selects the best draw for each scene and metric using the reference. Score of the mean evaluates the image obtained by averaging the draws. EAGLE reports one geocell estimate per scene, so its mean and best geolocation scores coincide. Failed TerraMind coordinate generations (1.9\% of its draws in the main comparison) are excluded.

We obtain 95\% percentile intervals from 2,000 bootstrap resamples of $1^\circ\times1^\circ$ geographic blocks, pairing resamples across models for comparisons. Two-sided bootstrap $p$-values are Holm-corrected across configurations (\cref{tab:context_sweep}) or input additions (\cref{tab:supp_additions}).

%% file: Sup/implementation_table.tex
\begin{table*}[t]
\centering
\caption{\textbf{Evaluation settings and generation models.} Draws per scene; EAGLE's geocell head gives one location
estimate. S2: L2A in the ablations.}
\label{tab:supp_eval_sets}
\label{tab:supp_baselines}
\scriptsize
\setlength{\tabcolsep}{3pt}
\renewcommand{\arraystretch}{1.15}

\begin{tabularx}{\textwidth}{@{}p{0.21\textwidth} X l c c p{0.15\textwidth}@{}}
\toprule
\textbf{Study} & \textbf{Inputs} & \textbf{Data} &
\textbf{Scenes} & \textbf{Draws} & \textbf{Score} \\
\midrule
Main comparison (\cref{tab:loo}(a), \cref{tab:peak_capability}) &
S2 L1C, S2 L2A, S1 RTC, DEM and coordinates; target withheld &
MajorTOM & 3,037 & 20 & mean over draws; best of 20 \\

Leave-one-out (\cref{tab:loo}(b)) &
as above, one further input withheld &
MajorTOM & 500 & 20 & mean over draws; best of 20 \\

Input configurations (\cref{tab:context_sweep}) &
non-empty subsets of other modalities, excluding LULC alone &
MajorTOM & 500 & 8 & mean over draws \\

Uncertainty (\cref{tab:uncertainty}) &
S1, S2 L2A or all other inputs, as listed &
MajorTOM & 500 & 8 & ensemble scores \\

Additional locations (\cref{tab:supp_additional_locations}) &
as listed; S1 GRD replaces RTC &
SSL4EO-S12 & 1,386 & 20 & mean over draws \\

Fusion (\cref{tab:fusion_location_ablation}(a)) &
S2; S2+S1; S1; DEM; S1+DEM &
TerraMesh val. & 4,423 & -- & reconstruction MAE \\

Location input (\cref{tab:fusion_location_ablation}(d)) &
coordinates supplied &
TerraMesh val. & 4,423 & -- & reconstruction MAE \\

Completion, DEM (\cref{tab:completion_ablation}(b,c)) &
S1 or S2+S1; coordinates supplied &
TerraMesh val. & 4,423 & 32 & CRPS \\

Completion, S1/S2 (\cref{tab:completion_ablation}(b,c)) &
S1 to S2 or S2 to S1; coordinates supplied &
TerraMesh val. & 600 & 16 & CRPS \\

Geolocation (\cref{tab:fusion_location_ablation}(e)) &
S2 L1C, S2 L2A, S1 RTC and DEM &
MajorTOM & 3,037 & 20$^\dagger$ & mean and median distance \\

Missing bands, S2 (\cref{tab:missing_bands_published}) &
available S2 bands &
MajorTOM & 500 & 20 & score of the mean \\

Hyperspectral transfer (\cref{tab:hsi}) &
S1, S2 or both to HySpex; HySpex to S2 &
MDAS & test area & 8 & score of the mean \\

Change detection (\cref{tab:flood_cd}) &
S1, S2 or both at each date &
SenForFlood & 357--741 & 8 & pixel-pooled F1 and IoU \\
\midrule
\multicolumn{3}{@{}l}{\textbf{Model and checkpoint}} &
\multicolumn{2}{l}{\textbf{Sampler}} &
\textbf{Steps} \\
\midrule
\multicolumn{3}{@{}l}{COP-GEN: \texttt{cop\_gen\_base}, 500k EMA} &
\multicolumn{2}{l}{DPM-Solver} & 50 \\

\multicolumn{3}{@{}l}{TerraMind-B: \texttt{terramind\_v1\_base\_generate}} &
\multicolumn{2}{l}{diffusion} & 50 \\

\multicolumn{3}{@{}l}{TerraMind-L: \texttt{terramind\_v1\_large\_generate}} &
\multicolumn{2}{l}{diffusion} & 50 \\

\rowcolor{A2Eshade}
\multicolumn{3}{@{}l}{EAGLE: this work} &
\multicolumn{2}{l}{Heun} & 64 \\
\bottomrule
\multicolumn{6}{@{}l}{$^\dagger$Flow-based geolocation only.}
\end{tabularx}
\end{table*}

%% file: Tables/best20.tex
% Requires booktabs, multirow, array and xcolor[table].
\definecolor{A2Eshade}{HTML}{EDF4F4}
\definecolor{A2Eink}{HTML}{204E56}
\providecommand{\aBest}[1]{\textcolor{A2Eink}{\textbf{#1}}}
\providecommand{\aOurs}{\textcolor{A2Eink}{\textbf{EAGLE}}}
\providecommand{\aSd}[1]{\,{\tiny$\pm$#1}}
\providecommand{\aMainCell}[2]{%
  \shortstack[c]{#1\\[-1pt]{\fontsize{6}{6.5}\selectfont #2}}}

\begin{table*}[t]
\centering
\caption{\textbf{EAGLE has the lowest MAE on every sensor target, for the mean and the best of 20 draws.} 3,037 MajorTOM
scenes; inputs as in \cref{tab:loo}(a) ($^{\dagger}$without L2A, $^{\ddagger}$without L1C; $^{*}$L2A from L1C).
$\pm$: SD over draws; brackets: 95\% block-bootstrap intervals. TerraMind: -B. S2 MAE in reflectance, PSNR in dB.
\textbf{Bold}: best; \underline{underline}: second-best.}
\label{tab:peak_capability}
\scriptsize
\setlength{\tabcolsep}{2pt}
\renewcommand{\arraystretch}{1.06}

\begin{tabular*}{\textwidth}{@{\extracolsep{\fill}}ll
    ccc >{\columncolor{A2Eshade}}c
    ccc >{\columncolor{A2Eshade}}c@{}}
\toprule
& & \multicolumn{4}{c}{\textbf{Mean of 20 draws}}
    & \multicolumn{4}{c}{\textbf{Best of 20 (oracle)}} \\
\cmidrule(lr){3-6}\cmidrule(l){7-10}
\textbf{Target} & \textbf{Metric}
& COP-GEN & TerraMind & TerraMind-L & \aOurs{}
& COP-GEN & TerraMind & TerraMind-L & \aOurs{} \\
\midrule

\multirow{2}{*}{\textbf{DEM}}
& MAE (m) $\downarrow$
& \aMainCell{318.2\aSd{5.3}}{[307.4, 328.8]}
& \aMainCell{198.3\aSd{0.5}}{[174.5, 223.3]}
& \aMainCell{\underline{194.1}\aSd{0.5}}{[169.8, 219.5]}
& \aMainCell{\aBest{88.2}\aSd{1.1}}{[84.9, 91.7]}
& \aMainCell{\underline{41.1}}{[38.8, 43.5]}
& \aMainCell{185.2}{[161.4, 209.7]}
& \aMainCell{182.9}{[159.4, 207.7]}
& \aMainCell{\aBest{18.5}}{[16.9, 20.4]} \\
& SSIM $\uparrow$
& \aMainCell{0.239\aSd{0.004}}{[0.232, 0.245]}
& \aMainCell{\underline{0.369}\aSd{0.001}}{[0.357, 0.380]}
& \aMainCell{\textbf{0.386}\aSd{0.001}}{[0.374, 0.398]}
& \aMainCell{0.357\aSd{0.003}}{[0.348, 0.365]}
& \aMainCell{\underline{0.547}}{[0.539, 0.556]}
& \aMainCell{0.433}{[0.420, 0.445]}
& \aMainCell{0.438}{[0.425, 0.450]}
& \aMainCell{\aBest{0.616}}{[0.606, 0.625]} \\
\midrule

\multirow{2}{*}{\textbf{LULC}}
& Top-1 $\uparrow$
& \aMainCell{0.493\aSd{0.005}}{[0.486, 0.500]}
& \aMainCell{\underline{0.889}\aSd{0.0003}}{[0.883, 0.894]}
& \aMainCell{\textbf{0.894}\aSd{0.0002}}{[0.889, 0.899]}
& \aMainCell{0.881\aSd{0.001}}{[0.875, 0.887]}
& \aMainCell{0.801}{[0.795, 0.808]}
& \aMainCell{\underline{0.905}}{[0.900, 0.910]}
& \aMainCell{\textbf{0.910}}{[0.905, 0.914]}
& \aMainCell{0.901}{[0.896, 0.906]} \\
& mIoU $\uparrow$
& \aMainCell{0.167\aSd{0.003}}{[0.163, 0.170]}
& \aMainCell{0.559\aSd{0.002}}{[0.550, 0.568]}
& \aMainCell{\underline{0.577}\aSd{0.002}}{[0.568, 0.586]}
& \aMainCell{\aBest{0.579}\aSd{0.001}}{[0.569, 0.588]}
& \aMainCell{0.359}{[0.352, 0.365]}
& \aMainCell{\underline{0.635}}{[0.626, 0.644]}
& \aMainCell{\textbf{0.647}}{[0.638, 0.656]}
& \aMainCell{0.614}{[0.605, 0.623]} \\
\midrule

\multirow{2}{*}{\textbf{S1 RTC}}
& MAE (dB) $\downarrow$
& \aMainCell{3.65\aSd{0.02}}{[3.61, 3.70]}
& \aMainCell{2.84\aSd{0.003}}{[2.79, 2.89]}
& \aMainCell{\underline{2.82}\aSd{0.001}}{[2.77, 2.87]}
& \aMainCell{\aBest{2.36}\aSd{0.005}}{[2.33, 2.38]}
& \aMainCell{\underline{2.72}}{[2.70, 2.75]}
& \aMainCell{2.73}{[2.69, 2.78]}
& \aMainCell{2.73}{[2.68, 2.78]}
& \aMainCell{\aBest{2.09}}{[2.07, 2.10]} \\
& PSNR $\uparrow$
& \aMainCell{14.47\aSd{0.03}}{[14.39, 14.54]}
& \aMainCell{16.57\aSd{0.003}}{[16.48, 16.66]}
& \aMainCell{\underline{16.63}\aSd{0.002}}{[16.54, 16.72]}
& \aMainCell{\aBest{17.80}\aSd{0.01}}{[17.74, 17.86]}
& \aMainCell{16.42}{[16.36, 16.47]}
& \aMainCell{16.82}{[16.73, 16.91]}
& \aMainCell{\underline{16.87}}{[16.78, 16.95]}
& \aMainCell{\aBest{18.70}}{[18.64, 18.75]} \\
\midrule

\multirow{2}{*}{\textbf{S2 L1C}}
& MAE $\downarrow$
& \aMainCell{\underline{0.0627}\aSd{0.0006}}{[0.0619, 0.0635]}
& \aMainCell{0.1083\aSd{0.0001}}{[0.1042, 0.1126]}
& \aMainCell{0.1085\aSd{0.00003}}{[0.1043, 0.1128]}
& \aMainCell{\aBest{0.0140}\aSd{0.00005}}{[0.0136, 0.0144]}
& \aMainCell{\underline{0.0181}}{[0.0175, 0.0187]}
& \aMainCell{0.1041}{[0.0999, 0.1083]}
& \aMainCell{0.1043}{[0.1001, 0.1085]}
& \aMainCell{\aBest{0.0115}}{[0.0112, 0.0119]} \\
& PSNR $\uparrow$
& \aMainCell{\underline{11.88}\aSd{0.07}}{[11.75, 12.02]}
& \aMainCell{9.97\aSd{0.004}}{[9.88, 10.06]}
& \aMainCell{9.95\aSd{0.003}}{[9.86, 10.04]}
& \aMainCell{\aBest{20.04}\aSd{0.02}}{[19.90, 20.19]}
& \aMainCell{\underline{19.35}}{[19.22, 19.48]}
& \aMainCell{10.30}{[10.21, 10.39]}
& \aMainCell{10.28}{[10.18, 10.37]}
& \aMainCell{\aBest{22.37}}{[22.23, 22.51]} \\
\midrule

\multirow{2}{*}{\textbf{S2 L1C$^{\dagger}$}}
& MAE $\downarrow$
& \aMainCell{0.1312\aSd{0.0009}}{[0.1288, 0.1338]}
& \aMainCell{0.1222\aSd{0.0001}}{[0.1181, 0.1263]}
& \aMainCell{\underline{0.1221}\aSd{0.0001}}{[0.1182, 0.1262]}
& \aMainCell{\aBest{0.0405}\aSd{0.0002}}{[0.0391, 0.0418]}
& \aMainCell{\underline{0.0558}}{[0.0541, 0.0575]}
& \aMainCell{0.1134}{[0.1094, 0.1175]}
& \aMainCell{0.1131}{[0.1093, 0.1172]}
& \aMainCell{\aBest{0.0311}}{[0.0300, 0.0322]} \\
& PSNR $\uparrow$
& \aMainCell{7.23\aSd{0.04}}{[7.16, 7.30]}
& \aMainCell{\underline{8.70}\aSd{0.01}}{[8.61, 8.79]}
& \aMainCell{8.69\aSd{0.01}}{[8.60, 8.78]}
& \aMainCell{\aBest{13.65}\aSd{0.03}}{[13.54, 13.75]}
& \aMainCell{\underline{11.61}}{[11.51, 11.71]}
& \aMainCell{9.23}{[9.13, 9.32]}
& \aMainCell{9.22}{[9.13, 9.32]}
& \aMainCell{\aBest{15.74}}{[15.63, 15.83]} \\
\midrule

\multirow{2}{*}{\textbf{S2 L2A$^{*}$}}
& MAE $\downarrow$
& \aMainCell{\underline{0.0376}\aSd{0.0006}}{[0.0366, 0.0385]}
& \aMainCell{0.0957\aSd{0.00004}}{[0.0906, 0.1011]}
& \aMainCell{0.0961\aSd{0.0001}}{[0.0911, 0.1014]}
& \aMainCell{\aBest{0.0243}\aSd{0.0001}}{[0.0233, 0.0254]}
& \aMainCell{\underline{0.0220}}{[0.0212, 0.0226]}
& \aMainCell{0.0908}{[0.0856, 0.0961]}
& \aMainCell{0.0912}{[0.0861, 0.0965]}
& \aMainCell{\aBest{0.0204}}{[0.0195, 0.0214]} \\
& PSNR $\uparrow$
& \aMainCell{\underline{17.37}\aSd{0.08}}{[17.23, 17.50]}
& \aMainCell{12.59\aSd{0.01}}{[12.43, 12.73]}
& \aMainCell{12.54\aSd{0.01}}{[12.39, 12.69]}
& \aMainCell{\aBest{20.21}\aSd{0.02}}{[20.04, 20.37]}
& \aMainCell{\underline{20.46}}{[20.35, 20.57]}
& \aMainCell{13.41}{[13.25, 13.58]}
& \aMainCell{13.34}{[13.18, 13.50]}
& \aMainCell{\aBest{21.79}}{[21.62, 21.95]} \\
\midrule

\multirow{2}{*}{\textbf{S2 L2A$^{\ddagger}$}}
& MAE $\downarrow$
& \aMainCell{0.1500\aSd{0.0011}}{[0.1469, 0.1535]}
& \aMainCell{0.1173\aSd{0.0001}}{[0.1123, 0.1225]}
& \aMainCell{\underline{0.1166}\aSd{0.0001}}{[0.1118, 0.1217]}
& \aMainCell{\aBest{0.0527}\aSd{0.0002}}{[0.0510, 0.0543]}
& \aMainCell{\underline{0.0667}}{[0.0647, 0.0687]}
& \aMainCell{0.1077}{[0.1025, 0.1127]}
& \aMainCell{0.1069}{[0.1020, 0.1119]}
& \aMainCell{\aBest{0.0412}}{[0.0398, 0.0427]} \\
& PSNR $\uparrow$
& \aMainCell{8.11\aSd{0.04}}{[8.03, 8.20]}
& \aMainCell{9.89\aSd{0.01}}{[9.76, 10.02]}
& \aMainCell{\underline{9.91}\aSd{0.01}}{[9.78, 10.03]}
& \aMainCell{\aBest{14.19}\aSd{0.02}}{[14.08, 14.30]}
& \aMainCell{\underline{12.48}}{[12.37, 12.59]}
& \aMainCell{10.81}{[10.66, 10.95]}
& \aMainCell{10.84}{[10.70, 10.99]}
& \aMainCell{\aBest{16.06}}{[15.95, 16.15]} \\
\midrule

\textbf{Location} & Mean km $\downarrow$
& \aMainCell{1188\aSd{28}}{[1116, 1265]}
& \aMainCell{\textbf{623}\aSd{24}}{[566, 684]}
& \aMainCell{\underline{732}\aSd{23}}{[671, 794]}
& \aMainCell{866\aSd{0}}{[773, 957]}
& \aMainCell{243}{[214, 273]}
& \aMainCell{\underline{153}}{[126, 182]}
& \aMainCell{\textbf{149}}{[128, 174]}
& \aMainCell{866}{[773, 957]} \\

\bottomrule
\end{tabular*}
\end{table*}

%% file: Tables/context_sweep.tex
% Generated by scripts/paper/build_sweep500_tex.py from results/peak_val_v2/sweep500/{sweep_table,monotone}.csv.
\begin{table*}[t]
\centering
\begin{minipage}[t]{0.485\textwidth}
\centering
\caption{\textbf{EAGLE is best in 148 of 181 configurations with non-coordinate targets, and no added input makes it significantly
worse (\cref{tab:supp_additions}).} 211 input configurations (every non-empty subset of the other modalities, except LULC alone), 500 MajorTOM
scenes, mean of 8 draws. $n$: configurations; \emph{Best}: EAGLE beats both baselines; \emph{Better / Worse}:
significantly better / worse than the better baseline (paired block bootstrap, Holm). Per configuration:
\crefrange{tab:full-dem}{tab:full-geolocation}.}
\label{tab:context_sweep}
\scriptsize
\setlength{\tabcolsep}{2pt}
\renewcommand{\arraystretch}{1.05}
\begin{tabular*}{\linewidth}{@{\extracolsep{\fill}}llcccc@{}}
\toprule
\textbf{Target} & \textbf{Metric} & $n$ & \textbf{Best} & \textbf{Better} & \textbf{Worse} \\
\midrule
DEM & MAE & 30 & 16 & 16 & 4 \\
S1 & MAE & 30 & 28 & 26 & 2 \\
S2 L1C & PSNR & 30 & 29 & 29 & 0 \\
S2 L2A & PSNR & 30 & 29 & 29 & 1 \\
NDVI & MAE & 30 & 28 & 28 & 2 \\
LULC & mIoU & 31 & 18 & 6 & 1 \\
Location & km & 30 & 1 & 0 & 25 \\
\midrule
\textit{Non-coordinate targets} &  & 181 & 148 & 134 & 10 \\
\quad $\geq$1 sensor input &  & 170 & 143 & 129 & 5 \\
\bottomrule
\end{tabular*}
\end{minipage}
\end{table*}

%% file: Sup/fullcontext.tex
% Generated by scripts/paper/build_sweep500_tex.py from results/peak_val_v2/sweep500/sweep_table.csv (do not edit
% numbers by hand). Needs \usepackage{booktabs,array,colortbl,pifont} and \usepackage[table]{xcolor}.
\definecolor{A2Eshade}{HTML}{EDF4F4}
\definecolor{A2Eink}{HTML}{204E56}
\definecolor{A2Emuted}{HTML}{B0B0B0}
\providecommand{\aYes}{\ding{51}}
\providecommand{\aNo}{\textcolor{A2Emuted}{\ding{55}}}
\providecommand{\aBest}[1]{\textcolor{A2Eink}{\textbf{#1}}}
\providecommand{\aOurs}{\textcolor{A2Eink}{\textbf{EAGLE}}}
\providecommand{\aSweepSetup}{\scriptsize\setlength{\tabcolsep}{2pt}\renewcommand{\arraystretch}{0.88}}
\providecommand{\aSweepNote}[1]{\par\vspace{4pt}\begin{minipage}{\textwidth}\scriptsize\raggedright
  \aYes/\aNo: input given / absent. Bold: best per row (teal: EAGLE); underline: second-best. #1\end{minipage}}
\providecommand{\aSweepEval}{500 MajorTOM scenes, 8 draws; best of 8 selected with the reference. Small numbers: SD
  over draws. TerraMind: -B.}
\providecommand{\aSd}[1]{\,{\tiny$\pm$#1}}
\providecommand{\aSweepLULC}{$^{\ddagger}$Baselines only; EAGLE does not take LULC, so its score is that without LULC.}

% ============================================================
% DEM generation
% ============================================================
\begin{table*}[t]
\centering
\caption{\textbf{EAGLE has the lowest DEM MAE in 16 of 30 configurations for the mean of 8 draws, and in 27 for the best of 8.}
MAE $\downarrow$ and SSIM $\uparrow$. \aSweepEval}
\label{tab:full-dem}
\begingroup
\aSweepSetup
\resizebox{\textwidth}{!}{%
\begin{tabular}{@{}ccccc rr>{\columncolor{A2Eshade}}r rr>{\columncolor{A2Eshade}}r rr>{\columncolor{A2Eshade}}r rr>{\columncolor{A2Eshade}}r@{}}
\toprule
\multicolumn{5}{c}{\textbf{Available inputs}} & \multicolumn{6}{c}{\textbf{Mean of 8 draws}} & \multicolumn{6}{c}{\textbf{Best of 8 draws (oracle)}} \\
\cmidrule(lr){1-5}\cmidrule(lr){6-11}\cmidrule(l){12-17}
& & & & & \multicolumn{3}{c}{MAE $\downarrow$} & \multicolumn{3}{c}{SSIM $\uparrow$} & \multicolumn{3}{c}{MAE $\downarrow$} & \multicolumn{3}{c}{SSIM $\uparrow$} \\
\cmidrule(l){6-8}\cmidrule(l){9-11}\cmidrule(l){12-14}\cmidrule(l){15-17}
Coord. & LULC$^{\ddagger}$ & S1 & L1C & L2A & COP-GEN & TerraMind & \aOurs & COP-GEN & TerraMind & \aOurs & COP-GEN & TerraMind & \aOurs & COP-GEN & TerraMind & \aOurs \\
\midrule
\aYes & \aYes & \aYes & \aYes & \aYes & 317.3\aSd{9.3} & \underline{183.3}\aSd{1.0} & \aBest{88.8}\aSd{2.6} & 0.241\aSd{0.009} & \textbf{0.374}\aSd{0.004} & \underline{0.365}\aSd{0.006} & \underline{75.8} & 172.9 & \aBest{27.2} & \underline{0.495} & 0.429 & \aBest{0.571} \\
\addlinespace[1pt]
\aYes & \aYes & \aYes & \aYes & \aNo  & 338.4\aSd{12.1} & \underline{194.3}\aSd{1.1} & \aBest{97.2}\aSd{2.3} & 0.237\aSd{0.008} & \underline{0.367}\aSd{0.004} & \aBest{0.376}\aSd{0.010} & \underline{84.6} & 181.1 & \aBest{32.1} & \underline{0.493} & 0.417 & \aBest{0.574} \\
\aYes & \aYes & \aYes & \aNo  & \aYes & 445.8\aSd{18.7} & \underline{192.2}\aSd{1.1} & \aBest{91.4}\aSd{3.3} & 0.234\aSd{0.006} & \textbf{0.368}\aSd{0.004} & \underline{0.358}\aSd{0.010} & \underline{98.3} & 179.0 & \aBest{29.1} & \underline{0.495} & 0.420 & \aBest{0.573} \\
\aYes & \aYes & \aNo  & \aYes & \aYes & 345.9\aSd{9.7} & \underline{184.5}\aSd{0.8} & \aBest{95.6}\aSd{2.2} & 0.258\aSd{0.010} & \textbf{0.367}\aSd{0.002} & \underline{0.349}\aSd{0.008} & \underline{95.4} & 174.4 & \aBest{30.1} & \underline{0.490} & 0.423 & \aBest{0.547} \\
\aYes & \aNo  & \aYes & \aYes & \aYes & 318.0\aSd{7.2} & \underline{183.8}\aSd{0.7} & \aBest{88.8}\aSd{2.6} & 0.230\aSd{0.006} & \textbf{0.375}\aSd{0.003} & \underline{0.365}\aSd{0.006} & \underline{77.4} & 173.7 & \aBest{27.2} & \underline{0.498} & 0.428 & \aBest{0.571} \\
\aNo  & \aYes & \aYes & \aYes & \aYes & 330.2\aSd{14.4} & \textbf{184.9}\aSd{0.7} & \underline{222.0}\aSd{9.8} & 0.241\aSd{0.006} & \textbf{0.381}\aSd{0.004} & \underline{0.304}\aSd{0.006} & \underline{83.8} & 173.7 & \aBest{60.9} & \underline{0.495} & 0.434 & \aBest{0.539} \\
\addlinespace[1pt]
\aYes & \aYes & \aYes & \aNo  & \aNo  & 619.8\aSd{35.2} & \underline{241.5}\aSd{2.2} & \aBest{106.4}\aSd{3.9} & 0.219\aSd{0.011} & \underline{0.333}\aSd{0.002} & \aBest{0.351}\aSd{0.010} & \underline{137.4} & 219.6 & \aBest{31.9} & \underline{0.486} & 0.384 & \aBest{0.562} \\
\aYes & \aYes & \aNo  & \aYes & \aNo  & 377.3\aSd{21.6} & \underline{193.7}\aSd{0.8} & \aBest{116.2}\aSd{3.7} & 0.257\aSd{0.008} & \textbf{0.364}\aSd{0.002} & \underline{0.324}\aSd{0.009} & \underline{94.7} & 180.1 & \aBest{46.2} & \underline{0.488} & 0.418 & \aBest{0.505} \\
\aYes & \aYes & \aNo  & \aNo  & \aYes & 524.8\aSd{20.4} & \underline{190.4}\aSd{1.1} & \aBest{100.6}\aSd{1.9} & 0.249\aSd{0.009} & \textbf{0.367}\aSd{0.003} & \underline{0.333}\aSd{0.013} & \underline{115.6} & 176.9 & \aBest{31.3} & \underline{0.486} & 0.421 & \aBest{0.545} \\
\aYes & \aNo  & \aYes & \aYes & \aNo  & 328.8\aSd{16.7} & \underline{192.2}\aSd{1.0} & \aBest{97.2}\aSd{2.3} & 0.241\aSd{0.007} & \underline{0.373}\aSd{0.005} & \aBest{0.376}\aSd{0.010} & \underline{77.0} & 179.3 & \aBest{32.1} & \underline{0.504} & 0.423 & \aBest{0.574} \\
\aYes & \aNo  & \aYes & \aNo  & \aYes & 456.4\aSd{17.5} & \underline{193.4}\aSd{0.8} & \aBest{91.4}\aSd{3.3} & 0.226\aSd{0.011} & \textbf{0.377}\aSd{0.002} & \underline{0.358}\aSd{0.010} & \underline{108.0} & 181.1 & \aBest{29.1} & \underline{0.484} & 0.426 & \aBest{0.573} \\
\aYes & \aNo  & \aNo  & \aYes & \aYes & 340.5\aSd{15.7} & \underline{185.9}\aSd{1.0} & \aBest{95.6}\aSd{2.2} & 0.257\aSd{0.010} & \textbf{0.368}\aSd{0.004} & \underline{0.349}\aSd{0.008} & \underline{86.9} & 175.3 & \aBest{30.1} & \underline{0.485} & 0.425 & \aBest{0.547} \\
\aNo  & \aYes & \aYes & \aYes & \aNo  & 354.0\aSd{14.8} & \textbf{198.4}\aSd{1.1} & \underline{264.5}\aSd{13.3} & 0.234\aSd{0.009} & \textbf{0.366}\aSd{0.002} & \underline{0.310}\aSd{0.007} & \underline{79.3} & 184.2 & \aBest{74.5} & \underline{0.486} & 0.417 & \aBest{0.534} \\
\aNo  & \aYes & \aYes & \aNo  & \aYes & 480.5\aSd{34.5} & \textbf{198.1}\aSd{0.5} & \underline{233.2}\aSd{14.5} & 0.226\aSd{0.012} & \textbf{0.366}\aSd{0.002} & \underline{0.294}\aSd{0.011} & \underline{106.7} & 183.9 & \aBest{66.1} & \underline{0.491} & 0.417 & \aBest{0.531} \\
\aNo  & \aYes & \aNo  & \aYes & \aYes & 355.4\aSd{16.2} & \textbf{186.4}\aSd{0.5} & \underline{247.6}\aSd{14.5} & 0.262\aSd{0.013} & \textbf{0.374}\aSd{0.003} & \underline{0.289}\aSd{0.009} & \underline{93.8} & 175.9 & \aBest{60.4} & \underline{0.489} & 0.426 & \aBest{0.521} \\
\aNo  & \aNo  & \aYes & \aYes & \aYes & 324.1\aSd{14.6} & \textbf{185.6}\aSd{0.7} & \underline{222.0}\aSd{9.8} & 0.237\aSd{0.010} & \textbf{0.384}\aSd{0.002} & \underline{0.304}\aSd{0.006} & \underline{78.2} & 173.9 & \aBest{60.9} & \underline{0.503} & 0.434 & \aBest{0.539} \\
\addlinespace[1pt]
\aYes & \aYes & \aNo  & \aNo  & \aNo  & 903.6\aSd{28.0} & \underline{332.6}\aSd{3.1} & \aBest{235.1}\aSd{2.3} & \textbf{0.264}\aSd{0.009} & \underline{0.235}\aSd{0.005} & 0.212\aSd{0.004} & \textbf{182.8} & 294.0 & \underline{185.7} & \textbf{0.471} & 0.285 & \underline{0.290} \\
\aYes & \aNo  & \aYes & \aNo  & \aNo  & 737.8\aSd{51.9} & \underline{243.8}\aSd{2.2} & \aBest{106.4}\aSd{3.9} & 0.205\aSd{0.011} & \underline{0.329}\aSd{0.004} & \aBest{0.351}\aSd{0.010} & \underline{151.1} & 217.0 & \aBest{31.9} & \underline{0.466} & 0.397 & \aBest{0.562} \\
\aYes & \aNo  & \aNo  & \aYes & \aNo  & 363.2\aSd{7.5} & \underline{194.3}\aSd{0.9} & \aBest{116.2}\aSd{3.7} & 0.242\aSd{0.007} & \textbf{0.376}\aSd{0.003} & \underline{0.324}\aSd{0.009} & \underline{93.7} & 181.4 & \aBest{46.2} & \underline{0.482} & 0.428 & \aBest{0.505} \\
\aYes & \aNo  & \aNo  & \aNo  & \aYes & 524.0\aSd{21.1} & \underline{194.3}\aSd{1.0} & \aBest{100.6}\aSd{1.9} & 0.239\aSd{0.011} & \textbf{0.375}\aSd{0.003} & \underline{0.333}\aSd{0.013} & \underline{115.8} & 181.3 & \aBest{31.3} & \underline{0.487} & 0.431 & \aBest{0.545} \\
\aNo  & \aYes & \aYes & \aNo  & \aNo  & 822.3\aSd{66.4} & \textbf{292.2}\aSd{2.5} & \underline{403.1}\aSd{11.4} & 0.215\aSd{0.010} & \textbf{0.306}\aSd{0.003} & \underline{0.274}\aSd{0.006} & \underline{179.9} & 260.9 & \aBest{112.6} & \underline{0.470} & 0.361 & \aBest{0.508} \\
\aNo  & \aYes & \aNo  & \aYes & \aNo  & 398.1\aSd{13.1} & \textbf{198.0}\aSd{0.7} & \underline{356.9}\aSd{16.7} & 0.252\aSd{0.005} & \textbf{0.365}\aSd{0.004} & \underline{0.267}\aSd{0.006} & \textbf{93.7} & 183.0 & \underline{115.2} & \textbf{0.484} & \underline{0.416} & \aBest{0.484} \\
\aNo  & \aYes & \aNo  & \aNo  & \aYes & 601.2\aSd{23.8} & \textbf{196.7}\aSd{1.6} & \underline{284.6}\aSd{19.6} & 0.250\aSd{0.012} & \textbf{0.364}\aSd{0.002} & \underline{0.275}\aSd{0.008} & \underline{117.2} & 182.7 & \aBest{74.3} & \underline{0.483} & 0.419 & \aBest{0.506} \\
\aNo  & \aNo  & \aYes & \aYes & \aNo  & 355.5\aSd{10.5} & \textbf{196.3}\aSd{1.0} & \underline{264.5}\aSd{13.3} & 0.236\aSd{0.011} & \textbf{0.370}\aSd{0.003} & \underline{0.310}\aSd{0.007} & \underline{84.1} & 181.6 & \aBest{74.5} & \underline{0.495} & 0.427 & \aBest{0.534} \\
\aNo  & \aNo  & \aYes & \aNo  & \aYes & 483.0\aSd{36.0} & \textbf{197.2}\aSd{0.5} & \underline{233.2}\aSd{14.5} & 0.229\aSd{0.009} & \textbf{0.370}\aSd{0.004} & \underline{0.294}\aSd{0.011} & \underline{102.5} & 183.5 & \aBest{66.1} & \underline{0.496} & 0.419 & \aBest{0.531} \\
\aNo  & \aNo  & \aNo  & \aYes & \aYes & 361.3\aSd{12.3} & \textbf{186.5}\aSd{0.3} & \underline{247.6}\aSd{14.5} & 0.254\aSd{0.007} & \textbf{0.379}\aSd{0.003} & \underline{0.289}\aSd{0.009} & \underline{90.4} & 176.1 & \aBest{60.4} & \underline{0.490} & 0.431 & \aBest{0.521} \\
\addlinespace[1pt]
\aYes & \aNo  & \aNo  & \aNo  & \aNo  & 1091.1\aSd{28.8} & \underline{590.6}\aSd{1.0} & \aBest{235.1}\aSd{2.3} & \underline{0.204}\aSd{0.011} & 0.148\aSd{0.001} & \aBest{0.212}\aSd{0.004} & \underline{219.7} & 564.6 & \aBest{185.7} & \textbf{0.460} & 0.167 & \underline{0.290} \\
\aNo  & \aNo  & \aYes & \aNo  & \aNo  & 926.1\aSd{52.8} & \textbf{289.9}\aSd{2.7} & \underline{403.1}\aSd{11.4} & 0.204\aSd{0.011} & \textbf{0.320}\aSd{0.004} & \underline{0.274}\aSd{0.006} & \underline{199.5} & 254.5 & \aBest{112.6} & \underline{0.467} & 0.377 & \aBest{0.508} \\
\aNo  & \aNo  & \aNo  & \aYes & \aNo  & 394.6\aSd{14.8} & \textbf{197.8}\aSd{1.3} & \underline{356.9}\aSd{16.7} & 0.246\aSd{0.011} & \textbf{0.369}\aSd{0.002} & \underline{0.267}\aSd{0.006} & \textbf{103.1} & 183.4 & \underline{115.2} & \underline{0.475} & 0.420 & \aBest{0.484} \\
\aNo  & \aNo  & \aNo  & \aNo  & \aYes & 624.1\aSd{29.3} & \textbf{195.2}\aSd{1.4} & \underline{284.6}\aSd{19.6} & 0.237\aSd{0.010} & \textbf{0.372}\aSd{0.003} & \underline{0.275}\aSd{0.008} & \underline{120.8} & 182.2 & \aBest{74.3} & \underline{0.477} & 0.428 & \aBest{0.506} \\
\bottomrule
\end{tabular}}
\aSweepNote{\aSweepLULC}
\endgroup
\end{table*}

% ============================================================
% Sentinel-1 RTC generation
% ============================================================
\begin{table*}[t]
\centering
\caption{\textbf{EAGLE has the lowest mean-of-8 S1 RTC MAE in 28 of 30 configurations.}
MAE $\downarrow$ and PSNR $\uparrow$. \aSweepEval}
\label{tab:full-s1rtc}
\begingroup
\aSweepSetup
\resizebox{\textwidth}{!}{%
\begin{tabular}{@{}ccccc rr>{\columncolor{A2Eshade}}r rr>{\columncolor{A2Eshade}}r rr>{\columncolor{A2Eshade}}r rr>{\columncolor{A2Eshade}}r@{}}
\toprule
\multicolumn{5}{c}{\textbf{Available inputs}} & \multicolumn{6}{c}{\textbf{Mean of 8 draws}} & \multicolumn{6}{c}{\textbf{Best of 8 draws (oracle)}} \\
\cmidrule(lr){1-5}\cmidrule(lr){6-11}\cmidrule(l){12-17}
& & & & & \multicolumn{3}{c}{MAE $\downarrow$} & \multicolumn{3}{c}{PSNR $\uparrow$} & \multicolumn{3}{c}{MAE $\downarrow$} & \multicolumn{3}{c}{PSNR $\uparrow$} \\
\cmidrule(l){6-8}\cmidrule(l){9-11}\cmidrule(l){12-14}\cmidrule(l){15-17}
Coord. & LULC$^{\ddagger}$ & DEM & L1C & L2A & COP-GEN & TerraMind & \aOurs & COP-GEN & TerraMind & \aOurs & COP-GEN & TerraMind & \aOurs & COP-GEN & TerraMind & \aOurs \\
\midrule
\aYes & \aYes & \aYes & \aYes & \aYes & 3.65\aSd{0.05} & \underline{2.83}\aSd{0.01} & \aBest{2.36}\aSd{0.02} & 14.41\aSd{0.06} & \underline{16.58}\aSd{0.01} & \aBest{17.74}\aSd{0.05} & 2.85 & \underline{2.74} & \aBest{2.13} & 16.02 & \underline{16.79} & \aBest{18.49} \\
\addlinespace[1pt]
\aYes & \aYes & \aYes & \aYes & \aNo  & 3.60\aSd{0.04} & \underline{2.82}\aSd{0.01} & \aBest{2.38}\aSd{0.01} & 14.59\aSd{0.07} & \underline{16.59}\aSd{0.01} & \aBest{17.71}\aSd{0.02} & 2.80 & \underline{2.73} & \aBest{2.14} & 16.14 & \underline{16.80} & \aBest{18.48} \\
\aYes & \aYes & \aYes & \aNo  & \aYes & 3.59\aSd{0.05} & \underline{2.82}\aSd{0.00} & \aBest{2.36}\aSd{0.02} & 14.51\aSd{0.06} & \underline{16.59}\aSd{0.01} & \aBest{17.74}\aSd{0.05} & 2.84 & \underline{2.74} & \aBest{2.12} & 16.04 & \underline{16.79} & \aBest{18.53} \\
\aYes & \aYes & \aNo  & \aYes & \aYes & 3.80\aSd{0.02} & \underline{2.82}\aSd{0.01} & \aBest{2.42}\aSd{0.02} & 14.07\aSd{0.02} & \underline{16.59}\aSd{0.01} & \aBest{17.57}\aSd{0.05} & 2.95 & \underline{2.73} & \aBest{2.14} & 15.74 & \underline{16.79} & \aBest{18.44} \\
\aYes & \aNo  & \aYes & \aYes & \aYes & 3.64\aSd{0.04} & \underline{2.82}\aSd{0.00} & \aBest{2.36}\aSd{0.02} & 14.33\aSd{0.04} & \underline{16.59}\aSd{0.01} & \aBest{17.74}\aSd{0.05} & 2.88 & \underline{2.74} & \aBest{2.13} & 15.92 & \underline{16.79} & \aBest{18.49} \\
\aNo  & \aYes & \aYes & \aYes & \aYes & 3.61\aSd{0.06} & \underline{2.83}\aSd{0.01} & \aBest{2.42}\aSd{0.01} & 14.48\aSd{0.04} & \underline{16.58}\aSd{0.01} & \aBest{17.55}\aSd{0.03} & 2.85 & \underline{2.74} & \aBest{2.14} & 16.03 & \underline{16.79} & \aBest{18.46} \\
\addlinespace[1pt]
\aYes & \aYes & \aYes & \aNo  & \aNo  & 4.23\aSd{0.05} & \underline{3.01}\aSd{0.01} & \aBest{2.61}\aSd{0.01} & 13.15\aSd{0.03} & \underline{15.88}\aSd{0.01} & \aBest{16.97}\aSd{0.05} & 3.16 & \underline{2.90} & \aBest{2.28} & 15.09 & \underline{16.16} & \aBest{17.93} \\
\aYes & \aYes & \aNo  & \aYes & \aNo  & 3.77\aSd{0.06} & \underline{2.81}\aSd{0.00} & \aBest{2.50}\aSd{0.01} & 14.19\aSd{0.07} & \underline{16.60}\aSd{0.01} & \aBest{17.37}\aSd{0.02} & 2.88 & \underline{2.72} & \aBest{2.18} & 15.90 & \underline{16.81} & \aBest{18.35} \\
\aYes & \aYes & \aNo  & \aNo  & \aYes & 3.83\aSd{0.05} & \underline{2.82}\aSd{0.01} & \aBest{2.43}\aSd{0.02} & 14.03\aSd{0.08} & \underline{16.59}\aSd{0.01} & \aBest{17.53}\aSd{0.04} & 2.92 & \underline{2.73} & \aBest{2.15} & 15.79 & \underline{16.80} & \aBest{18.43} \\
\aYes & \aNo  & \aYes & \aYes & \aNo  & 3.63\aSd{0.03} & \underline{2.82}\aSd{0.00} & \aBest{2.38}\aSd{0.01} & 14.44\aSd{0.06} & \underline{16.59}\aSd{0.01} & \aBest{17.71}\aSd{0.02} & 2.84 & \underline{2.73} & \aBest{2.14} & 16.04 & \underline{16.80} & \aBest{18.48} \\
\aYes & \aNo  & \aYes & \aNo  & \aYes & 3.65\aSd{0.06} & \underline{2.82}\aSd{0.01} & \aBest{2.36}\aSd{0.02} & 14.39\aSd{0.07} & \underline{16.60}\aSd{0.01} & \aBest{17.74}\aSd{0.05} & 2.86 & \underline{2.74} & \aBest{2.12} & 15.94 & \underline{16.79} & \aBest{18.53} \\
\aYes & \aNo  & \aNo  & \aYes & \aYes & 3.83\aSd{0.04} & \underline{2.82}\aSd{0.01} & \aBest{2.42}\aSd{0.02} & 13.95\aSd{0.05} & \underline{16.59}\aSd{0.01} & \aBest{17.57}\aSd{0.05} & 2.97 & \underline{2.73} & \aBest{2.14} & 15.66 & \underline{16.79} & \aBest{18.44} \\
\aNo  & \aYes & \aYes & \aYes & \aNo  & 3.57\aSd{0.06} & \underline{2.82}\aSd{0.01} & \aBest{2.47}\aSd{0.01} & 14.61\aSd{0.08} & \underline{16.58}\aSd{0.01} & \aBest{17.42}\aSd{0.03} & 2.80 & \underline{2.73} & \aBest{2.17} & 16.15 & \underline{16.79} & \aBest{18.37} \\
\aNo  & \aYes & \aYes & \aNo  & \aYes & 3.58\aSd{0.07} & \underline{2.83}\aSd{0.01} & \aBest{2.45}\aSd{0.01} & 14.54\aSd{0.08} & \underline{16.58}\aSd{0.01} & \aBest{17.49}\aSd{0.05} & 2.85 & \underline{2.74} & \aBest{2.14} & 16.00 & \underline{16.79} & \aBest{18.44} \\
\aNo  & \aYes & \aNo  & \aYes & \aYes & 3.79\aSd{0.07} & \underline{2.82}\aSd{0.01} & \aBest{2.49}\aSd{0.02} & 14.08\aSd{0.07} & \underline{16.58}\aSd{0.01} & \aBest{17.33}\aSd{0.05} & 2.94 & \underline{2.73} & \aBest{2.16} & 15.75 & \underline{16.79} & \aBest{18.38} \\
\aNo  & \aNo  & \aYes & \aYes & \aYes & 3.68\aSd{0.07} & \underline{2.82}\aSd{0.00} & \aBest{2.42}\aSd{0.01} & 14.28\aSd{0.10} & \underline{16.59}\aSd{0.00} & \aBest{17.55}\aSd{0.03} & 2.91 & \underline{2.73} & \aBest{2.14} & 15.85 & \underline{16.79} & \aBest{18.46} \\
\addlinespace[1pt]
\aYes & \aYes & \aNo  & \aNo  & \aNo  & \underline{4.99}\aSd{0.08} & \textbf{3.14}\aSd{0.01} & 9.61\aSd{0.03} & \underline{12.00}\aSd{0.06} & \textbf{15.61}\aSd{0.01} & 9.37\aSd{0.03} & \underline{3.62} & \textbf{3.01} & 8.73 & \underline{14.05} & \textbf{15.89} & 10.04 \\
\aYes & \aNo  & \aYes & \aNo  & \aNo  & 5.04\aSd{0.10} & \underline{3.12}\aSd{0.01} & \aBest{2.61}\aSd{0.01} & 11.77\aSd{0.08} & \underline{15.58}\aSd{0.01} & \aBest{16.97}\aSd{0.05} & 3.68 & \underline{3.00} & \aBest{2.28} & 13.84 & \underline{15.87} & \aBest{17.93} \\
\aYes & \aNo  & \aNo  & \aYes & \aNo  & 3.81\aSd{0.07} & \underline{2.81}\aSd{0.01} & \aBest{2.50}\aSd{0.01} & 14.08\aSd{0.05} & \underline{16.59}\aSd{0.01} & \aBest{17.37}\aSd{0.02} & 2.90 & \underline{2.72} & \aBest{2.18} & 15.83 & \underline{16.80} & \aBest{18.35} \\
\aYes & \aNo  & \aNo  & \aNo  & \aYes & 3.87\aSd{0.06} & \underline{2.82}\aSd{0.00} & \aBest{2.43}\aSd{0.02} & 13.95\aSd{0.05} & \underline{16.59}\aSd{0.01} & \aBest{17.53}\aSd{0.04} & 2.93 & \underline{2.73} & \aBest{2.15} & 15.76 & \underline{16.79} & \aBest{18.43} \\
\aNo  & \aYes & \aYes & \aNo  & \aNo  & 4.32\aSd{0.09} & \underline{3.08}\aSd{0.01} & \aBest{2.99}\aSd{0.03} & 12.99\aSd{0.07} & \underline{15.76}\aSd{0.01} & \aBest{16.03}\aSd{0.06} & 3.21 & \underline{2.94} & \aBest{2.41} & 14.97 & \underline{16.04} & \aBest{17.52} \\
\aNo  & \aYes & \aNo  & \aYes & \aNo  & 3.73\aSd{0.06} & \underline{2.82}\aSd{0.00} & \aBest{2.61}\aSd{0.02} & 14.23\aSd{0.09} & \underline{16.58}\aSd{0.01} & \aBest{17.05}\aSd{0.05} & 2.87 & \underline{2.73} & \aBest{2.21} & 15.93 & \underline{16.79} & \aBest{18.23} \\
\aNo  & \aYes & \aNo  & \aNo  & \aYes & 3.83\aSd{0.06} & \underline{2.82}\aSd{0.01} & \aBest{2.52}\aSd{0.01} & 14.08\aSd{0.07} & \underline{16.58}\aSd{0.01} & \aBest{17.26}\aSd{0.04} & 2.88 & \underline{2.74} & \aBest{2.17} & 15.90 & \underline{16.78} & \aBest{18.34} \\
\aNo  & \aNo  & \aYes & \aYes & \aNo  & 3.64\aSd{0.05} & \underline{2.82}\aSd{0.00} & \aBest{2.47}\aSd{0.01} & 14.46\aSd{0.07} & \underline{16.59}\aSd{0.01} & \aBest{17.42}\aSd{0.03} & 2.83 & \underline{2.73} & \aBest{2.17} & 16.02 & \underline{16.80} & \aBest{18.37} \\
\aNo  & \aNo  & \aYes & \aNo  & \aYes & 3.64\aSd{0.06} & \underline{2.82}\aSd{0.00} & \aBest{2.45}\aSd{0.01} & 14.38\aSd{0.05} & \underline{16.58}\aSd{0.01} & \aBest{17.49}\aSd{0.05} & 2.86 & \underline{2.74} & \aBest{2.14} & 15.97 & \underline{16.78} & \aBest{18.44} \\
\aNo  & \aNo  & \aNo  & \aYes & \aYes & 3.82\aSd{0.04} & \underline{2.81}\aSd{0.01} & \aBest{2.49}\aSd{0.02} & 13.98\aSd{0.06} & \underline{16.60}\aSd{0.01} & \aBest{17.33}\aSd{0.05} & 2.92 & \underline{2.73} & \aBest{2.16} & 15.76 & \underline{16.79} & \aBest{18.38} \\
\addlinespace[1pt]
\aYes & \aNo  & \aNo  & \aNo  & \aNo  & \underline{5.75}\aSd{0.08} & \textbf{3.94}\aSd{0.01} & 9.61\aSd{0.03} & \underline{10.96}\aSd{0.05} & \textbf{13.66}\aSd{0.01} & 9.37\aSd{0.03} & \underline{4.03} & \textbf{3.63} & 8.73 & \underline{13.17} & \textbf{14.23} & 10.04 \\
\aNo  & \aNo  & \aYes & \aNo  & \aNo  & 5.13\aSd{0.06} & \underline{3.13}\aSd{0.01} & \aBest{2.99}\aSd{0.03} & 11.64\aSd{0.06} & \underline{15.55}\aSd{0.01} & \aBest{16.03}\aSd{0.06} & 3.72 & \underline{3.00} & \aBest{2.41} & 13.75 & \underline{15.84} & \aBest{17.52} \\
\aNo  & \aNo  & \aNo  & \aYes & \aNo  & 3.82\aSd{0.11} & \underline{2.81}\aSd{0.00} & \aBest{2.61}\aSd{0.02} & 14.07\aSd{0.09} & \underline{16.60}\aSd{0.01} & \aBest{17.05}\aSd{0.05} & 2.88 & \underline{2.72} & \aBest{2.21} & 15.88 & \underline{16.80} & \aBest{18.23} \\
\aNo  & \aNo  & \aNo  & \aNo  & \aYes & 3.90\aSd{0.05} & \underline{2.82}\aSd{0.01} & \aBest{2.52}\aSd{0.01} & 13.97\aSd{0.05} & \underline{16.59}\aSd{0.01} & \aBest{17.26}\aSd{0.04} & 2.92 & \underline{2.73} & \aBest{2.17} & 15.78 & \underline{16.79} & \aBest{18.34} \\
\bottomrule
\end{tabular}}
\aSweepNote{\aSweepLULC}
\endgroup
\end{table*}

% ============================================================
% Sentinel-2 L1C generation
% ============================================================
\begin{table*}[t]
\centering
\caption{\textbf{EAGLE has the lowest mean-of-8 S2 L1C MAE in all 30 configurations and the highest PSNR in 29.}
MAE $\downarrow$ and PSNR $\uparrow$. \aSweepEval}
\label{tab:full-s2l1c}
\begingroup
\aSweepSetup
\resizebox{\textwidth}{!}{%
\begin{tabular}{@{}ccccc rr>{\columncolor{A2Eshade}}r rr>{\columncolor{A2Eshade}}r rr>{\columncolor{A2Eshade}}r rr>{\columncolor{A2Eshade}}r@{}}
\toprule
\multicolumn{5}{c}{\textbf{Available inputs}} & \multicolumn{6}{c}{\textbf{Mean of 8 draws}} & \multicolumn{6}{c}{\textbf{Best of 8 draws (oracle)}} \\
\cmidrule(lr){1-5}\cmidrule(lr){6-11}\cmidrule(l){12-17}
& & & & & \multicolumn{3}{c}{MAE $\downarrow$} & \multicolumn{3}{c}{PSNR $\uparrow$} & \multicolumn{3}{c}{MAE $\downarrow$} & \multicolumn{3}{c}{PSNR $\uparrow$} \\
\cmidrule(l){6-8}\cmidrule(l){9-11}\cmidrule(l){12-14}\cmidrule(l){15-17}
Coord. & LULC$^{\ddagger}$ & DEM & S1 & L2A & COP-GEN & TerraMind & \aOurs & COP-GEN & TerraMind & \aOurs & COP-GEN & TerraMind & \aOurs & COP-GEN & TerraMind & \aOurs \\
\midrule
\aYes & \aYes & \aYes & \aYes & \aYes & \underline{0.0626}\aSd{0.0010} & 0.1108\aSd{0.0001} & \aBest{0.0140}\aSd{0.0001} & \underline{11.98}\aSd{0.19} & 9.91\aSd{0.01} & \aBest{20.22}\aSd{0.06} & \underline{0.0217} & 0.1076 & \aBest{0.0120} & \underline{18.42} & 10.16 & \aBest{22.09} \\
\addlinespace[1pt]
\aYes & \aYes & \aYes & \aYes & \aNo  & 0.1311\aSd{0.0023} & \underline{0.1237}\aSd{0.0002} & \aBest{0.0418}\aSd{0.0003} & 7.31\aSd{0.08} & \underline{8.68}\aSd{0.01} & \aBest{13.68}\aSd{0.03} & \underline{0.0680} & 0.1169 & \aBest{0.0345} & \underline{10.62} & 9.08 & \aBest{15.25} \\
\aYes & \aYes & \aYes & \aNo  & \aYes & \underline{0.0631}\aSd{0.0019} & 0.1107\aSd{0.0001} & \aBest{0.0145}\aSd{0.0001} & \underline{11.89}\aSd{0.26} & 9.92\aSd{0.01} & \aBest{20.03}\aSd{0.05} & \underline{0.0209} & 0.1074 & \aBest{0.0121} & \underline{18.35} & 10.17 & \aBest{22.02} \\
\aYes & \aYes & \aNo  & \aYes & \aYes & \underline{0.0641}\aSd{0.0009} & 0.1106\aSd{0.0001} & \aBest{0.0148}\aSd{0.0001} & \underline{11.64}\aSd{0.14} & 9.93\aSd{0.01} & \aBest{19.86}\aSd{0.07} & \underline{0.0226} & 0.1073 & \aBest{0.0124} & \underline{17.89} & 10.18 & \aBest{21.89} \\
\aYes & \aNo  & \aYes & \aYes & \aYes & \underline{0.0621}\aSd{0.0017} & 0.1106\aSd{0.0001} & \aBest{0.0140}\aSd{0.0001} & \underline{12.03}\aSd{0.22} & 9.93\aSd{0.01} & \aBest{20.22}\aSd{0.06} & \underline{0.0213} & 0.1074 & \aBest{0.0120} & \underline{18.39} & 10.18 & \aBest{22.09} \\
\aNo  & \aYes & \aYes & \aYes & \aYes & \underline{0.0636}\aSd{0.0015} & 0.1111\aSd{0.0001} & \aBest{0.0151}\aSd{0.0001} & \underline{11.84}\aSd{0.22} & 9.90\aSd{0.01} & \aBest{19.67}\aSd{0.08} & \underline{0.0227} & 0.1079 & \aBest{0.0124} & \underline{18.16} & 10.15 & \aBest{21.83} \\
\addlinespace[1pt]
\aYes & \aYes & \aYes & \aNo  & \aNo  & 0.1422\aSd{0.0022} & \underline{0.1234}\aSd{0.0003} & \aBest{0.0495}\aSd{0.0002} & 6.95\aSd{0.06} & \underline{8.69}\aSd{0.02} & \aBest{12.52}\aSd{0.04} & \underline{0.0775} & 0.1165 & \aBest{0.0396} & \underline{9.75} & 9.07 & \aBest{14.21} \\
\aYes & \aYes & \aNo  & \aYes & \aNo  & 0.1352\aSd{0.0027} & \underline{0.1259}\aSd{0.0003} & \aBest{0.0451}\aSd{0.0004} & 7.26\aSd{0.10} & \underline{8.55}\aSd{0.01} & \aBest{13.33}\aSd{0.03} & \underline{0.0698} & 0.1188 & \aBest{0.0368} & \underline{10.46} & 8.95 & \aBest{14.92} \\
\aYes & \aYes & \aNo  & \aNo  & \aYes & \underline{0.0666}\aSd{0.0018} & 0.1106\aSd{0.0001} & \aBest{0.0161}\aSd{0.0001} & \underline{11.29}\aSd{0.21} & 9.93\aSd{0.01} & \aBest{19.42}\aSd{0.06} & \underline{0.0227} & 0.1074 & \aBest{0.0132} & \underline{17.67} & 10.18 & \aBest{21.56} \\
\aYes & \aNo  & \aYes & \aYes & \aNo  & 0.1387\aSd{0.0023} & \underline{0.1242}\aSd{0.0002} & \aBest{0.0418}\aSd{0.0003} & 7.11\aSd{0.07} & \underline{8.54}\aSd{0.01} & \aBest{13.68}\aSd{0.03} & \underline{0.0753} & 0.1168 & \aBest{0.0345} & \underline{10.08} & 8.95 & \aBest{15.25} \\
\aYes & \aNo  & \aYes & \aNo  & \aYes & \underline{0.0621}\aSd{0.0017} & 0.1106\aSd{0.0001} & \aBest{0.0145}\aSd{0.0001} & \underline{11.97}\aSd{0.23} & 9.93\aSd{0.01} & \aBest{20.03}\aSd{0.05} & \underline{0.0212} & 0.1072 & \aBest{0.0121} & \underline{18.39} & 10.18 & \aBest{22.02} \\
\aYes & \aNo  & \aNo  & \aYes & \aYes & \underline{0.0637}\aSd{0.0011} & 0.1106\aSd{0.0002} & \aBest{0.0148}\aSd{0.0001} & \underline{11.66}\aSd{0.10} & 9.93\aSd{0.01} & \aBest{19.86}\aSd{0.07} & \underline{0.0222} & 0.1073 & \aBest{0.0124} & \underline{17.93} & 10.18 & \aBest{21.89} \\
\aNo  & \aYes & \aYes & \aYes & \aNo  & 0.1343\aSd{0.0023} & \underline{0.1247}\aSd{0.0003} & \aBest{0.0513}\aSd{0.0006} & 7.26\aSd{0.06} & \underline{8.65}\aSd{0.01} & \aBest{12.59}\aSd{0.09} & \underline{0.0687} & 0.1179 & \aBest{0.0393} & \underline{10.54} & 9.05 & \aBest{14.44} \\
\aNo  & \aYes & \aYes & \aNo  & \aYes & \underline{0.0638}\aSd{0.0019} & 0.1110\aSd{0.0001} & \aBest{0.0157}\aSd{0.0001} & \underline{11.80}\aSd{0.28} & 9.90\aSd{0.00} & \aBest{19.37}\aSd{0.04} & \underline{0.0227} & 0.1077 & \aBest{0.0124} & \underline{18.16} & 10.16 & \aBest{21.64} \\
\aNo  & \aYes & \aNo  & \aYes & \aYes & \underline{0.0645}\aSd{0.0016} & 0.1109\aSd{0.0001} & \aBest{0.0160}\aSd{0.0001} & \underline{11.56}\aSd{0.24} & 9.92\aSd{0.01} & \aBest{19.13}\aSd{0.06} & \underline{0.0228} & 0.1076 & \aBest{0.0129} & \underline{17.88} & 10.17 & \aBest{21.38} \\
\aNo  & \aNo  & \aYes & \aYes & \aYes & \underline{0.0618}\aSd{0.0016} & 0.1109\aSd{0.0001} & \aBest{0.0151}\aSd{0.0001} & \underline{12.01}\aSd{0.21} & 9.92\aSd{0.01} & \aBest{19.67}\aSd{0.08} & \underline{0.0217} & 0.1076 & \aBest{0.0124} & \underline{18.29} & 10.16 & \aBest{21.83} \\
\addlinespace[1pt]
\aYes & \aYes & \aNo  & \aNo  & \aNo  & 0.1540\aSd{0.0025} & \underline{0.1248}\aSd{0.0002} & \aBest{0.1247}\aSd{0.0004} & 6.77\aSd{0.05} & \textbf{8.74}\aSd{0.01} & \underline{8.56}\aSd{0.02} & \textbf{0.0879} & 0.1188 & \underline{0.1179} & \textbf{9.13} & \underline{9.09} & 8.88 \\
\aYes & \aNo  & \aYes & \aNo  & \aNo  & 0.1547\aSd{0.0037} & \underline{0.1261}\aSd{0.0003} & \aBest{0.0495}\aSd{0.0002} & 6.61\aSd{0.08} & \underline{8.30}\aSd{0.02} & \aBest{12.52}\aSd{0.04} & \underline{0.0895} & 0.1180 & \aBest{0.0396} & \underline{8.93} & 8.71 & \aBest{14.21} \\
\aYes & \aNo  & \aNo  & \aYes & \aNo  & 0.1430\aSd{0.0025} & \underline{0.1256}\aSd{0.0002} & \aBest{0.0451}\aSd{0.0004} & 7.00\aSd{0.09} & \underline{8.50}\aSd{0.02} & \aBest{13.33}\aSd{0.03} & \underline{0.0789} & 0.1185 & \aBest{0.0368} & \underline{9.75} & 8.91 & \aBest{14.92} \\
\aYes & \aNo  & \aNo  & \aNo  & \aYes & \underline{0.0651}\aSd{0.0017} & 0.1105\aSd{0.0001} & \aBest{0.0161}\aSd{0.0001} & \underline{11.46}\aSd{0.19} & 9.93\aSd{0.01} & \aBest{19.42}\aSd{0.06} & \underline{0.0223} & 0.1072 & \aBest{0.0132} & \underline{17.76} & 10.17 & \aBest{21.56} \\
\aNo  & \aYes & \aYes & \aNo  & \aNo  & 0.1477\aSd{0.0025} & \underline{0.1253}\aSd{0.0002} & \aBest{0.0787}\aSd{0.0011} & 6.88\aSd{0.15} & \underline{8.60}\aSd{0.01} & \aBest{10.39}\aSd{0.06} & \underline{0.0814} & 0.1181 & \aBest{0.0545} & \underline{9.59} & 8.99 & \aBest{12.48} \\
\aNo  & \aYes & \aNo  & \aYes & \aNo  & 0.1407\aSd{0.0021} & \underline{0.1269}\aSd{0.0001} & \aBest{0.0596}\aSd{0.0012} & 7.11\aSd{0.09} & \underline{8.52}\aSd{0.01} & \aBest{12.08}\aSd{0.10} & \underline{0.0734} & 0.1202 & \aBest{0.0449} & \underline{10.16} & 8.90 & \aBest{13.93} \\
\aNo  & \aYes & \aNo  & \aNo  & \aYes & \underline{0.0670}\aSd{0.0026} & 0.1108\aSd{0.0001} & \aBest{0.0178}\aSd{0.0002} & \underline{11.28}\aSd{0.32} & 9.92\aSd{0.01} & \aBest{18.54}\aSd{0.06} & \underline{0.0233} & 0.1075 & \aBest{0.0139} & \underline{17.60} & 10.16 & \aBest{20.97} \\
\aNo  & \aNo  & \aYes & \aYes & \aNo  & 0.1389\aSd{0.0027} & \underline{0.1260}\aSd{0.0003} & \aBest{0.0513}\aSd{0.0006} & 7.11\aSd{0.11} & \underline{8.40}\aSd{0.02} & \aBest{12.59}\aSd{0.09} & \underline{0.0736} & 0.1186 & \aBest{0.0393} & \underline{10.00} & 8.81 & \aBest{14.44} \\
\aNo  & \aNo  & \aYes & \aNo  & \aYes & \underline{0.0620}\aSd{0.0011} & 0.1108\aSd{0.0001} & \aBest{0.0157}\aSd{0.0001} & \underline{11.99}\aSd{0.15} & 9.91\aSd{0.01} & \aBest{19.37}\aSd{0.04} & \underline{0.0221} & 0.1075 & \aBest{0.0124} & \underline{18.28} & 10.17 & \aBest{21.64} \\
\aNo  & \aNo  & \aNo  & \aYes & \aYes & \underline{0.0638}\aSd{0.0012} & 0.1107\aSd{0.0001} & \aBest{0.0160}\aSd{0.0001} & \underline{11.65}\aSd{0.19} & 9.94\aSd{0.01} & \aBest{19.13}\aSd{0.06} & \underline{0.0234} & 0.1074 & \aBest{0.0129} & \underline{17.87} & 10.19 & \aBest{21.38} \\
\addlinespace[1pt]
\aYes & \aNo  & \aNo  & \aNo  & \aNo  & 0.1720\aSd{0.0036} & \underline{0.1340}\aSd{0.0003} & \aBest{0.1247}\aSd{0.0004} & 6.34\aSd{0.07} & \underline{7.20}\aSd{0.01} & \aBest{8.56}\aSd{0.02} & \textbf{0.0988} & 0.1227 & \underline{0.1179} & \underline{8.38} & 7.71 & \aBest{8.88} \\
\aNo  & \aNo  & \aYes & \aNo  & \aNo  & 0.1639\aSd{0.0034} & \underline{0.1293}\aSd{0.0003} & \aBest{0.0787}\aSd{0.0011} & 6.45\aSd{0.06} & \underline{8.08}\aSd{0.02} & \aBest{10.39}\aSd{0.06} & \underline{0.0928} & 0.1211 & \aBest{0.0545} & \underline{8.64} & 8.50 & \aBest{12.48} \\
\aNo  & \aNo  & \aNo  & \aYes & \aNo  & 0.1532\aSd{0.0015} & \underline{0.1276}\aSd{0.0003} & \aBest{0.0596}\aSd{0.0012} & 6.83\aSd{0.08} & \underline{8.39}\aSd{0.01} & \aBest{12.08}\aSd{0.10} & \underline{0.0824} & 0.1203 & \aBest{0.0449} & \underline{9.50} & 8.80 & \aBest{13.93} \\
\aNo  & \aNo  & \aNo  & \aNo  & \aYes & \underline{0.0653}\aSd{0.0020} & 0.1105\aSd{0.0001} & \aBest{0.0178}\aSd{0.0002} & \underline{11.44}\aSd{0.23} & 9.93\aSd{0.01} & \aBest{18.54}\aSd{0.06} & \underline{0.0227} & 0.1072 & \aBest{0.0139} & \underline{17.62} & 10.18 & \aBest{20.97} \\
\bottomrule
\end{tabular}}
\aSweepNote{\aSweepLULC}
\endgroup
\end{table*}

% ============================================================
% Sentinel-2 L2A generation
% ============================================================
\begin{table*}[t]
\centering
\caption{\textbf{EAGLE has the lowest mean-of-8 S2 L2A MAE in 28 of 30 configurations and the highest PSNR in 29.}
MAE $\downarrow$ and PSNR $\uparrow$. \aSweepEval}
\label{tab:full-s2l2a}
\begingroup
\aSweepSetup
\resizebox{\textwidth}{!}{%
\begin{tabular}{@{}ccccc rr>{\columncolor{A2Eshade}}r rr>{\columncolor{A2Eshade}}r rr>{\columncolor{A2Eshade}}r rr>{\columncolor{A2Eshade}}r@{}}
\toprule
\multicolumn{5}{c}{\textbf{Available inputs}} & \multicolumn{6}{c}{\textbf{Mean of 8 draws}} & \multicolumn{6}{c}{\textbf{Best of 8 draws (oracle)}} \\
\cmidrule(lr){1-5}\cmidrule(lr){6-11}\cmidrule(l){12-17}
& & & & & \multicolumn{3}{c}{MAE $\downarrow$} & \multicolumn{3}{c}{PSNR $\uparrow$} & \multicolumn{3}{c}{MAE $\downarrow$} & \multicolumn{3}{c}{PSNR $\uparrow$} \\
\cmidrule(l){6-8}\cmidrule(l){9-11}\cmidrule(l){12-14}\cmidrule(l){15-17}
Coord. & LULC$^{\ddagger}$ & DEM & S1 & L1C & COP-GEN & TerraMind & \aOurs & COP-GEN & TerraMind & \aOurs & COP-GEN & TerraMind & \aOurs & COP-GEN & TerraMind & \aOurs \\
\midrule
\aYes & \aYes & \aYes & \aYes & \aYes & \underline{0.0363}\aSd{0.0007} & 0.0985\aSd{0.0002} & \aBest{0.0250}\aSd{0.0003} & \underline{17.53}\aSd{0.07} & 12.55\aSd{0.03} & \aBest{20.06}\aSd{0.07} & \underline{0.0229} & 0.0948 & \aBest{0.0215} & \underline{20.16} & 13.15 & \aBest{21.33} \\
\addlinespace[1pt]
\aYes & \aYes & \aYes & \aYes & \aNo  & 0.1490\aSd{0.0039} & \underline{0.1191}\aSd{0.0003} & \aBest{0.0548}\aSd{0.0003} & 8.23\aSd{0.10} & \underline{9.94}\aSd{0.02} & \aBest{14.16}\aSd{0.04} & \underline{0.0790} & 0.1117 & \aBest{0.0453} & \underline{11.54} & 10.63 & \aBest{15.64} \\
\aYes & \aYes & \aYes & \aNo  & \aYes & \underline{0.0354}\aSd{0.0009} & 0.0988\aSd{0.0001} & \aBest{0.0246}\aSd{0.0001} & \underline{17.66}\aSd{0.12} & 12.54\aSd{0.03} & \aBest{20.15}\aSd{0.04} & \underline{0.0230} & 0.0950 & \aBest{0.0210} & \underline{20.12} & 13.16 & \aBest{21.53} \\
\aYes & \aYes & \aNo  & \aYes & \aYes & \underline{0.0368}\aSd{0.0009} & 0.0987\aSd{0.0001} & \aBest{0.0257}\aSd{0.0002} & \underline{17.46}\aSd{0.13} & 12.55\aSd{0.02} & \aBest{19.94}\aSd{0.05} & \underline{0.0230} & 0.0951 & \aBest{0.0223} & \underline{20.14} & 13.14 & \aBest{21.28} \\
\aYes & \aNo  & \aYes & \aYes & \aYes & \underline{0.0340}\aSd{0.0007} & 0.0984\aSd{0.0001} & \aBest{0.0250}\aSd{0.0003} & \underline{17.86}\aSd{0.09} & 12.58\aSd{0.03} & \aBest{20.06}\aSd{0.07} & \underline{0.0226} & 0.0947 & \aBest{0.0215} & \underline{20.20} & 13.19 & \aBest{21.33} \\
\aNo  & \aYes & \aYes & \aYes & \aYes & \underline{0.0358}\aSd{0.0017} & 0.0988\aSd{0.0002} & \aBest{0.0256}\aSd{0.0001} & \underline{17.59}\aSd{0.25} & 12.52\aSd{0.03} & \aBest{19.86}\aSd{0.03} & \underline{0.0227} & 0.0949 & \aBest{0.0217} & \underline{20.18} & 13.14 & \aBest{21.33} \\
\addlinespace[1pt]
\aYes & \aYes & \aYes & \aNo  & \aNo  & 0.1622\aSd{0.0028} & \underline{0.1193}\aSd{0.0001} & \aBest{0.0648}\aSd{0.0009} & 7.73\aSd{0.07} & \underline{9.97}\aSd{0.02} & \aBest{12.94}\aSd{0.07} & \underline{0.0890} & 0.1121 & \aBest{0.0522} & \underline{10.63} & 10.59 & \aBest{14.52} \\
\aYes & \aYes & \aNo  & \aYes & \aNo  & 0.1549\aSd{0.0031} & \underline{0.1214}\aSd{0.0004} & \aBest{0.0574}\aSd{0.0002} & 8.14\aSd{0.08} & \underline{9.75}\aSd{0.03} & \aBest{13.91}\aSd{0.05} & \underline{0.0804} & 0.1136 & \aBest{0.0472} & \underline{11.43} & 10.44 & \aBest{15.39} \\
\aYes & \aYes & \aNo  & \aNo  & \aYes & \underline{0.0371}\aSd{0.0010} & 0.0988\aSd{0.0001} & \aBest{0.0265}\aSd{0.0002} & \underline{17.31}\aSd{0.14} & 12.53\aSd{0.01} & \aBest{19.75}\aSd{0.06} & \underline{0.0236} & 0.0951 & \aBest{0.0224} & \underline{19.96} & 13.15 & \aBest{21.25} \\
\aYes & \aNo  & \aYes & \aYes & \aNo  & 0.1604\aSd{0.0031} & \underline{0.1207}\aSd{0.0003} & \aBest{0.0548}\aSd{0.0003} & 7.87\aSd{0.05} & \underline{9.64}\aSd{0.03} & \aBest{14.16}\aSd{0.04} & \underline{0.0882} & 0.1127 & \aBest{0.0453} & \underline{10.87} & 10.30 & \aBest{15.64} \\
\aYes & \aNo  & \aYes & \aNo  & \aYes & \underline{0.0338}\aSd{0.0007} & 0.0987\aSd{0.0001} & \aBest{0.0246}\aSd{0.0001} & \underline{17.83}\aSd{0.11} & 12.56\aSd{0.02} & \aBest{20.15}\aSd{0.04} & \underline{0.0228} & 0.0949 & \aBest{0.0210} & \underline{20.12} & 13.17 & \aBest{21.53} \\
\aYes & \aNo  & \aNo  & \aYes & \aYes & \underline{0.0345}\aSd{0.0009} & 0.0986\aSd{0.0001} & \aBest{0.0257}\aSd{0.0002} & \underline{17.74}\aSd{0.16} & 12.57\aSd{0.01} & \aBest{19.94}\aSd{0.05} & \underline{0.0228} & 0.0948 & \aBest{0.0223} & \underline{20.13} & 13.19 & \aBest{21.28} \\
\aNo  & \aYes & \aYes & \aYes & \aNo  & 0.1533\aSd{0.0032} & \underline{0.1204}\aSd{0.0001} & \aBest{0.0656}\aSd{0.0009} & 8.13\aSd{0.04} & \underline{9.91}\aSd{0.02} & \aBest{13.13}\aSd{0.04} & \underline{0.0802} & 0.1129 & \aBest{0.0500} & \underline{11.39} & 10.60 & \aBest{14.97} \\
\aNo  & \aYes & \aYes & \aNo  & \aYes & \underline{0.0358}\aSd{0.0012} & 0.0990\aSd{0.0001} & \aBest{0.0255}\aSd{0.0002} & \underline{17.59}\aSd{0.16} & 12.52\aSd{0.03} & \aBest{19.87}\aSd{0.08} & \underline{0.0229} & 0.0953 & \aBest{0.0212} & \underline{20.15} & 13.13 & \aBest{21.45} \\
\aNo  & \aYes & \aNo  & \aYes & \aYes & \underline{0.0353}\aSd{0.0011} & 0.0989\aSd{0.0002} & \aBest{0.0266}\aSd{0.0001} & \underline{17.65}\aSd{0.17} & 12.52\aSd{0.02} & \aBest{19.69}\aSd{0.05} & \underline{0.0229} & 0.0951 & \aBest{0.0223} & \underline{20.13} & 13.12 & \aBest{21.24} \\
\aNo  & \aNo  & \aYes & \aYes & \aYes & \underline{0.0340}\aSd{0.0015} & 0.0984\aSd{0.0001} & \aBest{0.0256}\aSd{0.0001} & \underline{17.87}\aSd{0.25} & 12.59\aSd{0.03} & \aBest{19.86}\aSd{0.03} & \underline{0.0226} & 0.0947 & \aBest{0.0217} & \underline{20.20} & 13.20 & \aBest{21.33} \\
\addlinespace[1pt]
\aYes & \aYes & \aNo  & \aNo  & \aNo  & 0.1761\aSd{0.0035} & \textbf{0.1200}\aSd{0.0003} & \underline{0.1576}\aSd{0.0004} & 7.57\aSd{0.10} & \textbf{10.08}\aSd{0.03} & \underline{8.81}\aSd{0.02} & \textbf{0.0943} & \underline{0.1134} & 0.1484 & \underline{10.40} & \textbf{10.69} & 9.17 \\
\aYes & \aNo  & \aYes & \aNo  & \aNo  & 0.1803\aSd{0.0042} & \underline{0.1238}\aSd{0.0003} & \aBest{0.0648}\aSd{0.0009} & 7.20\aSd{0.08} & \underline{9.31}\aSd{0.02} & \aBest{12.94}\aSd{0.07} & \underline{0.1042} & 0.1154 & \aBest{0.0522} & 9.65 & \underline{9.92} & \aBest{14.52} \\
\aYes & \aNo  & \aNo  & \aYes & \aNo  & 0.1659\aSd{0.0032} & \underline{0.1217}\aSd{0.0002} & \aBest{0.0574}\aSd{0.0002} & 7.71\aSd{0.08} & \underline{9.61}\aSd{0.02} & \aBest{13.91}\aSd{0.05} & \underline{0.0921} & 0.1138 & \aBest{0.0472} & \underline{10.52} & 10.29 & \aBest{15.39} \\
\aYes & \aNo  & \aNo  & \aNo  & \aYes & \underline{0.0345}\aSd{0.0011} & 0.0988\aSd{0.0001} & \aBest{0.0265}\aSd{0.0002} & \underline{17.64}\aSd{0.16} & 12.55\aSd{0.02} & \aBest{19.75}\aSd{0.06} & \underline{0.0235} & 0.0951 & \aBest{0.0224} & \underline{20.01} & 13.15 & \aBest{21.25} \\
\aNo  & \aYes & \aYes & \aNo  & \aNo  & 0.1693\aSd{0.0033} & \underline{0.1217}\aSd{0.0002} & \aBest{0.0984}\aSd{0.0021} & 7.67\aSd{0.13} & \underline{9.85}\aSd{0.01} & \aBest{10.84}\aSd{0.13} & \underline{0.0903} & 0.1144 & \aBest{0.0697} & \underline{10.67} & 10.45 & \aBest{12.96} \\
\aNo  & \aYes & \aNo  & \aYes & \aNo  & 0.1614\aSd{0.0031} & \underline{0.1229}\aSd{0.0003} & \aBest{0.0751}\aSd{0.0009} & 7.95\aSd{0.09} & \underline{9.67}\aSd{0.02} & \aBest{12.58}\aSd{0.06} & \underline{0.0835} & 0.1150 & \aBest{0.0569} & \underline{11.19} & 10.35 & \aBest{14.37} \\
\aNo  & \aYes & \aNo  & \aNo  & \aYes & \underline{0.0352}\aSd{0.0007} & 0.0992\aSd{0.0001} & \aBest{0.0276}\aSd{0.0002} & \underline{17.57}\aSd{0.12} & 12.49\aSd{0.01} & \aBest{19.40}\aSd{0.05} & \underline{0.0234} & 0.0954 & \aBest{0.0224} & \underline{20.00} & 13.10 & \aBest{21.16} \\
\aNo  & \aNo  & \aYes & \aYes & \aNo  & 0.1605\aSd{0.0039} & \underline{0.1229}\aSd{0.0004} & \aBest{0.0656}\aSd{0.0009} & 7.84\aSd{0.14} & \underline{9.45}\aSd{0.02} & \aBest{13.13}\aSd{0.04} & \underline{0.0873} & 0.1143 & \aBest{0.0500} & \underline{10.77} & 10.14 & \aBest{14.97} \\
\aNo  & \aNo  & \aYes & \aNo  & \aYes & \underline{0.0336}\aSd{0.0014} & 0.0988\aSd{0.0001} & \aBest{0.0255}\aSd{0.0002} & \underline{17.89}\aSd{0.19} & 12.54\aSd{0.03} & \aBest{19.87}\aSd{0.08} & \underline{0.0229} & 0.0951 & \aBest{0.0212} & \underline{20.14} & 13.13 & \aBest{21.45} \\
\aNo  & \aNo  & \aNo  & \aYes & \aYes & \underline{0.0342}\aSd{0.0014} & 0.0985\aSd{0.0001} & \aBest{0.0266}\aSd{0.0001} & \underline{17.83}\aSd{0.16} & 12.61\aSd{0.02} & \aBest{19.69}\aSd{0.05} & \underline{0.0228} & 0.0947 & \aBest{0.0223} & \underline{20.15} & 13.22 & \aBest{21.24} \\
\addlinespace[1pt]
\aYes & \aNo  & \aNo  & \aNo  & \aNo  & 0.1999\aSd{0.0038} & \textbf{0.1422}\aSd{0.0004} & \underline{0.1576}\aSd{0.0004} & 6.90\aSd{0.10} & \underline{7.78}\aSd{0.02} & \aBest{8.81}\aSd{0.02} & \textbf{0.1108} & \underline{0.1297} & 0.1484 & \textbf{9.23} & 8.34 & \underline{9.17} \\
\aNo  & \aNo  & \aYes & \aNo  & \aNo  & 0.1902\aSd{0.0020} & \underline{0.1278}\aSd{0.0002} & \aBest{0.0984}\aSd{0.0021} & 7.10\aSd{0.03} & \underline{8.98}\aSd{0.02} & \aBest{10.84}\aSd{0.13} & \underline{0.1066} & 0.1186 & \aBest{0.0697} & 9.52 & \underline{9.61} & \aBest{12.96} \\
\aNo  & \aNo  & \aNo  & \aYes & \aNo  & 0.1751\aSd{0.0024} & \underline{0.1244}\aSd{0.0002} & \aBest{0.0751}\aSd{0.0009} & 7.59\aSd{0.07} & \underline{9.44}\aSd{0.02} & \aBest{12.58}\aSd{0.06} & \underline{0.0930} & 0.1160 & \aBest{0.0569} & \underline{10.42} & 10.12 & \aBest{14.37} \\
\aNo  & \aNo  & \aNo  & \aNo  & \aYes & \underline{0.0344}\aSd{0.0015} & 0.0990\aSd{0.0001} & \aBest{0.0276}\aSd{0.0002} & \underline{17.66}\aSd{0.18} & 12.53\aSd{0.03} & \aBest{19.40}\aSd{0.05} & \underline{0.0234} & 0.0952 & \aBest{0.0224} & \underline{20.02} & 13.14 & \aBest{21.16} \\
\bottomrule
\end{tabular}}
\aSweepNote{\aSweepLULC}
\endgroup
\end{table*}

% ============================================================
% NDVI generation
% ============================================================
\begin{table*}[t]
\centering
\caption{\textbf{EAGLE has the lowest mean-of-8 NDVI MAE in 28 of 30 configurations.}
MAE $\downarrow$ and PSNR $\uparrow$. \aSweepEval}
\label{tab:full-ndvi}
\begingroup
\aSweepSetup
\resizebox{\textwidth}{!}{%
\begin{tabular}{@{}ccccc rr>{\columncolor{A2Eshade}}r rr>{\columncolor{A2Eshade}}r rr>{\columncolor{A2Eshade}}r rr>{\columncolor{A2Eshade}}r@{}}
\toprule
\multicolumn{5}{c}{\textbf{Available inputs}} & \multicolumn{6}{c}{\textbf{Mean of 8 draws}} & \multicolumn{6}{c}{\textbf{Best of 8 draws (oracle)}} \\
\cmidrule(lr){1-5}\cmidrule(lr){6-11}\cmidrule(l){12-17}
& & & & & \multicolumn{3}{c}{MAE $\downarrow$} & \multicolumn{3}{c}{PSNR $\uparrow$} & \multicolumn{3}{c}{MAE $\downarrow$} & \multicolumn{3}{c}{PSNR $\uparrow$} \\
\cmidrule(l){6-8}\cmidrule(l){9-11}\cmidrule(l){12-14}\cmidrule(l){15-17}
Coord. & LULC$^{\ddagger}$ & DEM & S1 & L1C & COP-GEN & TerraMind & \aOurs & COP-GEN & TerraMind & \aOurs & COP-GEN & TerraMind & \aOurs & COP-GEN & TerraMind & \aOurs \\
\midrule
\aYes & \aYes & \aYes & \aYes & \aYes & \underline{0.086}\aSd{0.002} & 0.192\aSd{0.001} & \aBest{0.051}\aSd{0.001} & \underline{18.87}\aSd{0.08} & 14.50\aSd{0.03} & \aBest{22.56}\aSd{0.06} & \underline{0.053} & 0.176 & \aBest{0.042} & \underline{21.17} & 15.32 & \aBest{24.05} \\
\addlinespace[1pt]
\aYes & \aYes & \aYes & \aYes & \aNo  & 0.258\aSd{0.006} & \underline{0.219}\aSd{0.001} & \aBest{0.110}\aSd{0.002} & 10.10\aSd{0.10} & \underline{12.27}\aSd{0.03} & \aBest{15.55}\aSd{0.09} & \underline{0.146} & 0.190 & \aBest{0.088} & \underline{13.65} & 13.36 & \aBest{17.28} \\
\aYes & \aYes & \aYes & \aNo  & \aYes & \underline{0.083}\aSd{0.002} & 0.192\aSd{0.001} & \aBest{0.050}\aSd{0.001} & \underline{19.01}\aSd{0.10} & 14.46\aSd{0.03} & \aBest{22.70}\aSd{0.06} & \underline{0.053} & 0.176 & \aBest{0.040} & \underline{21.19} & 15.28 & \aBest{24.33} \\
\aYes & \aYes & \aNo  & \aYes & \aYes & \underline{0.083}\aSd{0.003} & 0.193\aSd{0.001} & \aBest{0.050}\aSd{0.001} & \underline{19.00}\aSd{0.13} & 14.46\aSd{0.02} & \aBest{22.68}\aSd{0.07} & \underline{0.053} & 0.175 & \aBest{0.040} & \underline{21.32} & 15.29 & \aBest{24.29} \\
\aYes & \aNo  & \aYes & \aYes & \aYes & \underline{0.081}\aSd{0.004} & 0.192\aSd{0.001} & \aBest{0.051}\aSd{0.001} & \underline{19.17}\aSd{0.16} & 14.49\aSd{0.02} & \aBest{22.56}\aSd{0.06} & \underline{0.052} & 0.175 & \aBest{0.042} & \underline{21.32} & 15.33 & \aBest{24.05} \\
\aNo  & \aYes & \aYes & \aYes & \aYes & \underline{0.084}\aSd{0.003} & 0.193\aSd{0.001} & \aBest{0.052}\aSd{0.001} & \underline{18.99}\aSd{0.15} & 14.49\aSd{0.03} & \aBest{22.29}\aSd{0.08} & \underline{0.053} & 0.176 & \aBest{0.040} & \underline{21.26} & 15.33 & \aBest{24.00} \\
\addlinespace[1pt]
\aYes & \aYes & \aYes & \aNo  & \aNo  & 0.288\aSd{0.004} & \underline{0.249}\aSd{0.001} & \aBest{0.136}\aSd{0.002} & 9.44\aSd{0.08} & \underline{11.21}\aSd{0.03} & \aBest{13.94}\aSd{0.08} & \underline{0.171} & 0.217 & \aBest{0.105} & \underline{12.63} & 12.27 & \aBest{15.96} \\
\aYes & \aYes & \aNo  & \aYes & \aNo  & 0.266\aSd{0.004} & \underline{0.220}\aSd{0.001} & \aBest{0.113}\aSd{0.001} & 9.98\aSd{0.13} & \underline{12.31}\aSd{0.03} & \aBest{15.35}\aSd{0.05} & \underline{0.147} & 0.191 & \aBest{0.089} & \underline{13.61} & 13.41 & \aBest{17.17} \\
\aYes & \aYes & \aNo  & \aNo  & \aYes & \underline{0.081}\aSd{0.002} & 0.193\aSd{0.001} & \aBest{0.052}\aSd{0.001} & \underline{19.04}\aSd{0.09} & 14.43\aSd{0.02} & \aBest{22.60}\aSd{0.07} & \underline{0.053} & 0.176 & \aBest{0.040} & \underline{21.26} & 15.26 & \aBest{24.29} \\
\aYes & \aNo  & \aYes & \aYes & \aNo  & 0.288\aSd{0.002} & \underline{0.217}\aSd{0.001} & \aBest{0.110}\aSd{0.002} & 9.25\aSd{0.11} & \underline{12.07}\aSd{0.04} & \aBest{15.55}\aSd{0.09} & \underline{0.167} & 0.188 & \aBest{0.088} & 12.60 & \underline{13.14} & \aBest{17.28} \\
\aYes & \aNo  & \aYes & \aNo  & \aYes & \underline{0.080}\aSd{0.002} & 0.192\aSd{0.001} & \aBest{0.050}\aSd{0.001} & \underline{19.16}\aSd{0.10} & 14.47\aSd{0.02} & \aBest{22.70}\aSd{0.06} & \underline{0.053} & 0.176 & \aBest{0.040} & \underline{21.24} & 15.28 & \aBest{24.33} \\
\aYes & \aNo  & \aNo  & \aYes & \aYes & \underline{0.081}\aSd{0.003} & 0.193\aSd{0.001} & \aBest{0.050}\aSd{0.001} & \underline{19.14}\aSd{0.15} & 14.46\aSd{0.01} & \aBest{22.68}\aSd{0.07} & \underline{0.052} & 0.176 & \aBest{0.040} & \underline{21.35} & 15.24 & \aBest{24.29} \\
\aNo  & \aYes & \aYes & \aYes & \aNo  & 0.262\aSd{0.005} & \underline{0.224}\aSd{0.001} & \aBest{0.126}\aSd{0.002} & 10.04\aSd{0.09} & \underline{12.20}\aSd{0.03} & \aBest{14.60}\aSd{0.08} & \underline{0.148} & 0.197 & \aBest{0.092} & \underline{13.56} & 13.26 & \aBest{16.79} \\
\aNo  & \aYes & \aYes & \aNo  & \aYes & \underline{0.083}\aSd{0.002} & 0.192\aSd{0.001} & \aBest{0.053}\aSd{0.001} & \underline{18.97}\aSd{0.14} & 14.44\aSd{0.03} & \aBest{22.30}\aSd{0.07} & \underline{0.053} & 0.176 & \aBest{0.040} & \underline{21.22} & 15.26 & \aBest{24.21} \\
\aNo  & \aYes & \aNo  & \aYes & \aYes & \underline{0.081}\aSd{0.003} & 0.193\aSd{0.001} & \aBest{0.052}\aSd{0.001} & \underline{19.05}\aSd{0.15} & 14.43\aSd{0.03} & \aBest{22.39}\aSd{0.10} & \underline{0.052} & 0.177 & \aBest{0.039} & \underline{21.35} & 15.24 & \aBest{24.22} \\
\aNo  & \aNo  & \aYes & \aYes & \aYes & \underline{0.080}\aSd{0.003} & 0.193\aSd{0.001} & \aBest{0.052}\aSd{0.001} & \underline{19.22}\aSd{0.16} & 14.47\aSd{0.05} & \aBest{22.29}\aSd{0.08} & \underline{0.053} & 0.176 & \aBest{0.040} & \underline{21.26} & 15.30 & \aBest{24.00} \\
\addlinespace[1pt]
\aYes & \aYes & \aNo  & \aNo  & \aNo  & \underline{0.302}\aSd{0.004} & \textbf{0.256}\aSd{0.001} & 0.656\aSd{0.002} & \underline{9.17}\aSd{0.12} & \textbf{11.17}\aSd{0.03} & 8.12\aSd{0.04} & \textbf{0.175} & \underline{0.227} & 0.591 & \textbf{12.37} & \underline{12.22} & 8.93 \\
\aYes & \aNo  & \aYes & \aNo  & \aNo  & 0.369\aSd{0.007} & \underline{0.246}\aSd{0.001} & \aBest{0.136}\aSd{0.002} & 7.90\aSd{0.12} & \underline{10.86}\aSd{0.04} & \aBest{13.94}\aSd{0.08} & 0.228 & \underline{0.212} & \aBest{0.105} & 10.82 & \underline{11.95} & \aBest{15.96} \\
\aYes & \aNo  & \aNo  & \aYes & \aNo  & 0.309\aSd{0.002} & \underline{0.221}\aSd{0.001} & \aBest{0.113}\aSd{0.001} & 8.90\aSd{0.07} & \underline{12.06}\aSd{0.04} & \aBest{15.35}\aSd{0.05} & \underline{0.182} & 0.192 & \aBest{0.089} & 12.14 & \underline{13.18} & \aBest{17.17} \\
\aYes & \aNo  & \aNo  & \aNo  & \aYes & \underline{0.080}\aSd{0.002} & 0.192\aSd{0.001} & \aBest{0.052}\aSd{0.001} & \underline{19.13}\aSd{0.13} & 14.45\aSd{0.03} & \aBest{22.60}\aSd{0.07} & \underline{0.053} & 0.176 & \aBest{0.040} & \underline{21.28} & 15.23 & \aBest{24.29} \\
\aNo  & \aYes & \aYes & \aNo  & \aNo  & 0.302\aSd{0.003} & \underline{0.256}\aSd{0.001} & \aBest{0.202}\aSd{0.002} & 9.17\aSd{0.08} & \underline{11.06}\aSd{0.04} & \aBest{11.41}\aSd{0.08} & \underline{0.175} & 0.224 & \aBest{0.127} & \underline{12.55} & 12.10 & \aBest{14.39} \\
\aNo  & \aYes & \aNo  & \aYes & \aNo  & 0.274\aSd{0.003} & \underline{0.219}\aSd{0.001} & \aBest{0.133}\aSd{0.002} & 9.78\aSd{0.07} & \underline{12.26}\aSd{0.04} & \aBest{14.13}\aSd{0.08} & \underline{0.158} & 0.192 & \aBest{0.095} & 13.22 & \underline{13.35} & \aBest{16.49} \\
\aNo  & \aYes & \aNo  & \aNo  & \aYes & \underline{0.080}\aSd{0.002} & 0.193\aSd{0.001} & \aBest{0.054}\aSd{0.001} & \underline{19.12}\aSd{0.11} & 14.41\aSd{0.02} & \aBest{22.26}\aSd{0.07} & \underline{0.053} & 0.176 & \aBest{0.039} & \underline{21.26} & 15.23 & \aBest{24.29} \\
\aNo  & \aNo  & \aYes & \aYes & \aNo  & 0.293\aSd{0.007} & \underline{0.217}\aSd{0.002} & \aBest{0.126}\aSd{0.002} & 9.15\aSd{0.17} & \underline{12.02}\aSd{0.04} & \aBest{14.60}\aSd{0.08} & \underline{0.170} & 0.187 & \aBest{0.092} & 12.55 & \underline{13.13} & \aBest{16.79} \\
\aNo  & \aNo  & \aYes & \aNo  & \aYes & \underline{0.078}\aSd{0.003} & 0.193\aSd{0.001} & \aBest{0.053}\aSd{0.001} & \underline{19.24}\aSd{0.13} & 14.46\aSd{0.02} & \aBest{22.30}\aSd{0.07} & \underline{0.053} & 0.176 & \aBest{0.040} & \underline{21.22} & 15.28 & \aBest{24.21} \\
\aNo  & \aNo  & \aNo  & \aYes & \aYes & \underline{0.078}\aSd{0.002} & 0.194\aSd{0.000} & \aBest{0.052}\aSd{0.001} & \underline{19.27}\aSd{0.14} & 14.43\aSd{0.03} & \aBest{22.39}\aSd{0.10} & \underline{0.052} & 0.177 & \aBest{0.039} & \underline{21.37} & 15.25 & \aBest{24.22} \\
\addlinespace[1pt]
\aYes & \aNo  & \aNo  & \aNo  & \aNo  & \underline{0.415}\aSd{0.004} & \textbf{0.321}\aSd{0.002} & 0.656\aSd{0.002} & 7.52\aSd{0.08} & \underline{7.78}\aSd{0.04} & \aBest{8.12}\aSd{0.04} & \textbf{0.252} & \underline{0.272} & 0.591 & \textbf{10.42} & 8.79 & \underline{8.93} \\
\aNo  & \aNo  & \aYes & \aNo  & \aNo  & 0.381\aSd{0.007} & \underline{0.249}\aSd{0.002} & \aBest{0.202}\aSd{0.002} & 7.82\aSd{0.09} & \underline{10.76}\aSd{0.04} & \aBest{11.41}\aSd{0.08} & 0.233 & \underline{0.214} & \aBest{0.127} & 10.80 & \underline{11.82} & \aBest{14.39} \\
\aNo  & \aNo  & \aNo  & \aYes & \aNo  & 0.314\aSd{0.004} & \underline{0.222}\aSd{0.002} & \aBest{0.133}\aSd{0.002} & 8.83\aSd{0.09} & \underline{12.01}\aSd{0.04} & \aBest{14.13}\aSd{0.08} & \underline{0.182} & 0.190 & \aBest{0.095} & 12.10 & \underline{13.13} & \aBest{16.49} \\
\aNo  & \aNo  & \aNo  & \aNo  & \aYes & \underline{0.079}\aSd{0.003} & 0.193\aSd{0.000} & \aBest{0.054}\aSd{0.001} & \underline{19.19}\aSd{0.11} & 14.42\aSd{0.03} & \aBest{22.26}\aSd{0.07} & \underline{0.053} & 0.176 & \aBest{0.039} & \underline{21.27} & 15.24 & \aBest{24.29} \\
\bottomrule
\end{tabular}}
\aSweepNote{\aSweepLULC\space NDVI from each model's generated L2A; PSNR on $(\mathrm{NDVI}+1)/2$.}
\endgroup
\end{table*}

% ============================================================
% Land-cover prediction
% ============================================================
\begin{table*}[t]
\centering
\caption{\textbf{EAGLE has the highest mean-of-8 mIoU in 18 of 31 configurations; TerraMind-B has the highest top-1 accuracy in 29.}
Top-1 acc. $\uparrow$ and mIoU $\uparrow$. \aSweepEval}
\label{tab:full-lulc}
\begingroup
\aSweepSetup
\resizebox{\textwidth}{!}{%
\begin{tabular}{@{}ccccc rr>{\columncolor{A2Eshade}}r rr>{\columncolor{A2Eshade}}r rr>{\columncolor{A2Eshade}}r rr>{\columncolor{A2Eshade}}r@{}}
\toprule
\multicolumn{5}{c}{\textbf{Available inputs}} & \multicolumn{6}{c}{\textbf{Mean of 8 draws}} & \multicolumn{6}{c}{\textbf{Best of 8 draws (oracle)}} \\
\cmidrule(lr){1-5}\cmidrule(lr){6-11}\cmidrule(l){12-17}
& & & & & \multicolumn{3}{c}{top-1 acc. $\uparrow$} & \multicolumn{3}{c}{mIoU $\uparrow$} & \multicolumn{3}{c}{top-1 acc. $\uparrow$} & \multicolumn{3}{c}{mIoU $\uparrow$} \\
\cmidrule(l){6-8}\cmidrule(l){9-11}\cmidrule(l){12-14}\cmidrule(l){15-17}
Coord. & DEM & S1 & L1C & L2A & COP-GEN & TerraMind & \aOurs & COP-GEN & TerraMind & \aOurs & COP-GEN & TerraMind & \aOurs & COP-GEN & TerraMind & \aOurs \\
\midrule
\aYes & \aYes & \aYes & \aYes & \aYes & 0.489\aSd{0.006} & \textbf{0.891}\aSd{0.001} & \underline{0.881}\aSd{0.002} & 0.165\aSd{0.004} & \underline{0.565}\aSd{0.003} & \aBest{0.581}\aSd{0.004} & 0.745 & \textbf{0.904} & \underline{0.898} & 0.304 & \textbf{0.629} & \underline{0.612} \\
\addlinespace[1pt]
\aYes & \aYes & \aYes & \aYes & \aNo  & 0.510\aSd{0.008} & \textbf{0.890}\aSd{0.001} & \underline{0.871}\aSd{0.001} & 0.172\aSd{0.005} & \underline{0.564}\aSd{0.002} & \aBest{0.571}\aSd{0.001} & 0.749 & \textbf{0.902} & \underline{0.891} & 0.309 & \textbf{0.627} & \underline{0.597} \\
\aYes & \aYes & \aYes & \aNo  & \aYes & 0.518\aSd{0.008} & \textbf{0.892}\aSd{0.000} & \underline{0.879}\aSd{0.002} & 0.178\aSd{0.005} & \underline{0.570}\aSd{0.003} & \aBest{0.578}\aSd{0.003} & 0.755 & \textbf{0.905} & \underline{0.900} & 0.323 & \textbf{0.632} & \underline{0.611} \\
\aYes & \aYes & \aNo  & \aYes & \aYes & 0.488\aSd{0.007} & \textbf{0.887}\aSd{0.000} & \underline{0.876}\aSd{0.001} & 0.164\aSd{0.005} & \underline{0.560}\aSd{0.004} & \aBest{0.573}\aSd{0.003} & 0.747 & \underline{0.900} & \aBest{0.902} & 0.305 & \textbf{0.625} & \underline{0.617} \\
\aYes & \aNo  & \aYes & \aYes & \aYes & 0.495\aSd{0.005} & \textbf{0.891}\aSd{0.000} & \underline{0.878}\aSd{0.002} & 0.165\aSd{0.004} & \underline{0.563}\aSd{0.004} & \aBest{0.574}\aSd{0.002} & 0.751 & \textbf{0.904} & \underline{0.901} & 0.309 & \textbf{0.627} & \underline{0.618} \\
\aNo  & \aYes & \aYes & \aYes & \aYes & 0.494\aSd{0.010} & \textbf{0.891}\aSd{0.001} & \underline{0.861}\aSd{0.004} & 0.167\aSd{0.004} & \textbf{0.566}\aSd{0.005} & \underline{0.564}\aSd{0.004} & \underline{0.754} & \textbf{0.904} & \aBest{0.904} & 0.310 & \textbf{0.628} & \underline{0.626} \\
\addlinespace[1pt]
\aYes & \aYes & \aYes & \aNo  & \aNo  & 0.306\aSd{0.009} & \textbf{0.783}\aSd{0.002} & \underline{0.777}\aSd{0.005} & 0.078\aSd{0.003} & \underline{0.358}\aSd{0.002} & \aBest{0.435}\aSd{0.005} & 0.593 & \underline{0.809} & \aBest{0.839} & 0.173 & \underline{0.439} & \aBest{0.509} \\
\aYes & \aYes & \aNo  & \aYes & \aNo  & 0.490\aSd{0.009} & \textbf{0.885}\aSd{0.001} & \underline{0.865}\aSd{0.002} & 0.162\aSd{0.005} & \textbf{0.559}\aSd{0.002} & \underline{0.556}\aSd{0.002} & 0.739 & \textbf{0.898} & \underline{0.891} & 0.299 & \textbf{0.622} & \underline{0.596} \\
\aYes & \aYes & \aNo  & \aNo  & \aYes & 0.502\aSd{0.010} & \textbf{0.888}\aSd{0.000} & \underline{0.872}\aSd{0.002} & 0.171\aSd{0.005} & \underline{0.564}\aSd{0.004} & \aBest{0.568}\aSd{0.003} & 0.749 & \underline{0.901} & \aBest{0.902} & 0.315 & \textbf{0.626} & \underline{0.618} \\
\aYes & \aNo  & \aYes & \aYes & \aNo  & 0.505\aSd{0.006} & \textbf{0.889}\aSd{0.001} & \underline{0.867}\aSd{0.002} & 0.169\aSd{0.004} & \underline{0.560}\aSd{0.004} & \aBest{0.565}\aSd{0.002} & 0.752 & \textbf{0.902} & \underline{0.889} & 0.305 & \textbf{0.624} & \underline{0.598} \\
\aYes & \aNo  & \aYes & \aNo  & \aYes & 0.509\aSd{0.009} & \textbf{0.891}\aSd{0.001} & \underline{0.878}\aSd{0.002} & 0.173\aSd{0.004} & \underline{0.564}\aSd{0.004} & \aBest{0.575}\aSd{0.003} & 0.755 & \textbf{0.905} & \underline{0.900} & 0.318 & \textbf{0.628} & \underline{0.618} \\
\aYes & \aNo  & \aNo  & \aYes & \aYes & 0.490\aSd{0.009} & \textbf{0.887}\aSd{0.001} & \underline{0.874}\aSd{0.003} & 0.163\aSd{0.003} & \underline{0.561}\aSd{0.003} & \aBest{0.571}\aSd{0.002} & 0.752 & \underline{0.900} & \aBest{0.904} & 0.309 & \textbf{0.625} & \underline{0.622} \\
\aNo  & \aYes & \aYes & \aYes & \aNo  & 0.514\aSd{0.010} & \textbf{0.889}\aSd{0.000} & \underline{0.856}\aSd{0.003} & 0.174\aSd{0.005} & \textbf{0.564}\aSd{0.004} & \underline{0.553}\aSd{0.004} & 0.755 & \textbf{0.902} & \underline{0.897} & 0.315 & \textbf{0.627} & \underline{0.609} \\
\aNo  & \aYes & \aYes & \aNo  & \aYes & 0.508\aSd{0.013} & \textbf{0.891}\aSd{0.000} & \underline{0.863}\aSd{0.004} & 0.173\aSd{0.006} & \underline{0.563}\aSd{0.003} & \aBest{0.564}\aSd{0.003} & 0.752 & \underline{0.904} & \aBest{0.907} & 0.316 & \textbf{0.629} & \underline{0.623} \\
\aNo  & \aYes & \aNo  & \aYes & \aYes & 0.493\aSd{0.010} & \textbf{0.885}\aSd{0.001} & \underline{0.855}\aSd{0.003} & 0.165\aSd{0.003} & \textbf{0.560}\aSd{0.005} & \underline{0.555}\aSd{0.003} & 0.747 & \underline{0.898} & \aBest{0.905} & 0.309 & \underline{0.624} & \aBest{0.625} \\
\aNo  & \aNo  & \aYes & \aYes & \aYes & 0.500\aSd{0.008} & \textbf{0.891}\aSd{0.000} & \underline{0.854}\aSd{0.004} & 0.169\aSd{0.005} & \textbf{0.563}\aSd{0.004} & \underline{0.556}\aSd{0.004} & 0.751 & \underline{0.904} & \aBest{0.905} & 0.310 & \textbf{0.630} & \underline{0.622} \\
\addlinespace[1pt]
\aYes & \aYes & \aNo  & \aNo  & \aNo  & 0.189\aSd{0.009} & \underline{0.599}\aSd{0.001} & \aBest{0.680}\aSd{0.008} & 0.041\aSd{0.001} & \underline{0.201}\aSd{0.004} & \aBest{0.349}\aSd{0.005} & 0.446 & \underline{0.640} & \aBest{0.767} & 0.109 & \underline{0.258} & \aBest{0.427} \\
\aYes & \aNo  & \aYes & \aNo  & \aNo  & 0.280\aSd{0.008} & \textbf{0.779}\aSd{0.001} & \underline{0.768}\aSd{0.005} & 0.069\aSd{0.003} & \underline{0.350}\aSd{0.004} & \aBest{0.424}\aSd{0.005} & 0.564 & \underline{0.803} & \aBest{0.837} & 0.152 & \underline{0.425} & \aBest{0.505} \\
\aYes & \aNo  & \aNo  & \aYes & \aNo  & 0.495\aSd{0.009} & \textbf{0.884}\aSd{0.001} & \underline{0.858}\aSd{0.003} & 0.162\aSd{0.004} & \textbf{0.556}\aSd{0.003} & \underline{0.547}\aSd{0.003} & 0.753 & \textbf{0.897} & \underline{0.889} & 0.298 & \textbf{0.618} & \underline{0.592} \\
\aYes & \aNo  & \aNo  & \aNo  & \aYes & 0.498\aSd{0.012} & \textbf{0.888}\aSd{0.001} & \underline{0.876}\aSd{0.001} & 0.167\aSd{0.006} & \underline{0.562}\aSd{0.005} & \aBest{0.573}\aSd{0.002} & 0.755 & \underline{0.901} & \aBest{0.904} & \underline{0.313} & \textbf{0.623} & \aBest{0.623} \\
\aNo  & \aYes & \aYes & \aNo  & \aNo  & 0.293\aSd{0.009} & \textbf{0.767}\aSd{0.001} & \underline{0.729}\aSd{0.007} & 0.075\aSd{0.002} & \underline{0.342}\aSd{0.004} & \aBest{0.394}\aSd{0.006} & 0.580 & \underline{0.794} & \aBest{0.826} & 0.169 & \underline{0.414} & \aBest{0.495} \\
\aNo  & \aYes & \aNo  & \aYes & \aNo  & 0.500\aSd{0.008} & \textbf{0.884}\aSd{0.001} & \underline{0.844}\aSd{0.006} & 0.166\aSd{0.002} & \textbf{0.556}\aSd{0.005} & \underline{0.537}\aSd{0.005} & 0.754 & \textbf{0.897} & \underline{0.896} & 0.306 & \textbf{0.618} & \underline{0.605} \\
\aNo  & \aYes & \aNo  & \aNo  & \aYes & 0.503\aSd{0.005} & \textbf{0.884}\aSd{0.000} & \underline{0.854}\aSd{0.005} & 0.169\aSd{0.004} & \textbf{0.560}\aSd{0.003} & \underline{0.550}\aSd{0.005} & 0.748 & \underline{0.898} & \aBest{0.900} & 0.306 & \textbf{0.624} & \underline{0.618} \\
\aNo  & \aNo  & \aYes & \aYes & \aNo  & 0.507\aSd{0.010} & \textbf{0.890}\aSd{0.001} & \underline{0.846}\aSd{0.004} & 0.171\aSd{0.008} & \textbf{0.564}\aSd{0.003} & \underline{0.545}\aSd{0.005} & 0.755 & \textbf{0.902} & \underline{0.898} & 0.311 & \textbf{0.627} & \underline{0.606} \\
\aNo  & \aNo  & \aYes & \aNo  & \aYes & 0.519\aSd{0.007} & \textbf{0.892}\aSd{0.001} & \underline{0.859}\aSd{0.006} & 0.181\aSd{0.003} & \textbf{0.561}\aSd{0.003} & \underline{0.557}\aSd{0.007} & 0.758 & \underline{0.905} & \aBest{0.908} & \underline{0.326} & \textbf{0.626} & \aBest{0.626} \\
\aNo  & \aNo  & \aNo  & \aYes & \aYes & 0.493\aSd{0.015} & \textbf{0.884}\aSd{0.001} & \underline{0.849}\aSd{0.002} & 0.165\aSd{0.007} & \textbf{0.558}\aSd{0.004} & \underline{0.544}\aSd{0.002} & 0.753 & \underline{0.898} & \aBest{0.902} & 0.309 & \textbf{0.623} & \underline{0.616} \\
\addlinespace[1pt]
\aYes & \aNo  & \aNo  & \aNo  & \aNo  & 0.160\aSd{0.006} & \underline{0.176}\aSd{0.002} & \aBest{0.287}\aSd{0.003} & 0.036\aSd{0.002} & \underline{0.039}\aSd{0.000} & \aBest{0.171}\aSd{0.002} & \textbf{0.419} & 0.239 & \underline{0.330} & \underline{0.102} & 0.055 & \aBest{0.197} \\
\aNo  & \aYes & \aNo  & \aNo  & \aNo  & 0.180\aSd{0.006} & \textbf{0.593}\aSd{0.002} & \underline{0.537}\aSd{0.010} & 0.039\aSd{0.002} & \underline{0.196}\aSd{0.002} & \aBest{0.247}\aSd{0.006} & 0.445 & \underline{0.638} & \aBest{0.715} & 0.104 & \underline{0.253} & \aBest{0.368} \\
\aNo  & \aNo  & \aYes & \aNo  & \aNo  & 0.278\aSd{0.012} & \textbf{0.774}\aSd{0.001} & \underline{0.699}\aSd{0.011} & 0.067\aSd{0.004} & \underline{0.341}\aSd{0.004} & \aBest{0.367}\aSd{0.006} & 0.571 & \underline{0.801} & \aBest{0.810} & 0.157 & \underline{0.428} & \aBest{0.475} \\
\aNo  & \aNo  & \aNo  & \aYes & \aNo  & 0.501\aSd{0.011} & \textbf{0.884}\aSd{0.001} & \underline{0.826}\aSd{0.006} & 0.165\aSd{0.006} & \textbf{0.558}\aSd{0.002} & \underline{0.517}\aSd{0.005} & 0.755 & \textbf{0.897} & \underline{0.892} & 0.307 & \textbf{0.620} & \underline{0.593} \\
\aNo  & \aNo  & \aNo  & \aNo  & \aYes & 0.502\aSd{0.008} & \textbf{0.885}\aSd{0.000} & \underline{0.850}\aSd{0.004} & 0.165\aSd{0.003} & \textbf{0.555}\aSd{0.003} & \underline{0.545}\aSd{0.002} & 0.754 & \underline{0.899} & \aBest{0.905} & 0.302 & \underline{0.620} & \aBest{0.621} \\
\bottomrule
\end{tabular}}
\aSweepNote{}
\endgroup
\end{table*}

% ============================================================
% Geolocation
% ============================================================
\begin{table*}[t]
\centering
\caption{\textbf{Location error for each input configuration.}
Distance (km) $\downarrow$. \aSweepEval}
\label{tab:full-geolocation}
\begingroup
\aSweepSetup
\begin{tabular*}{\textwidth}{@{\extracolsep{\fill}}ccccc rr>{\columncolor{A2Eshade}}r rr>{\columncolor{A2Eshade}}r@{}}
\toprule
\multicolumn{5}{c}{\textbf{Available inputs}} & \multicolumn{3}{c}{\textbf{Mean of 8 draws}} & \multicolumn{3}{c}{\textbf{Best of 8 draws (oracle)}} \\
\cmidrule(lr){1-5}\cmidrule(lr){6-8}\cmidrule(l){9-11}
& & & & & \multicolumn{3}{c}{distance (km) $\downarrow$} & \multicolumn{3}{c}{distance (km) $\downarrow$} \\
\cmidrule(l){6-8}\cmidrule(l){9-11}
LULC$^{\ddagger}$ & DEM & S1 & L1C & L2A & COP-GEN & TerraMind & \aOurs & COP-GEN & TerraMind & \aOurs \\
\midrule
\aYes & \aYes & \aYes & \aYes & \aYes & 1227\aSd{61} & \underline{705}\aSd{62} & \aBest{703}\aSd{0} & \underline{441} & \textbf{238} & 703 \\
\addlinespace[1pt]
\aYes & \aYes & \aYes & \aYes & \aNo  & \underline{1449}\aSd{92} & \textbf{627}\aSd{57} & 2103\aSd{0} & \underline{482} & \textbf{250} & 2103 \\
\aYes & \aYes & \aYes & \aNo  & \aYes & 1391\aSd{56} & \textbf{711}\aSd{49} & \underline{821}\aSd{0} & \underline{405} & \textbf{285} & 821 \\
\aYes & \aYes & \aNo  & \aYes & \aYes & 1563\aSd{81} & \textbf{674}\aSd{39} & \underline{1131}\aSd{0} & \underline{492} & \textbf{261} & 1131 \\
\aYes & \aNo  & \aYes & \aYes & \aYes & 1979\aSd{101} & \textbf{707}\aSd{37} & \underline{970}\aSd{0} & \underline{596} & \textbf{276} & 970 \\
\aNo  & \aYes & \aYes & \aYes & \aYes & 1310\aSd{63} & \textbf{662}\aSd{37} & \underline{703}\aSd{0} & \underline{411} & \textbf{242} & 703 \\
\addlinespace[1pt]
\aYes & \aYes & \aYes & \aNo  & \aNo  & 6040\aSd{219} & \textbf{1666}\aSd{80} & \underline{2734}\aSd{0} & \underline{1569} & \textbf{409} & 2734 \\
\aYes & \aYes & \aNo  & \aYes & \aNo  & \underline{1820}\aSd{111} & \textbf{634}\aSd{37} & 2499\aSd{0} & \underline{586} & \textbf{288} & 2499 \\
\aYes & \aYes & \aNo  & \aNo  & \aYes & 1787\aSd{67} & \textbf{643}\aSd{54} & \underline{1229}\aSd{0} & \underline{538} & \textbf{244} & 1229 \\
\aYes & \aNo  & \aYes & \aYes & \aNo  & \underline{2233}\aSd{112} & \textbf{722}\aSd{74} & 2402\aSd{0} & \underline{493} & \textbf{241} & 2402 \\
\aYes & \aNo  & \aYes & \aNo  & \aYes & 2417\aSd{153} & \textbf{701}\aSd{45} & \underline{1266}\aSd{0} & \underline{584} & \textbf{254} & 1266 \\
\aYes & \aNo  & \aNo  & \aYes & \aYes & 2324\aSd{127} & \textbf{651}\aSd{44} & \underline{1780}\aSd{0} & \underline{622} & \textbf{261} & 1780 \\
\aNo  & \aYes & \aYes & \aYes & \aNo  & \underline{1597}\aSd{98} & \textbf{608}\aSd{53} & 2103\aSd{0} & \underline{392} & \textbf{245} & 2103 \\
\aNo  & \aYes & \aYes & \aNo  & \aYes & 1476\aSd{79} & \textbf{663}\aSd{27} & \underline{821}\aSd{0} & \underline{413} & \textbf{270} & 821 \\
\aNo  & \aYes & \aNo  & \aYes & \aYes & 1646\aSd{112} & \textbf{576}\aSd{53} & \underline{1131}\aSd{0} & \underline{455} & \textbf{241} & 1131 \\
\aNo  & \aNo  & \aYes & \aYes & \aYes & 2051\aSd{119} & \textbf{580}\aSd{57} & \underline{970}\aSd{0} & \underline{461} & \textbf{204} & 970 \\
\addlinespace[1pt]
\aYes & \aYes & \aNo  & \aNo  & \aNo  & 7031\aSd{140} & \textbf{3450}\aSd{66} & \underline{5994}\aSd{0} & \underline{2384} & \textbf{1629} & 5994 \\
\aYes & \aNo  & \aYes & \aNo  & \aNo  & 6970\aSd{221} & \textbf{2414}\aSd{104} & \underline{3466}\aSd{0} & \underline{2103} & \textbf{708} & 3466 \\
\aYes & \aNo  & \aNo  & \aYes & \aNo  & \underline{2722}\aSd{124} & \textbf{709}\aSd{65} & 3636\aSd{0} & \underline{631} & \textbf{301} & 3636 \\
\aYes & \aNo  & \aNo  & \aNo  & \aYes & 3068\aSd{158} & \textbf{723}\aSd{45} & \underline{1922}\aSd{0} & \underline{712} & \textbf{228} & 1922 \\
\aNo  & \aYes & \aYes & \aNo  & \aNo  & 7134\aSd{214} & \textbf{1698}\aSd{64} & \underline{2734}\aSd{0} & \underline{1980} & \textbf{558} & 2734 \\
\aNo  & \aYes & \aNo  & \aYes & \aNo  & \underline{2034}\aSd{59} & \textbf{573}\aSd{54} & 2499\aSd{0} & \underline{547} & \textbf{240} & 2499 \\
\aNo  & \aYes & \aNo  & \aNo  & \aYes & 1989\aSd{129} & \textbf{580}\aSd{47} & \underline{1229}\aSd{0} & \underline{452} & \textbf{251} & 1229 \\
\aNo  & \aNo  & \aYes & \aYes & \aNo  & 2439\aSd{95} & \textbf{634}\aSd{50} & \underline{2402}\aSd{0} & \underline{585} & \textbf{180} & 2402 \\
\aNo  & \aNo  & \aYes & \aNo  & \aYes & 2585\aSd{142} & \textbf{651}\aSd{53} & \underline{1266}\aSd{0} & \underline{549} & \textbf{180} & 1266 \\
\aNo  & \aNo  & \aNo  & \aYes & \aYes & 2765\aSd{79} & \textbf{636}\aSd{60} & \underline{1780}\aSd{0} & \underline{701} & \textbf{256} & 1780 \\
\addlinespace[1pt]
\aNo  & \aYes & \aNo  & \aNo  & \aNo  & 7977\aSd{211} & \textbf{3682}\aSd{121} & \underline{5994}\aSd{0} & \underline{3084} & \textbf{1584} & 5994 \\
\aNo  & \aNo  & \aYes & \aNo  & \aNo  & 7879\aSd{127} & \textbf{2187}\aSd{69} & \underline{3466}\aSd{0} & \underline{2803} & \textbf{595} & 3466 \\
\aNo  & \aNo  & \aNo  & \aYes & \aNo  & \underline{2919}\aSd{124} & \textbf{628}\aSd{62} & 3636\aSd{0} & \underline{728} & \textbf{255} & 3636 \\
\aNo  & \aNo  & \aNo  & \aNo  & \aYes & 3341\aSd{223} & \textbf{627}\aSd{94} & \underline{1922}\aSd{0} & \underline{743} & \textbf{263} & 1922 \\
\bottomrule
\end{tabular*}
\aSweepNote{\aSweepLULC\space EAGLE gives one geocell estimate per tile, so its mean and best of 8 coincide.}
\endgroup
\end{table*}

%% file: Sup/tab_uncertainty.tex
% Generated by scripts/paper/build_uncertainty_table.py (data: results/uncertainty500/{per_tile.parquet,summary.csv,selective.csv})
% Needs \usepackage{booktabs,array,colortbl} and \usepackage[table]{xcolor}.
\definecolor{A2Eshade}{HTML}{EDF4F4}
\definecolor{A2Eink}{HTML}{204E56}
\providecommand{\aBest}[1]{\textcolor{A2Eink}{\textbf{#1}}}
\providecommand{\aOurs}{\textcolor{A2Eink}{\textbf{EAGLE}}}
\providecommand{\aScoreCI}[2]{#1\,{\fontsize{6}{7}\selectfont[#2]}}
\begin{table*}[t]
\centering
\caption{\textbf{EAGLE has the lowest CRPS in nine of ten settings.} 500 scenes (draws separate from those of
\crefrange{tab:full-dem}{tab:full-geolocation}),
8 draws, same inputs for every model. Std: spread of the draws; RMSE: error of the ensemble mean;
SSR $=\sqrt{9/8}\,$Std$/$RMSE; Cov$_{90}$: coverage of the central 90\% interval from the randomised PIT of the truth's rank among the draws (0.90 when
calibrated, for any number of draws); CRPS: fair estimator;
$\rho$: Spearman correlation of per-scene spread and error [95\% interval]. LULC: Disagr., draws differing from the
majority class; Brier and NLL of the class frequencies. S2 in reflectance, S1 in dB, DEM in m. \textbf{Bold}: best
(SSR closest to 1, Cov$_{90}$ to 0.90).}
\label{tab:uncertainty}
\scriptsize\setlength{\tabcolsep}{2pt}\renewcommand{\arraystretch}{0.95}
\begin{tabular*}{\textwidth}{@{\extracolsep{\fill}}lll rrrrrl@{}}
\toprule
\textbf{Target} & \textbf{Input} & \textbf{Model} & \textbf{Std} & \textbf{RMSE $\downarrow$} & \textbf{SSR $\to$ 1} &
\textbf{Cov$_{90}$ $\to$ 0.90} & \textbf{CRPS $\downarrow$} & \textbf{$\rho$ (spread, error) $\uparrow$} \\
\midrule
DEM (m) & S1 & COP-GEN & 1326.9 & 904.2 & 1.56 & 0.95 & 376.4 & \aScoreCI{\textbf{0.59}}{0.51, 0.65} \\
 &  & TerraMind-B & 67.0 & 648.7 & 0.11 & 0.63 & 273.3 & \aScoreCI{0.37}{0.26, 0.47} \\
\rowcolor{A2Eshade}\cellcolor{white} &\cellcolor{white}  & \aOurs & 546.7 & \aBest{512.2} & \aBest{1.13} & \aBest{0.93} & \aBest{197.1} & \aScoreCI{0.55}{0.47, 0.61} \\
\addlinespace[2pt]
 & S2 L2A & COP-GEN & 893.5 & 679.9 & \textbf{1.39} & 0.94 & 251.7 & \aScoreCI{\textbf{0.70}}{0.64, 0.75} \\
 &  & TerraMind-B & 32.3 & 593.8 & 0.06 & 0.63 & 187.5 & \aScoreCI{0.39}{0.28, 0.49} \\
\rowcolor{A2Eshade}\cellcolor{white} &\cellcolor{white}  & \aOurs & 456.6 & \aBest{290.8} & 1.67 & \aBest{0.93} & \aBest{124.3} & \aScoreCI{0.55}{0.48, 0.62} \\
\addlinespace[2pt]
 & all others & COP-GEN & 406.1 & 326.1 & 1.32 & 0.91 & 151.0 & \aScoreCI{0.54}{0.47, 0.61} \\
 &  & TerraMind-B & 29.4 & 590.2 & 0.05 & 0.63 & 178.1 & \aScoreCI{0.32}{0.21, 0.43} \\
\rowcolor{A2Eshade}\cellcolor{white} &\cellcolor{white}  & \aOurs & 82.1 & \aBest{112.2} & \aBest{0.78} & \aBest{0.90} & \aBest{49.4} & \aScoreCI{\aBest{0.64}}{0.58, 0.70} \\
\addlinespace[2pt]
S2 L1C & S1 & COP-GEN & 0.1394 & 0.1731 & \textbf{0.85} & 0.84 & 0.0843 & \aScoreCI{0.46}{0.39, 0.52} \\
 &  & TerraMind-B & 0.0427 & 0.1993 & 0.23 & 0.70 & 0.1063 & \aScoreCI{0.15}{0.06, 0.24} \\
\rowcolor{A2Eshade}\cellcolor{white} &\cellcolor{white}  & \aOurs & 0.0477 & \aBest{0.1021} & 0.50 & \aBest{0.85} & \aBest{0.0389} & \aScoreCI{\aBest{0.59}}{0.52, 0.65} \\
\addlinespace[2pt]
 & S2 L2A & COP-GEN & 0.0532 & 0.0670 & \textbf{0.84} & 0.87 & 0.0378 & \aScoreCI{$-$0.10}{$-$0.20, 0.01} \\
 &  & TerraMind-B & 0.0248 & 0.1945 & 0.14 & 0.65 & 0.0992 & \aScoreCI{0.38}{0.30, 0.46} \\
\rowcolor{A2Eshade}\cellcolor{white} &\cellcolor{white}  & \aOurs & 0.0176 & \aBest{0.0261} & 0.72 & \aBest{0.88} & \aBest{0.0099} & \aScoreCI{\aBest{0.68}}{0.61, 0.73} \\
\addlinespace[2pt]
 & all others & COP-GEN & 0.0490 & 0.0639 & \textbf{0.81} & \textbf{0.86} & 0.0365 & \aScoreCI{$-$0.07}{$-$0.18, 0.04} \\
 &  & TerraMind-B & 0.0248 & 0.1946 & 0.14 & 0.65 & 0.0993 & \aScoreCI{0.38}{0.30, 0.45} \\
\rowcolor{A2Eshade}\cellcolor{white} &\cellcolor{white}  & \aOurs & 0.0116 & \aBest{0.0222} & 0.56 & 0.83 & \aBest{0.0089} & \aScoreCI{\aBest{0.73}}{0.68, 0.78} \\
\addlinespace[2pt]
S2 L2A & S1 & COP-GEN & 0.1665 & 0.2002 & \textbf{0.88} & 0.85 & 0.0945 & \aScoreCI{0.43}{0.35, 0.50} \\
 &  & TerraMind-B & 0.0428 & 0.2078 & 0.22 & 0.70 & 0.1030 & \aScoreCI{0.29}{0.20, 0.38} \\
\rowcolor{A2Eshade}\cellcolor{white} &\cellcolor{white}  & \aOurs & 0.0619 & \aBest{0.1205} & 0.54 & \aBest{0.86} & \aBest{0.0479} & \aScoreCI{\aBest{0.56}}{0.48, 0.62} \\
\addlinespace[2pt]
 & all others & COP-GEN & 0.0396 & \textbf{0.0461} & \textbf{0.91} & \textbf{0.85} & \textbf{0.0181} & \aScoreCI{\textbf{0.76}}{0.72, 0.81} \\
 &  & TerraMind-B & 0.0254 & 0.2023 & 0.13 & 0.69 & 0.0867 & \aScoreCI{0.47}{0.39, 0.54} \\
\rowcolor{A2Eshade}\cellcolor{white} &\cellcolor{white}  & \aOurs & 0.0106 & 0.0490 & 0.23 & 0.73 & 0.0200 & \aScoreCI{0.56}{0.48, 0.63} \\
\addlinespace[2pt]
S1 (dB) & S2 L2A & COP-GEN & 4.24 & 3.57 & \textbf{1.26} & \textbf{0.92} & 1.76 & \aScoreCI{\textbf{0.69}}{0.63, 0.74} \\
 &  & TerraMind-B & 1.69 & 3.41 & 0.53 & 0.79 & 1.88 & \aScoreCI{0.17}{0.08, 0.26} \\
\rowcolor{A2Eshade}\cellcolor{white} &\cellcolor{white}  & \aOurs & 1.59 & \aBest{2.90} & 0.58 & 0.80 & \aBest{1.65} & \aScoreCI{0.52}{0.44, 0.59} \\
\addlinespace[2pt]
 & all others & COP-GEN & 3.60 & 3.48 & \textbf{1.10} & \textbf{0.91} & 1.76 & \aScoreCI{\textbf{0.71}}{0.65, 0.76} \\
 &  & TerraMind-B & 1.69 & 3.43 & 0.52 & 0.79 & 1.89 & \aScoreCI{0.19}{0.10, 0.27} \\
\rowcolor{A2Eshade}\cellcolor{white} &\cellcolor{white}  & \aOurs & 1.32 & \aBest{2.80} & 0.50 & 0.77 & \aBest{1.64} & \aScoreCI{0.59}{0.51, 0.65} \\
\midrule
& & & \textbf{Disagr.} & \textbf{Acc.\ (vote) $\uparrow$} & \textbf{Acc.\ (draw) $\uparrow$} & \textbf{Brier $\downarrow$} &
\textbf{NLL $\downarrow$} & \textbf{$\rho$ (disagr., error) $\uparrow$} \\
\midrule
LULC & S1 & COP-GEN & 0.523 & 0.334 & 0.271 & 0.822 & 1.86 & \aScoreCI{0.16}{0.07, 0.25} \\
 &  & TerraMind-B & 0.102 & \textbf{0.803} & \textbf{0.774} & \textbf{0.312} & \textbf{0.87} & \aScoreCI{\textbf{0.70}}{0.64, 0.76} \\
\rowcolor{A2Eshade}\cellcolor{white} &\cellcolor{white}  & \aOurs & 0.140 & 0.748 & 0.697 & 0.414 & 1.03 & \aScoreCI{0.44}{0.36, 0.52} \\
\addlinespace[2pt]
 & S2 L2A & COP-GEN & 0.374 & 0.627 & 0.505 & 0.510 & 1.25 & \aScoreCI{0.63}{0.57, 0.69} \\
 &  & TerraMind-B & 0.052 & \textbf{0.898} & \textbf{0.885} & \textbf{0.159} & \textbf{0.65} & \aScoreCI{\textbf{0.86}}{0.81, 0.89} \\
\rowcolor{A2Eshade}\cellcolor{white} &\cellcolor{white}  & \aOurs & 0.067 & 0.866 & 0.848 & 0.212 & 0.72 & \aScoreCI{0.75}{0.70, 0.81} \\
\addlinespace[2pt]
 & all others & COP-GEN & 0.378 & 0.596 & 0.489 & 0.539 & 1.29 & \aScoreCI{0.62}{0.55, 0.67} \\
 &  & TerraMind-B & 0.049 & \textbf{0.902} & \textbf{0.891} & \textbf{0.152} & \textbf{0.64} & \aScoreCI{\textbf{0.86}}{0.81, 0.90} \\
\rowcolor{A2Eshade}\cellcolor{white} &\cellcolor{white}  & \aOurs & 0.018 & 0.884 & 0.882 & 0.212 & 0.72 & \aScoreCI{0.77}{0.71, 0.82} \\
\bottomrule
\end{tabular*}
\end{table*}

%% file: Sup/train_distance.tex
% Generated by scripts/paper/build_train_distance_tex.py from results/eval3037/train_distance/bins.csv; style edited by
% hand on 21 Nov (EAGLE columns shaded, \aBest, \cref): re-apply these if the table is regenerated.
\begin{table*}[t]
\centering
\caption{\textbf{EAGLE's error does not grow with distance from its training data.} MAE $\downarrow$ (mean of 20 draws) on
the 3,037 scenes of \cref{tab:peak_capability}, binned by distance to the nearest training tile and by the number of
training tiles within 100\,km; no evaluation tile overlaps a training tile.
Spearman $\rho$ of EAGLE's error with distance: $-0.09$ to $-0.01$. S2 MAE in reflectance $\times$100; TerraM.:
TerraMind-B; $^{*}$L2A from L1C. \textbf{Bold}: lowest; $^{\dagger}$difference to one baseline not significant.}
\label{tab:train_distance}
\scriptsize
\setlength{\tabcolsep}{2pt}
\renewcommand{\arraystretch}{1.05}
\begin{tabular}{@{}lr >{\columncolor{A2Eshade}}ccc >{\columncolor{A2Eshade}}ccc >{\columncolor{A2Eshade}}ccc >{\columncolor{A2Eshade}}ccc@{}}
\toprule
& & \multicolumn{3}{c}{\textbf{DEM} (m)} & \multicolumn{3}{c}{\textbf{S1 RTC} (dB)} & \multicolumn{3}{c}{\textbf{S2 L1C} ($\times$100)} & \multicolumn{3}{c}{\textbf{S2 L2A$^{*}$} ($\times$100)} \\
\cmidrule(lr){3-5}\cmidrule(lr){6-8}\cmidrule(lr){9-11}\cmidrule(l){12-14}
\textbf{Bin} & \textbf{Places} & \aOurs & COP-GEN & TerraM. & \aOurs & COP-GEN & TerraM. & \aOurs & COP-GEN & TerraM. & \aOurs & COP-GEN & TerraM. \\
\midrule
\multicolumn{14}{@{}l}{\textit{Distance to the nearest training tile (km), quintiles}} \\
4.7--10.0 km & 608 & \aBest{90.1} & 321.0 & 178.1 & \aBest{2.39} & 3.69 & 2.79 & \aBest{1.52} & 6.36 & 11.52 & \aBest{2.63} & 3.82 & 10.31 \\
10.0--12.0 km & 607 & \aBest{89.3} & 316.4 & 186.0 & \aBest{2.40} & 3.68 & 2.84 & \aBest{1.33} & 6.24 & 10.21 & \aBest{2.24} & 3.57 & 8.76 \\
12.0--12.7 km & 607 & \aBest{91.4} & 330.6 & 231.9 & \aBest{2.33} & 3.64 & 2.78 & \aBest{1.39} & 6.14 & 10.21 & \aBest{2.44} & 3.74 & 8.74 \\
12.7--15.5 km & 607 & \aBest{86.9} & 319.5 & 233.5 & \aBest{2.35} & 3.67 & 2.85 & \aBest{1.38} & 6.38 & 10.87 & \aBest{2.52} & 3.93 & 9.64 \\
15.5--112.1 km & 608 & \aBest{83.1} & 303.3 & 162.1 & \aBest{2.31} & 3.57 & 2.93 & \aBest{1.36} & 6.25 & 11.36 & \aBest{2.34} & 3.72 & 10.38 \\
\midrule
\multicolumn{14}{@{}l}{\textit{Distance to the nearest training tile, far tail}} \\
$\leq$20 km & 2,770 & \aBest{88.7} & 319.5 & 202.8 & \aBest{2.36} & 3.66 & 2.81 & \aBest{1.40} & 6.29 & 10.67 & \aBest{2.43} & 3.76 & 9.34 \\
20--30 km & 196 & \aBest{86.0} & 318.3 & 171.4 & \aBest{2.26} & 3.55 & 2.88 & \aBest{1.37} & 6.14 & 12.18 & \aBest{2.28} & 3.79 & 11.50 \\
$>$30 km & 71 & \aBest{72.9}$^{\dagger}$ & 265.5 & 96.5 & \aBest{2.37} & 3.73 & 3.81 & \aBest{1.43} & 5.97 & 13.66 & \aBest{2.88} & 3.65 & 13.06 \\
\midrule
\multicolumn{14}{@{}l}{\textit{Training tiles within 100\,km, quintiles}} \\
0--464 & 612 & \aBest{71.8} & 260.7 & 112.4 & \aBest{2.27} & 3.65 & 3.13 & \aBest{1.36} & 6.23 & 12.04 & \aBest{2.29} & 3.81 & 11.22 \\
464--625 & 604 & \aBest{80.5} & 277.3 & 161.2 & \aBest{2.26} & 3.51 & 2.74 & \aBest{1.37} & 6.38 & 11.13 & \aBest{2.33} & 3.73 & 9.97 \\
625--800 & 609 & \aBest{88.4} & 321.0 & 218.9 & \aBest{2.31} & 3.60 & 2.73 & \aBest{1.44} & 6.35 & 11.32 & \aBest{2.60} & 3.80 & 10.14 \\
800--1122 & 604 & \aBest{105.1} & 380.0 & 364.0 & \aBest{2.50} & 3.90 & 2.85 & \aBest{1.52} & 6.43 & 11.21 & \aBest{2.88} & 4.15 & 9.92 \\
1122--10563 & 608 & \aBest{95.3} & 352.4 & 136.4 & \aBest{2.44} & 3.61 & 2.74 & \aBest{1.30} & 5.97 & 8.47 & \aBest{2.06} & 3.29 & 6.58 \\
\bottomrule
\end{tabular}
\end{table*}